\documentclass[manuscript, screen, nonacm]{jair}

\usepackage{multirow}
\RequirePackage[
  datamodel=acmdatamodel,
  style=acmauthoryear,
  backend=bibtex,
  giveninits=true,
  uniquename=init
  ]{biblatex}

\usepackage{amsmath,amsfonts}
\usepackage{algorithmic}
\usepackage{algorithm}
\usepackage{array}
\usepackage[caption=false,font=normalsize,labelfont=sf,textfont=sf]{subfig}
\usepackage{textcomp}
\usepackage{stfloats}
\usepackage{url}
\usepackage{verbatim}
\usepackage{graphicx}
\usepackage{orcidlink}
\usepackage{pdfpages}
\usepackage{xcolor}
\usepackage{forest}
\usepackage{placeins}

\usepackage{tikz}
\usetikzlibrary{arrows.meta, shapes, positioning}

\newcommand{\eg}{\emph{e.g.}}

\newcommand{\etal}{\emph{et al.}}
\newcommand{\keyref}[1]{\citeauthor{#1}~\citeyear{#1}}

\setcopyright{none}
\acmDOI{}
\makeatletter
\def\@JAIRAE{}
\def\@JAIRTrack{}
\def\@mkbibcitation{}
\AtBeginDocument{%
  \fancyfoot{}%
  \fancyfoot[RO,LE]{\footnotesize\thepage}%
  \fancypagestyle{firstpagestyle}{%
    \fancyhf{}%
    \fancyfoot[RO,LE]{\footnotesize\thepage}%
  }%
}
\makeatother

\definecolor{collab-blue}{RGB}{194,232,247}
\definecolor{collab-orange}{RGB}{243,202,120}
\definecolor{collab-red}{RGB}{205,44,36}
\definecolor{collab-green}{RGB}{34,139,34}
\definecolor{collab-yellow}{RGB}{255,228,123}
\definecolor{collab-pink}{RGB}{255,245,247}
\definecolor{collab-dark}{RGB}{20,68,106}
\definecolor{collab-gray}{RGB}{90,90,90}
\definecolor{hidden-black}{RGB}{20,68,106}
\definecolor{darkblue}{rgb}{0, 0.40, 0.75}

\tikzstyle{my-box}=[
  rectangle,
  draw=hidden-black,
  rounded corners,
  text opacity=1,
  minimum height=1.5em,
  minimum width=5em,
  inner sep=2pt,
  align=center,
  fill opacity=.5,
]

\tikzstyle{leaf}=[
    my-box,
    minimum height=1.5em,
    fill=collab-red!20,
    text=black,
    align=left,
    font=\normalsize,
    inner xsep=5pt,
    inner ysep=4pt,
]

\tikzstyle{leaf3}=[
    my-box,
    minimum height=1.5em,
    fill=yellow!32,
    text=black,
    align=left,
    font=\normalsize,
    inner xsep=5pt,
    inner ysep=4pt,
    align=left,
    text width=45em,
]

\tikzstyle{leaf5}=[
    my-box,
    minimum height=1.5em,
    fill=darkblue!15,
    text=black,
    align=left,
    font=\normalsize,
    inner xsep=5pt,
    inner ysep=4pt,
]

\tikzstyle{leaf-blue}=[
    my-box,
    fill=collab-blue!60,
    text width=40em,
]

\tikzstyle{leaf-red}=[
    my-box,
    fill=collab-red,
    text width=40em,
]
\DeclareUnicodeCharacter{0300}{\`{}}
\begin{document}

\title[Towards Harnessing the Collaborative Power of Large and Small Models for Domain Tasks]{Towards Harnessing the Collaborative Power of Large and Small Models for Domain Tasks}

\author{Yang Liu}
\authornote{Corresponding author: yang-veronica.liu@polyu.edu.hk.}
\authornote{Equal contribution.}
\email{yang-veronica.liu@polyu.edu.hk}
\affiliation{
  \institution{The Hong Kong Polytechnic University}
  \city{Hong Kong SAR}
  \country{China}
}

\author{Kejia Zhang}
\authornotemark[2]
\email{kejiaz171@gmail.com}
\affiliation{%
  \institution{The Hong Kong Polytechnic University}
  \city{Hong Kong SAR}
  \country{China}
}

\author{Bingjie Yan}
\affiliation{%
  \institution{The Hong Kong Polytechnic University}
  \city{Hong Kong SAR}
  \country{China}
}

\author{Tianyuan Zou}
\affiliation{%
  \institution{Institute for AI Industry Research, Tsinghua University}
  \city{Beijing}
  \country{China}
}

\author{Jianqing Zhang}
\affiliation{%
  \institution{Institute for AI Industry Research, Tsinghua University}
  \city{Beijing}
  \country{China}
}
\affiliation{%
  \institution{Shanghai Jiao Tong University}
  \city{Shanghai}
  \country{China}
}

\author{Zixuan Gu}
\affiliation{%
  \institution{Institute for AI Industry Research \& School of Software, Tsinghua University}
  \city{Beijing}
  \country{China}
}

\author{Xiangsen Chen}
\affiliation{%
  \institution{The Hong Kong Polytechnic University}
  \city{Hong Kong SAR}
  \country{China}
}

\author{Jianbing Ding}
\affiliation{%
  \institution{AsiaInfo Technologies}
  \city{Beijing}
  \country{China}
}

\author{Xidong Wang}
\affiliation{%
  \institution{AsiaInfo Technologies}
  \city{Beijing}
  \country{China}
}

\author{Jingyi Li}
\affiliation{%
  \institution{AsiaInfo Technologies}
  \city{Beijing}
  \country{China}
}

\author{Xiaozhou Ye}
\affiliation{%
  \institution{AsiaInfo Technologies}
  \city{Beijing}
  \country{China}
}

\author{Ye Ouyang}
\affiliation{%
  \institution{AsiaInfo Technologies}
  \city{Beijing}
  \country{China}
}

\author{Qiang Yang}
\affiliation{%
  \institution{The Hong Kong Polytechnic University}
  \city{Hong Kong SAR}
  \country{China}
}

\author{Ya\--Qin Zhang}
\affiliation{%
  \institution{Institute for AI Industry Research, Tsinghua University}
  \city{Beijing}
  \country{China}
}


\renewcommand{\shortauthors}{Liu \etal}
\makeatletter
\let\@authorsaddresses\@empty
\makeatother

\begin{abstract}
Large language models (LMs) offer broad generalization capabilities but require vast amounts of data and computational resources for domain-specific tasks; small models (SMs), in contrast, are more efficient and tailored to specific domains yet lack general-purpose coverage. Taking a collaborative approach, where large and small models work synergistically, can accelerate the adaptation of LLMs to private domains and unlock new potential in AI. This survey presents a comprehensive overview of recent advances and challenges in harnessing the collaborative power of large and small models for private-domain adaptation. It specifically focuses on the unique constraints of cross-boundary environments, where models belong to distinct parties, and examines the resulting tensions among data privacy, model security, integrity, and resource limitations \footnote{We maintain an open repository tracking representative methods and resources at \url{https://github.com/KejiaZhang-Robust/Awesome-LM-SM-Domain-Collaboration}.}. By analyzing the information flow between distinct model and data stakeholders, we propose a unified taxonomy that classifies research into three primary directions: downward knowledge transfer (LM to SM), upward knowledge transfer (SM to LM), and inference-time collaboration across parties. Drawing on this taxonomy, we analyze the core challenges inherent to cross-boundary information exchange, including data-privacy, model-security, and integrity threats as well as efficiency constraints, and synthesize these into a multi-objective optimization problem that governs practical deployment. Finally, we review key open challenges inherent to such hybrid approaches and outline promising directions for future research. By offering a principled, boundary-centric view of this rapidly evolving landscape, this survey aims to serve as a structured resource for researchers and practitioners advancing privacy-aware, resource-efficient AI deployment.

\end{abstract}

\maketitle

\section{Introduction}

\begin{figure}[t]
    \centering
    \includegraphics[width=0.9\linewidth]{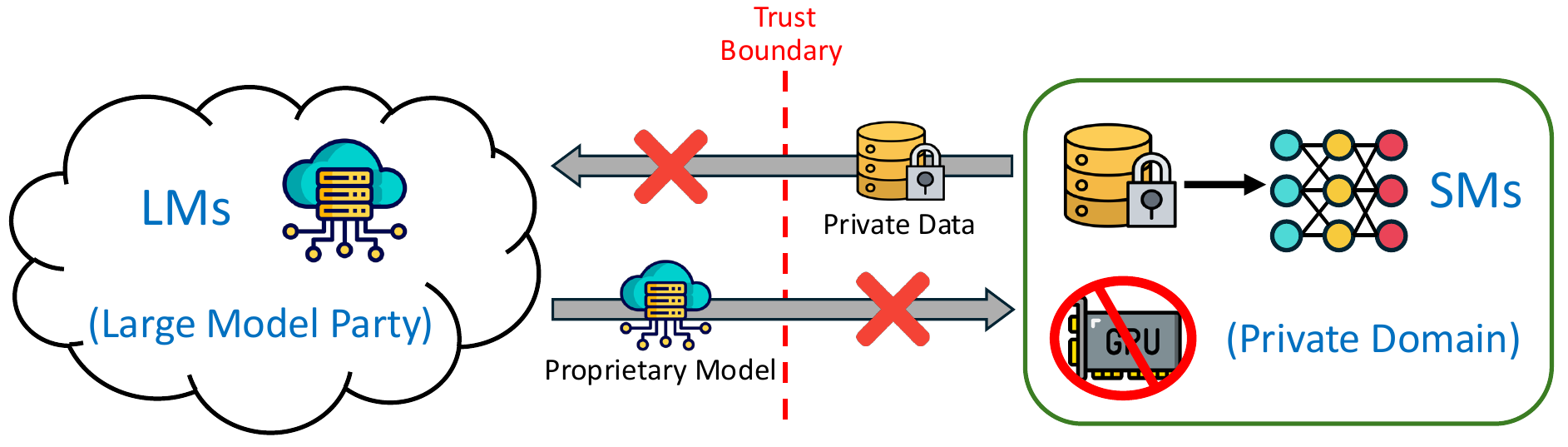}
    \caption{
        Collaboration of LMs and SMs for domain tasks: data privacy, model security, integrity risks, and resource limitations.
    }
    \label{fig:problem}
\end{figure}

The rapid scaling of large language models (LMs), driven by simultaneous growth in model size and training data, has produced systems that perform strongly on open-ended reasoning, code generation, and scientific analysis~\cite{kaplan2020scaling, weiemergent, muennighoff2023scaling, hoffmanntraining, grattafiori2024llama}. As high-quality public data approaches saturation~\cite{villalobos2024, song2025injecting}, the remaining headroom lies in the private domain. Institutions across sectors hold proprietary corpora whose long-tail knowledge and operational patterns are absent from public pre-training distributions~\cite{ren2024advancesopenchallengesfederated, li2024synergizing}, and general-purpose models consistently underperform on such domains without targeted adaptation~\cite{kandpal2023large, gururangan2020don}. However, because private data must remain within its local trust boundary and local environments often lack the computational resources to fine-tune frontier LMs, standard adaptation pipelines remain impractical for real-world private domains~\cite{yang2019federated, carlini2021extracting, menghani2023efficient, xiao2023offsite, li2025bild, chen2025fedc4, yi2025ecoagent}.
For example, one of the most recent open-sourced LLM models DeepSeek-V3~\cite{deepseekai2024deepseekv3technicalreport} has more than 671B parameters, and deploying it directly for domain tasks would require substantial high-speed GPU clusters and communication bandwidth, which would be infeasible for most small and medium-sized enterprises and institutions. 
Substantial investment in resources often leads to strict copyright protections and proprietary ownership of LLMs, hindering the development of new models based on these foundations~\cite{ren2024advancesopenchallengesfederated}.
Nevertheless, there are abundant domain tasks which do not rely on LLMs before. For example, a bank aims to develop an accurate credit rating model using customer financial data, or a security company seeks to build a better hazard detection model using surveillance footage. To achieve these goals, small domain models, such as ResNet~\cite{he2016deep} or LSTM~\cite{hochreiter1997long,chung2014empirical}, are commonly adopted. These small models are usually highly efficient and lightweight and can run on edge devices. While LLMs are gaining prominence, SMs remain crucial for private domain tasks and personal devices.

The complementary nature of large and small models presents a unique opportunity for AI innovation. Large models can enhance the capabilities of small models by sharing their pre-trained knowledge, leading to improved private intelligence. Conversely, small models can transfer their domain knowledge to improve domain-specific understanding in large models, enabling the development of domain-specific large models without compromising data privacy. Ultimately, the collaboration between large and small models opens new avenues for harnessing the best of both worlds, paving the way for efficient, private, and collaborative AI.


This survey examines a collaborative paradigm that leverages the complementary strengths of both model types, addressing the challenges arising from cross-silo collaboration within a unified framework. In this framework, LMs function as external reasoning engines, supplying the broad semantic understanding and generalization capabilities that are absent in compact architectures~\cite{hinton2015distilling, hsieh2023distilling, gunasekar2023textbooks}. Conversely, Small Models~(SMs) serve as resident local agents that operate strictly within the private domain boundary, ensuring rigorous adherence to data sovereignty and resource budgets~\cite{jiao2020tinybert, xu2024small, mehta2024openelm}. 
By fostering this bidirectional collaboration, LMs augment local execution with broad generalization capabilities, while SMs contribute private-domain knowledge back to enrich LM adaptation, together enabling high-utility domain adaptation without compromising data privacy or exposing proprietary model parameters~\cite{10882951, leviathan2023fast, shen2024learning}.

Recent surveys in this direction~\cite{xu2024survey, li2024synergizing} have actively explored the complementary strengths of LMs and SMs, through the lens of engineering mechanism~\cite{chen2025survey, li2025edge}, collaboration objective~\cite{wang2025survey}, or the functional role of small models~\cite{wang2024comprehensive}. However, these works often overlook the practical boundary between data and model parties as an explicit primitive; consequently, they fail to systematically characterize the information crossing this boundary or the associated privacy risks. In contrast, our survey treats the silo boundary as a fundamental constraint. We conduct a granular examination of the transferred carrier, defined as any object crossing the boundary, such as distillation signals, synthetic datasets, or retrieved contexts. This perspective allows us to simultaneously organize the taxonomy of transfer mechanisms and anchor a rigorous analysis of carrier-level privacy, integrity, and efficiency, a distinction largely absent in prior literature. In summary, this survey focuses exclusively on collaboration settings in which at least one party operates within a private-domain trust boundary and information exchange across this boundary is the central design concern. Consequently, we exclude fully centralized pipelines that assume unrestricted access to parameters or data, as these architectures do not necessitate a cross-boundary carrier.

Our paper is structured as follows: we first define the collaborative problem and its constraints in Section~\ref{sec:problem_definition}; then we reflect on why it is important to promote collaborative AI paradigms incorporating both large and small domain models in Section~\ref{sec:motivations}; next, we provide problem formulations and surveyed previous methodologies on enabling large and small model collaboration to solve domain tasks in Section~\ref{sec:taxonomy}. We then identify major challenges in Section~\ref{sec:challenges} and advocate for techniques that are more application-driven in Section~\ref{sec:future_directions}. 

Grounded in this perspective, Section~\ref{sec:taxonomy} organizes the literature along two axes: \textbf{training-time transfer}, classified by transfer direction (LM-to-SM and SM-to-LM) and by the form of the transmitted carrier; and \textbf{inference-time collaboration}, classified by the granularity of the exchanged carrier. Multi-party settings enter the taxonomy as a scale extension of these mechanisms rather than as a separate category.

In practice, however, domain utility cannot be pursued in isolation. LM--SM collaboration must simultaneously contend with two categories of fundamental constraints, both arising from the necessity of cross-boundary information exchange. On the privacy and security side, we identify three risk dimensions: \textbf{data privacy}, concerning the exposure of private training content and distributional properties through transmitted carriers~\cite{wang2025privacy}; \textbf{model security}, concerning the theft or reconstruction of proprietary model capabilities~\cite{tramer2016stealing, birch2023model, zhao2025survey}; and \textbf{integrity and robustness}, concerning the poisoning, backdooring, or manipulation of inbound carriers before they are consumed by the collaborator~\cite{greshake2023not, zou2025poisonedrag, chen2024agentpoison, wan2023poisoning}. Because each collaboration paradigm is characterized by the specific carrier it exchanges across the trust boundary (Section~\ref{sec:taxonomy}) and each carrier class introduces a distinct attack surface, Section~\ref{sec:challenges} maps these risk dimensions directly onto the carrier taxonomy and surveys the corresponding protection strategies. On the efficiency side, deployment cost decomposes into communication overhead for carrier transmission~\cite{cheng2021fedgems, li2024synergizing, liu2025hlora}, local computation within the resource-constrained SM environment~\cite{lin2020mcunet, alizadeh2024llm, kim2025pifi}, and remote LM invocation cost under per-token API pricing~\cite{chen2023frugalgpt, pan2024llmlingua, guo2025accrag}, with optimization strategies that likewise vary by collaboration paradigm.

These two constraint families do not act on disjoint resources. The same carrier that delivers domain utility is the channel through which privacy, security, and integrity risks and efficiency costs arise, so tightening either constraint consumes the capacity available for utility. We formalize this three-way interaction as the \textit{Collaborative Trilemma}: a Pareto frontier problem in which every LM--SM collaboration paradigm occupies a distinct operating point trading off domain utility, privacy and security assurance, and resource efficiency (Section~\ref{sec:challenges}).

Building on this diagnosis, Section~\ref{sec:future_directions} outlines a pathway for navigating the problem in practice along four complementary directions. At the measurement layer, we argue for moving beyond single-metric leaderboards toward multi-objective benchmarks that explicitly report operating points on the Pareto frontier of knowledge divergence, privacy and security constraints, and efficiency budget. At the mechanism layer, we discuss budget-aware routing and dynamic orchestration strategies that adapt carrier choice to the target operating point of a deployment~\cite{liu2022automix, ong2024routellm, ding2024hybrid}. At the deployment layer, we ground the framework in three representative real-world scenarios, urban intelligence, business intelligence, and personalized intelligence, where the interplay of utility, privacy and security assurance, and efficiency becomes most acute, providing scenario-conditioned deployment guidance. Finally, we discuss continuous adaptation under evolving tasks, data, and resource conditions (Section~\ref{sec:future_directions}).

\paragraph{\textbf{Contributions.}}
The aim of this survey is to provide a structured overview of recent advances in large and small model collaboration for private-domain adaptation, with the intent of serving as a principled resource for researchers and practitioners working toward privacy-aware and resource-efficient AI deployment. 
Our main contributions are as follows.

\begin{enumerate}
    \item We formally define the problem of large--small model collaboration for private-domain tasks. This setting captures a practical difficulty: domain knowledge is held behind local trust boundaries, while large models are often remote, proprietary, and costly to deploy locally. We use this boundary-centric formulation to clarify why collaboration must rely on controlled information carriers across the trust boundary rather than centralizing private data or exposing full model parameters (Section~\ref{sec:problem_definition}).

    \item We develop a novel taxonomy that organizes the literature by the direction, form, and granularity of the carrier exchanged across the trust boundary. The taxonomy covers training-time transfer from large models to small models (LM-to-SM) and from small models to large models (SM-to-LM), bidirectional inference-time collaboration, and multi-party extensions of these mechanisms (Section~\ref{sec:taxonomy}).

    \item We analyze the privacy, security, and efficiency implications of cross-boundary carrier exchange. On the privacy and security side, we survey carrier-specific attacks and protection techniques for data privacy, model security, and the integrity and robustness of exchanged carriers; on the efficiency side, we characterize the communication, computation, and query bottlenecks associated with each carrier. We then synthesize these analyses into a \textit{Collaborative Trilemma}, a three-way tension among domain utility, privacy and security assurance, and resource efficiency that shapes practical deployment (Section~\ref{sec:challenges}).

    \item We identify open challenges and future research directions covering multi-objective benchmarking, budget-aware collaboration protocols, deployment guidance for representative real-world scenarios, and continuous adaptation under evolving tasks, data, and resource conditions (Section~\ref{sec:future_directions}).
\end{enumerate}

\section{Problem Definition}
\label{sec:problem_definition}


As illustrated in Figure~\ref{fig:problem}, we consider a collaborative learning framework in which two complementary models operate under distinct trust and capability boundaries for a downstream task $\mathcal{T}$.

\paragraph{\textbf{Large Model (LM, $\mathcal{L}$).}}
Let $\mathcal{L}$ denote a foundation model parameterized by $\theta_{\mathcal{L}}$ and pre-trained on a broad public corpus $\mathcal{D}_{source}$~\cite{brown2020language, touvron2023llama, gao2020pile, bommasani2021opportunities}, thereby acquiring extensive knowledge across diverse tasks~\cite{ge2023openagi}. We denote its predictive function as $f_{\theta_{\mathcal{L}}}(\cdot)$. $\mathcal{L}$ serves as the primary source of general knowledge and reasoning capability~\cite{weiemergent, wei2022chain}, but its enormous size and high computational demands~\cite{kaplan2020scaling, hoffmanntraining} mean that $\mathcal{L}$ is typically hosted remotely, preventing local deployment or direct parameter access~\cite{openai2023gpt4, zhang2024cogenesis}.

\paragraph{\textbf{Small Model (SM, $\mathcal{S}$).}}
Let $\mathcal{S}$ denote a lightweight model parameterized by $\theta_{\mathcal{S}}$ and deployed directly within the private environment, instantiated either as a task-specific model~\cite{hochreiter1997long, chung2014empirical} or as a compact foundation model pre-trained on specialized domains~\cite{li2023llavamed, luo2022biogpt, lee2020biobert, wu2023bloomberggpt} or distilled from larger foundation models~\cite{javaheripi2023phi, abdin2024phi, meta2024llama3-1}. We denote its predictive function as $f_{\theta_{\mathcal{S}}}(\cdot)$. As the resident local agent, $\mathcal{S}$ is the sole component permitted to access raw private data from $\mathcal{D}_{target}$~\cite{kairouz2021advances}; its compact architecture enables efficient inference on a single local GPU, but its limited capacity often results in suboptimal performance without external guidance~\cite{zhang2024cogenesis, fancore, yi2025ecoagent}.


Given the setup above, we formalize the learning problem for private-domain adaptation task $\mathcal{T}$. We define the \textbf{domain utility functional} $\mathcal{F}$ as the performance metric evaluated strictly over $\mathcal{D}_{target}$, quantifying domain-specific effectiveness under the target distribution~\cite{wang2023self, hendrycks2020measuring, gururangan2020don, kandpal2023large}. Unlike standard machine learning, which assumes unconstrained data access and elastic compute, private-domain collaboration is governed by rigid systemic boundaries imposed by data isolation, model ownership, untrusted carrier exchange, and resource constraints~\cite{mcmahan2017communication, menghani2023efficient, yang2019federated, wang2024comprehensive}. We therefore represent the learning objective as an extended tuple:
\begin{equation}
\mathfrak{T} = \langle \mathcal{D}_{target}, \mathcal{T}, \mathcal{C} \rangle,
\end{equation}

where $\mathcal{C}$ denotes the constraint set induced by these systemic boundaries. Specifically, $\mathcal{C}$ comprises four systemic factors that distinguish private-domain tasks from general machine learning: \textbf{Knowledge Divergence}, which quantifies the distributional gap between the private domain (SM domain) and public domain (LM domain); and three hard constraints that bound the collaborative policy, namely the \textbf{Privacy Budget}, \textbf{Integrity Budget}, and the \textbf{Efficiency Budget}.

\paragraph{\textbf{Knowledge Divergence}}
Knowledge Divergence characterizes the distributional mismatch between the target 
domain $\mathcal{D}_{target}$ and the public source domain $\mathcal{D}_{source}$ used for 
general pre-training~\cite{gururangan2020don, kandpal2023large}
Formally, this discrepancy is expressed as:
\begin{equation}
\label{eq:divergence}
P(\mathcal{D}_{target}) \neq P(\mathcal{D}_{source}).
\end{equation}

where $P(\cdot)$ denotes the data distribution. This discrepancy identifies the domain gap and motivates domain adaptation.

\paragraph{\textbf{Privacy Budget}}
The privacy budget specifies the limits on information that may cross system boundaries during collaboration. More specifically, \textbf{Data Privacy} ensures that private datasets remain within their designated trust region and that any signals communicated across systems do not reveal sensitive information about individual samples or the underlying distribution~\cite{song2020information,ponomareva2023dp}. This risk is formally quantified by the data-leakage functional $M_p$, which measures the information gain of an adversary regarding the private data~\cite{Bakare2024DATAPL, carlini2021extracting, zhu2019deep}. 
\textbf{Model Security} ensures that proprietary LM parameters and intermediate states remain protected and cannot be reconstructed or inferred outside their trust region~\cite{gudibande2024false, sha2024prompt}. This risk is quantified by the model-leakage functional $M_L$, which measures the reduction in an adversary's uncertainty about proprietary model internals~\cite{liu2024vertical, tramer2016stealing, birch2023model}.  
we use $\epsilon_p$ and $\epsilon_L$ to denote their respective budget, quantifying these limits to govern both the protection of sensitive data and the confidentiality of LMs. 

\paragraph{\textbf{Integrity Budget}}
The integrity budget specifies the limits on behavioral deviations that received carriers may induce in the collaboration~\cite{greshake2023not,zou2025poisonedrag,chen2024agentpoison,wan2023poisoning}. The corresponding integrity constraint limits the extent to which cross-boundary carriers may alter intended training or inference behavior at the receiving party. Unlike $M_p$ and $M_L$, which quantify information leakage from released carriers, the integrity functional $M_r$ measures deviations induced by poisoned, backdoored, or otherwise manipulated carriers~\cite{liu2025loratk,yao2024poisonprompt}, and is bounded by the integrity budget $\epsilon_r$. 

\paragraph{\textbf{Efficiency Budget}}
The efficiency budget characterizes the computational and communication limits under
which collaboration must operate in real-world private
environments~\cite{xu2024survey, menghani2023efficient}. 
We use $M_e$ to measure the total resource consumption induced by the collaborative
policy, which typically includes three aspects: communication, local computation, and LM invocation. We use $\epsilon_e^{(comm)},\;\epsilon_e^{(comp)},\;\epsilon_e^{(query)}$ to specify the respective budgets.

\begin{itemize}
    \item \textbf{Local Computation Efficiency ($\epsilon_e^{(comp)}$):}
    Since the massive scale of $\mathcal{L}$ prohibits local deployment, $\mathcal{S}$ must bear all on-device computation within a fixed hardware envelope, which may offer only a few hundred megabytes of RAM and no dedicated GPU~\cite{lin2020mcunet, alizadeh2024llm, xu2024survey, menghani2023efficient}.

    \item \textbf{Communication Cost ($\epsilon_e^{(comm)}$):}
    The cross-silo collaboration must remain within bandwidth and latency budgets. The communication cost scales with both the volume and frequency of information exchanged across the boundary~\cite{cheng2021fedgems, li2024synergizing, liu2025hlora}.

    \item \textbf{LM Invocation Cost ($\epsilon_e^{(query)}$):}
    In API-based settings, per-token pricing accumulates rapidly under high-frequency inference~\cite{chen2023frugalgpt, pan2024llmlingua, guo2025accrag}, so collaborative policies must regulate how often and under what conditions $\mathcal{L}$ is queried.
\end{itemize}

To enable collaboration under the aforementioned constraints, we define a collaborative policy $\pi$, which governs the bidirectional information exchange:
\begin{itemize}
    \item $\mathcal{I}_{\pi}$: The information transferred from $\mathcal{L}$ to $\mathcal{S}$.
    \item $\mathcal{J}_{\pi}$: The information transferred from $\mathcal{S}$ to $\mathcal{L}$.
\end{itemize}
We define $\theta_\pi$ as the set of trainable parameters optimized during the collaboration. 
The objective is to maximize the utility $\mathcal{F}$ of the domain task $\mathcal{T}$ while strictly satisfying the boundary conditions. We formulate this as the following constrained optimization problem:
\begin{equation}
\label{eq:problem}
\max_{\theta_{\pi}} \mathcal{F}(\theta_{\mathcal{L}}, \mathcal{D}_{target}, \theta_{\mathcal{S}}, \pi), \quad  \text{s.t.} \quad
\begin{cases}
M_{p}(\mathcal{D}_{target}, \theta_{\mathcal{S}}, \mathcal{J}_{\pi}) \le \epsilon_p, \\
M_{L}(\theta_{\mathcal{L}}, \mathcal{I}_{\pi}) \le \epsilon_L, \\
M_{r}(\mathcal{I}_{\pi}, \mathcal{J}_{\pi}, \pi) \le \epsilon_r, \\
M_{e}(\mathcal{D}_{target}, \theta_{\mathcal{S}}, \pi) \preceq \boldsymbol{\epsilon}_e.
\end{cases}
\end{equation}

The components of Eq.~\eqref{eq:problem} instantiate the constraint set $\mathcal{C}$ together with the model-specific access limitations described. 
Note that real-world scenarios may involve additional constraints and objectives. Nevertheless, we aim to maintain a general framework. Additionally, the number of LMs and SMs in the collaboration 
can both be more than one, and we discuss the multi-party scenarios in Section~\ref{sec:multi-party}.

\section{Motivations}
\label{sec:motivations}

The motivation for cross-silo collaboration between large and small models is rooted in their complementary nature~\cite{wang2024comprehensive, chen2025survey}. In the framework above, this motivation corresponds to three coupled requirements: improving domain utility under knowledge divergence, preserving data privacy, model security, and integrity during collaboration, and satisfying the efficiency budget for practical deployment.
This section discusses the motivations from the perspectives of model performance, privacy and security, and system concerns, highlighting how this collaboration enables efficient AI deployment.

\subsection{Enhancing Performance}

\textbf{Enhancing SMs.} Large models exhibit strong generalization abilities across multiple tasks and domains~\cite{brown2020language, openai2023gpt4, weiemergent, wei2022chain}. 
In contrast, small models 
lack the depth and breadth of knowledge that large models possess, often showing limitations in handling complex reasoning or tasks requiring extensive background information~\cite{wang2024comprehensive}. Large models can transfer knowledge to small models through techniques such as knowledge distillation, synthetic data generation, and parameter-efficient adaptations~\cite{hinton2015distilling, ye2022zerogen, xiao2023offsite, guminillm, gholami2024gold} (Section~\ref{sec:knowledge_transfer_LM2SM}), or provide rich contextual information and world knowledge to small models at inference time~\cite{zhang2024cogenesis, xu2024small} (Section~\ref{sec:cross-silo-collaboration}), 
enabling them to perform specialized tasks more effectively without the computational cost of training from scratch. In the meantime, small models can offer valuable feedback to large models during knowledge transfer, helping LMs to adapt to domain tasks while maintaining generalization abilities and reducing the knowledge divergence between public and private domains~\cite{burns2024weak}.

\textbf{Enhancing LMs.} Training large models to achieve domain-specific capabilities requires high-quality labeled datasets from diverse fields, which is a significant challenge due to data scarcity ~\cite{villalobos2024, song2025injecting} and data privacy concerns 
~\cite{ren2024advancesopenchallengesfederated, wang2024public}. Techniques such as Federated Learning (FL) \cite{mcmahan2017communication, yang2019federated} allow private data parties to collaboratively train or fine-tune a large model without centralizing data~\cite{wang2024flora}. By establishing knowledge transfer from SMs to LMs, LMs can acquire domain knowledge without directly accessing private data held by SMs~\cite{mitchell2024emulator}. 





\subsection{Preserving Privacy and Security}



\textbf{Data Privacy.} Data privacy and confidentiality are critical concerns, leading to the enforcement of strict privacy regulations such as GDPR~\cite{regulation2016regulation}, CCPA~\cite{bonta2022california}, and HIPAA~\cite{act1996health}. Particularly in highly regulated industries like healthcare~\cite{joshi2022federated,li2023review,courtiol2019deep}, finance~\cite{chatterjee2023use,liu2023efficient}, and pharmaceuticals discovery~\cite{cordis2019machine}, valuable data resources are not yet fully exploited by the pre-training of LMs because they must remain within local trust boundaries~\cite{yang2019federated,kairouz2021advances,wang2024public}.  
A collaborative AI framework that integrates LMs and SMs empowers users to reap the benefits of advanced AI while retaining full control over their data, guaranteeing data privacy and confidentiality~\cite{kairouz2021advances, ren2024advancesopenchallengesfederated, li2024synergizing, wang2023privatelora}. Such collaboration must also account for privacy risks introduced by prompts, in-context examples, and other exchanged carriers~\cite{duan2023privacy, edemacu2025privacy}.

\textbf{Model Security and IP Protection.} Training large models involves huge investments in proprietary data and computational power, making LM parameters a valuable asset. The protection of the copyright and proprietary features of large models has become a crucial task~\cite{li2023protecting,chu2024history}. 
Adapting LMs for domain tasks in a centralized manner inevitably exposes model details to the domain party, creating opportunities for prompt stealing, model extraction, or adapter theft~\cite{sha2024prompt, zhao2025survey}. By facilitating collaboration with SMs at the domain party, large models can share specific capabilities or insights without exposing the entire underlying model or risking unauthorized duplication, reverse engineering, or adversarial exploitation~\cite{tramer2016stealing, orekondy2019knockoff, tang2024modelguard}, avoiding model attacks while maintaining strict boundaries on the intellectual property.

\textbf{Integrity and Robustness.} Cross-silo collaboration requires each party to use carriers produced outside its own trust boundary, and the receiving party may not fully control their provenance or construction~\cite{greshake2023not,debenedetti2024agentdojo}. These carriers can be adversarially manipulated before use: poisoned supervision or synthetic data corrupts downstream training, backdoored adapters introduce trigger-conditioned behavior, and injected context steers inference toward attacker-chosen outputs~\cite{wan2023poisoning,liu2025loratk,zou2025poisonedrag,chen2024agentpoison}. Trustworthy collaboration therefore must bound the behavioral influence of received carriers, complementing data privacy and model security, which constrain outbound leakage~\cite{debenedetti2024agentdojo,sun2025peftguard}.

\FloatBarrier
\begin{figure*}[!t]
\vspace{2mm}
\centering
\resizebox{1\textwidth}{!}{
    \begin{forest}
        for tree={
        grow=east,
        reversed=true,
        anchor=base west,
        parent anchor=east,
        child anchor=west,
        base=left,
        font=\large,
        rectangle,
        draw=hidden-black,
        rounded corners,
        align=left,
        minimum width=4em,
        edge path={
            \noexpand\path[\forestoption{edge}]
            (!u.parent anchor) -- +(5pt,0) |- (.child anchor)\forestoption{edge label};
        },
        edge+={darkgray, line width=1pt},
        s sep=3pt,
        inner xsep=2pt,
        inner ysep=4pt,
        line width=1.1pt,
        ver/.style={rotate=90, child anchor=north, parent anchor=south, anchor=center},
        },
        where level=0{font=\normalsize}{},
        where level=1{font=\normalsize, l sep=0.7em}{},
        where level=2{
            font=\normalsize, 
            text width=11.5em,
            l sep=0.7em
        }{},
        where level=3{
            text width=52.3em,
            font=\normalsize
        }{},
        [\qquad\ \textbf{Large-Small Model Collaboration} \qquad\ , ver
            [\qquad\ \ \textbf{Transfer from LMs to SMs} \qquad\ \ ,ver
                [\quad\ \ \textbf{Distillation-based}\ \\
                 \qquad\qquad\  (\S\,\ref{subsubsec:distillation-based-transfer-LM2SM}) \qquad\qquad
                [
                    , leaf
                    , text width=47.5em
                    , align=left
                    , content = {
                    \textbf{Single-silo:}
                    Offsite-Tuning~\cite{xiao2023offsite}{,} 
                    Emulator-Adapter~\cite{yu2023orchestration}{,}
                    AMD~\cite{han2024amd}{,}  \\
                    CRaSh~\cite{zhang2023crash}{,}
                    ScaleOT~\cite{yao2025scaleot}{,} 
                    GradOT~\cite{yao2025gradot}{,} 
                    FedPFT~\cite{peng2024fedpft}
                    \\
                    \textbf{Cross-silo:}
                    FedMD~\cite{li2019fedmd}{,} 
                    PPFL~\cite{gong2021ensemble}{,} 
                    Hard-label Stealing~\cite{sanyal2022towards}{,} \\
                    ZSKD~\cite{wang2021zero}{,} 
                    FedKD~\cite{gong2022preserving}{,}
                    IDEAL~\cite{zhang2022ideal}{,}
                    FedGMKD~\cite{zhang2024fedgmkd}{,} \\
                    EC-KD~\cite{chen2024datashunt}{,} 
                    FedTGP~\cite{zhang2024fedtgp}{,} 
                    FedKTL~\cite{zhang2024upload}{,} 
                    SD-CL~\cite{woo2025synthetic}
                    }
                    ]
                ]
                [\quad\ \ \textbf{Generation-based}\ \\
                 \qquad\qquad\  (\S\,\ref{subsubsec:generation-based-transfer-LM2SM}) \qquad\qquad
                [
                    , leaf
                    , text width=47.5em
                    , align=left
                    , content = {\textbf{Open-loop:} SuperGen~\cite{meng2022generating}{,} ZeroGen~\cite{ye2022zerogen}{,}  RetriKT~\cite{liu2023retrieval}{,} 
                    \\ 
                    GOLD~\cite{gholami2024gold}{,} SunGen~\cite{gao2023self}{,}   
                    CLTGen~\cite{latouche2024zero}{,}
                    LLKD~\cite{li2025learning}{,}
                    \\
                    PubSynth~\cite{wu2024promptpublic}{,}
                    SD-CL~\cite{woo2025synthetic}{,}
                    PRISM~\cite{ravichandran2025prism}
                    \\
                    \textbf{Closed-loop:}
                    CrossLM~\cite{deng2023mutual}{,}
                    Fusegen~\cite{zou2024fusegen}{,}
                    SoftSRV~\cite{desalvo2024no}{,} \\
                    EKD~\cite{liu2024evolving}{,}
                    ProGen~\cite{ye2022progen}{,}
                    Montessori-Instruct~\cite{li2025montessori}{,}
                    CRV~\cite{cai2025cognitive}
                    }
                    ]
                ]
                [\quad\ \ \textbf{Parameter-based}\ \\
                \qquad\qquad\ (\S\,\ref{subsubsec:parameter-based-transfer-LM2SM})\qquad\qquad
                [
                    , leaf
                    , text width=47.5em
                    , align=left
                    , content = {\textbf{Subspace-based:} MLAS-LoRA~\cite{dong2025mlas}{,} ICM-LoRA~\cite{shao2025context}{,} TrimLLM~\cite{hu2025trimllm}{,} \\
                    LoRA-Gen~\cite{xiao2025loragen}{,} PK-LoRA~\cite{zhongseeking} \\
                    \textbf{Module-based:} 4Ds~\cite{tang2024direct}{,} TAGI~\cite{liao2024instance}{,} PiFi~\cite{kim2025pifi}}
                    ]
                ]
            ]
            [\qquad\ \ \textbf{Transfer from SMs to LMs} \qquad\ \ ,ver
                [\quad\ \ \textbf{Distillation-based}\ \\
                \qquad\qquad\ (\S\,\ref{subsubsec:s2l_distillation})\qquad\qquad
                [
                    , leaf3
                    , text width=47.5em
                    , align=left
                    , content = {\textbf{Student-centered KD:} MetaDistil~\cite{zhou2022bert}{,} IKD~\cite{liu2021learning}{,}  IDEAL~\cite{zhang2022ideal}{,} \\  Meta-Pseudo-Labels~\cite{pham2021meta}{,} W2S~\cite{burns2024weak}{,} DataShunt~\cite{chen2024datashunt}{,} \\EKD~\cite{xiang2025evidential} {,}  SLMRec~\cite{xu2025slmrec}\\
                    \textbf{Backward/Reverse KD:} RKD~\cite{nasser2024reverse}{,} BiAlign~\cite{qin2025beyond}{,}  LLMD4Rec~\cite{wu2025bidirectional}{,} \\ DKDB~\cite{zhang2021dual}{,}  SHAKE~\cite{li2022shadow}{,} BiLD~\cite{li2025bild} \\
                    \textbf{Ensemble KD:} FedGEM~\cite{cheng2021fedgems}, CreamFL~\cite{yumultimodal}, FedMKT~\cite{fan2025fedmkt}
                    }
                    ]
                ]
                [\quad\ \ \textbf{Generation-based}\ \\
                \qquad\qquad\ (\S\,\ref{subsubsec:s2l_generation})\qquad\qquad
                [
                    , leaf3
                    , text width=47.5em
                    , align=left
                    , content = {\textbf{SM Generation:} Mixup~\cite{zhang2018mixup}{,} 
                    Dataset Distillation~\cite{wang2018dataset}{,} DOSFL~\cite{zhou2020distilled}{,} \\ 
                    Dataset Condensation~\cite{zhao2021dataset}{,}
                    GFL~\cite{cheng2023gfl}{,}
                    SFLD~\cite{kim2022stable}{,}
                    FedD3~\cite{song2023federated}{,}
                    \\
                    FL-GDS~\cite{li2022federated}{,} 
                    FedSD2C~\cite{zhang2024one}{,} 
                    FedC4~\cite{chen2025fedc4}{,} FedGC~\cite{yan2025federated}
                    \\
                    \textbf{LM Generation:} 
                    S3~\cite{ruida2023let}{,} 
                    CrossLM~\cite{deng2023mutual}{,} 
                    PE~\cite{lindifferentially}{,}
                    PPI~\cite{yu2024privacy}{,} 
                    \\ 
                    FedPCL~\cite{tan2022federated}{,}
                    PST~\cite{kurakin2023harnessing}{,} 
                    RetriKT~\cite{liu2023retrieval}{,} 
                    FedTGP~\cite{zhang2024fedtgp}{,} 
                    \\  
                    FedProto~\cite{tan2022fedproto}{,} 
                    DP Transformers~\cite{yue2023synthetic}{,}
                    DPSD~\cite{wu2024promptpublic}{,} 
                    DPE~\cite{xie2024differentially}{,}
                    \\ 
                    FedKTL~\cite{zhang2024upload}{,}  
                    DP Text~\cite{zhao2024dptext}{,}
                    Llamdex~\cite{wu2025model}{,} 
                    AdaDPSyn~\cite{gaodata}
                    }
                    ]
                ]
                [\quad\ \ \textbf{Parameter-based}\ \\
                \qquad\qquad\ (\S\,\ref{subsubsec:s2l_parameter})\qquad\qquad
                [
                    , leaf3
                    , text width=47.5em
                    , align=left
                    , content = {\textbf{Adapters Transfer:} FedPETuning~\cite{zhang2023fedpetuning}{,} MoLE~\cite{wumixture}{,} PET~\cite{jin2023parameter}{,} \\  Offsite-tuning~\cite{xiao2023offsite}{,} FedPA~\cite{zhang2024federated}{,} DoRA~\cite{liu2024dora}\\
                    \textbf{Tunable Prompts:} Prefix-tuning~\cite{li-liang-2021-prefix}{,} BlackVIP~\cite{BlackVIP_2023_CVPR}{,} BPO~\cite{cheng2024black}{,} \\ FedSP~\cite{dong2023tunable}{,}  FedPrompt~\cite{zhao2023fedprompt}{,} InstructZero~\cite{chen2024instructzero}}
                ]
                ]
            ]
            [\qquad\ \ \textbf{Collaborative Inference} \qquad\ \ ,ver
                [\quad \quad \ \textbf{Split Learning}\ \\
                \qquad\qquad\ (\S\,\ref{subsubsec:split_execution})\qquad\qquad
                    [
                    , leaf5
                    , text width=47.5em
                    , align=left
                    , content = {
                    SplitNN~\cite{vepakomma2018split}{,}
                    FedBERT~\cite{fedbert}{,}
                    SFPrompt~\cite{shen2023split}{,}
                    \\ 
                    PrivateLoRA~\cite{wang2023privatelora}{,}
                    SFSL~\cite{li2024introducing}{,}
                    VFL~\cite{liu2024vertical}{,} 
                    ALS~\cite{chen2024adaptive}{,}
                    \\
                    SplitLLM~\cite{mudvari2024splitllm}{,}
                    Fed-FSA~\cite{qi2025cross}
                    }
                    ]
                ]
                [\ \textbf{Collaborative Decoding}\ \\
                \qquad\qquad\ (\S\,\ref{subsubsec:collaborative_decoding})\qquad\qquad
                    [
                        , leaf5
                        , text width=47.5em
                        , align=left
                        , content = {
                        CoGenesis~\cite{zhang2024cogenesis}{,}
                        Dynamic Logits Fusion~\cite{fan2024giant}{,} 
                        \\
                        DP Fusion~\cite{thareja2026dp}{,}
                        DP-ICL~\cite{bhusal2026privacy}
                        }
                    ]
                ]
                [\; \;  \textbf{Context-Augmented}\ \\
                \qquad \ \textbf{Collaboration}\ \\
                \qquad\qquad\ (\S\,\ref{subsubsec:context_augmented_collaboration})\qquad\qquad
                    [
                        , leaf5
                        , text width=47.5em
                        , align=left
                        , content = {
                        \textbf{Retrieval Collaboration:}
                        RA-DIT~\cite{lin2023ra}{,} 
                        Self-RAG~\cite{asaiself}{,} 
                        SAGE~\cite{zeng2025mitigating}{,}
                        \\
                        RAG-LLM~\cite{NEURIPS2020_retrieval}{,}
                        RAG-end2end~\cite{RAG-end2end}{,} 
                        REPLUG~\cite{shi-etal-2024-replug}{,}
                        \\
                        RemoteRAG~\cite{cheng2025remoterag}{,}
                        Seakr~\cite{yao2025seakr}{,}
                        UniRAG~\cite{li2025unirag}
                         \\
                        \textbf{Agentic Workflow:}
                        HuggingGPT~\cite{shen2023hugginggpt}{,} 
                        Taskmatrix~\cite{liang2024taskmatrix}{,}
                        AutoGen~\cite{wu2024autogen}{,}
                        \\
                        MetaGPT~\cite{hong2023metagpt}{,}
                        ChatDev~\cite{qian2024chatdev}{,}
                        PlanGEN~\cite{parmar2025plangen}{,}
                        CORE~\cite{fancore}{,}
                        \\
                        Reflexion~\cite{shinn2023reflexion}{,} 
                        MAD~\cite{liang2024encouraging}{,}                        
                        ReConcile~\cite{chen2024reconcile}{,}
                        Smurfs~\cite{chen2025smurfs}{,}
                        \\
                        Mobile-Agent~\cite{wang2024mobile}{,} 
                        EcoAgent~\cite{yi2025ecoagent}
                        \\
                    }
                    ]
                ]
            ]
            ]
    \end{forest}%
}
\vspace{-5mm}
\caption{\textbf{A Taxonomy of Large--Small Model Collaboration on Domain Tasks.}
We organize existing approaches according to the direction of information flow and the structural interface.}
\label{fig:taxonomy-w2s-collaboration}
\end{figure*}

\subsection{Improving System Efficiency}



\textbf{Resource Efficiency.} The carbon footprint associated with LMs has become a significant global concern \cite{faiz2024llmcarbonmodelingendtoendcarbon}. One of the primary system-level motivations for collaboration is to alleviate the significant storage and computational overhead ~\cite{zhou2023comprehensive,xu2024survey,menghani2023efficient} associated with handling large models for private domains. This is crucial especially for real-time applications in resource-constrained environments, such as mobile devices or edge computing~\cite{lin2020mcunet, liu2024mobilellm, ren2024advancesopenchallengesfederated,li2024synergizing}.  
Collaborative approaches trade off this resource burden with communication, local computation, and LM invocation costs by transferring minimal knowledge carriers instead of original sensitive data to the LMs~\cite{chen2023frugalgpt, ong2024routellm}. 

\textbf{Scalability and Deployment Flexibility.} Smaller models are easier to deploy and update, simplifying maintenance and reducing operational costs~\cite{xu2024survey, bonawitz2019towards, woisetschlager2023federated}. The distributed nature of a collaborative system enables easier scalability and flexibility, because adding or removing edge devices can be done more easily to adapt to changing demands and accommodate new applications~\cite{ieee2025split, liu2025ecolora, li2025edge}.

\section{A Taxonomy of Cross-silo Collaboration between LMs and SMs}

\label{sec:taxonomy}

In recent years, there has been growing research interest in accomplishing the collaborative objective in Eq.~\ref{eq:problem} through knowledge transfer and coordination across data and model boundaries. We categorize and discuss existing methods based on their knowledge transfer mechanisms across the boundary. The first two categories address \textbf{training-time transfer}: from the LM side to the SM side (LM$\rightarrow$SM, Section~\ref{sec:knowledge_transfer_LM2SM}) and from the SM side to the LM side (SM$\rightarrow$LM,Section~\ref{sec:knowledge_transfer_SM2LM}), each further organized by the form of transferred information carrier, namely distillation-based, generation-based, and parameter-based transfer. Here, carriers are denoted by $\mathcal{I}_{\pi}$ (LM$\rightarrow$SM) and $\mathcal{J}_{\pi}$ (SM$\rightarrow$LM), consistent with Section~2. The third category addresses \textbf{inference-time collaboration} (Section~\ref{sec:cross-silo-collaboration}), where methods are also organized by the form of transferred information carrier, including intermediate activations, token-level signals, and output-level context. Additionally, \textbf{Multi-party settings} (Section~\ref{sec:multi-party}) are discussed as an extension of these mechanisms.

The distinction between $\mathcal{I}_{\pi}$ and $\mathcal{J}_{\pi}$ identifies not only carrier direction but also the boundary condition most directly engaged in Eq.~\eqref{eq:problem}: LM$\rightarrow$SM carriers $\mathcal{I}_{\pi}$ primarily expose LM-derived behavior or knowledge and are tied to $M_L$, whereas SM$\rightarrow$LM carriers $\mathcal{J}_{\pi}$ externalize private-domain signals and are tied to $M_p$. Some protocols instantiate only one direction, while closed-loop and inference-time protocols activate both across rounds.

\begin{figure}
    \centering
    \includegraphics[width=0.9\linewidth]{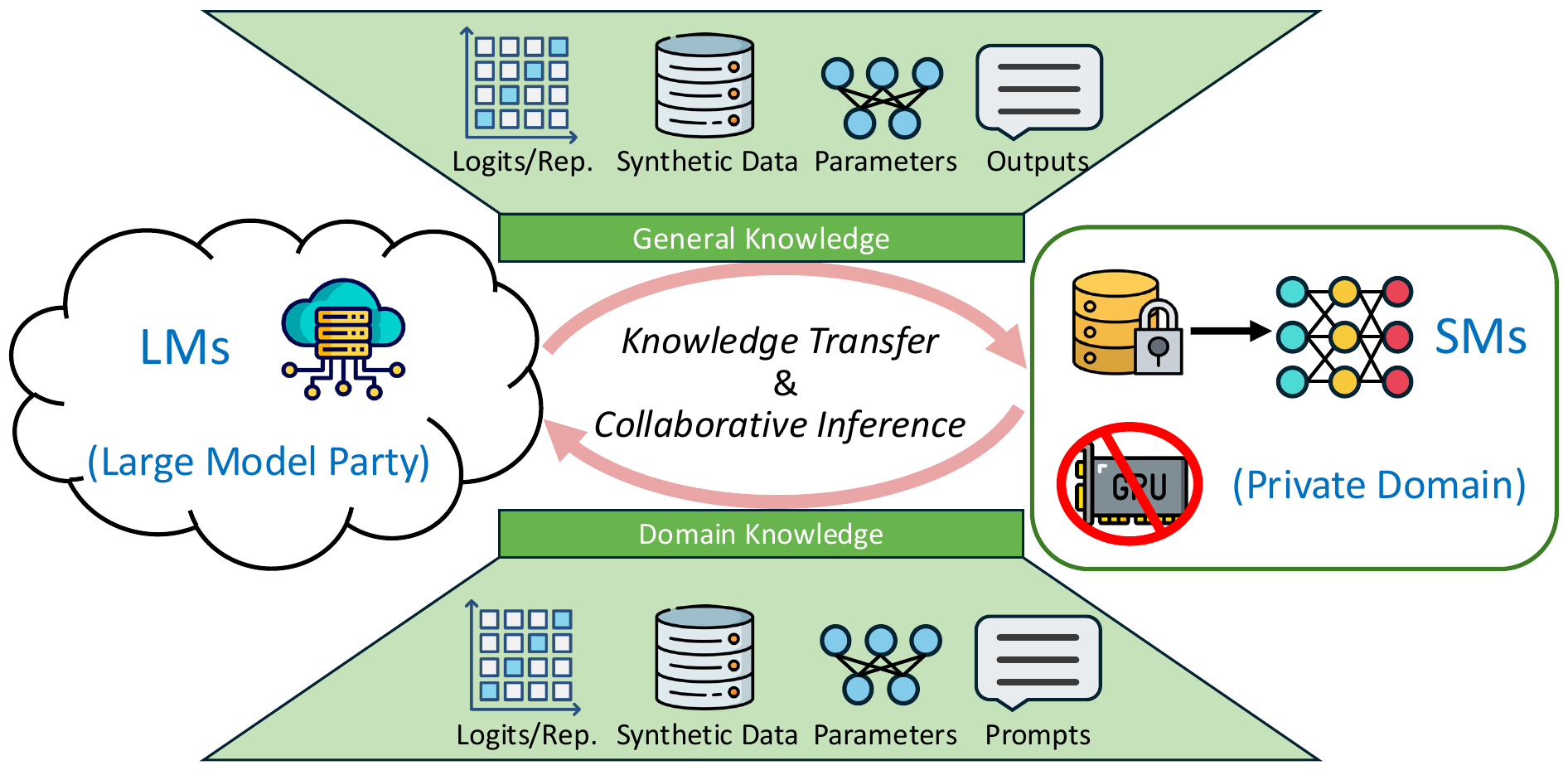}
    \caption{Overview of the cross-silo LM--SM collaboration taxonomy under the transferred-carrier view. The taxonomy distinguishes methods by carrier direction and carrier form, from training-time supervision, data, and parameter carriers to inference-time activation, token-level, and context carriers.}
    \label{fig:techniques}
\end{figure}

Table~\ref{tab:previous_techniques} summarizes the knowledge transfer mechanisms employed by this taxonomy, depicting the specific forms of knowledge carrier $\mathcal{I}_{\pi}$ and  $\mathcal{J}_{\pi}$ for each category. Figure~\ref{fig:techniques} highlights how LMs transfer general knowledge to SMs and how SMs convey domain-specific knowledge back to LMs. Figure~\ref{fig:taxonomy-w2s-collaboration} further organizes existing approaches according to our categorization. The following sections provide an overview of each paradigm.

\begin{table}[!t]
\caption{A Summary of LM--SM Collaboration Paradigms under the Transferred-Carrier View. $\theta_{\mathcal{L}}$ and $\theta_{\mathcal{S}}$ denote the parameters of large and small models, respectively, and $\theta_{\mathcal{A}}$ denotes adapters or tunable prompts.}
\centering
\newcommand{\tablehead}[1]{\begin{tabular}[c]{@{}c@{}}\rule{0pt}{2.7ex}#1\rule[-1.2ex]{0pt}{0pt}\end{tabular}}
\newcommand{\methodref}[2]{#1~(\hyperref[#2]{\S\,\ref*{#2}})}
\renewcommand{\arraystretch}{1.25}
\resizebox{\linewidth}{!}{
\begin{tabular}{c|l||c|c|c}
\hline
\tablehead{\textbf{Paradigm}} & \tablehead{\textbf{Method Category}} 
& \tablehead{\textbf{Trainable} \\ \textbf{Params}} 
& \tablehead{\textbf{Flow from LM ($\mathcal{I}_\pi$)} \\ (LM $\to$ SM)} 
& \tablehead{\textbf{Flow from SM ($\mathcal{J}_\pi$)} \\ (SM $\to$ LM)} \\
\hline\hline

\multirow[c]{3}{*}{\shortstack{Transfer\\from LMs \\to SMs}}
& \methodref{Distillation-based Transfer}{subsubsec:distillation-based-transfer-LM2SM}
& $\theta_\mathcal{S}$
& student model / logits / representations
& - \\ \cline{2-5}

& \methodref{Generation-based Transfer}{subsubsec:generation-based-transfer-LM2SM}
& $\theta_\mathcal{S}$
& synthetic data
& - \\ \cline{2-5}

& \methodref{Parameter-based Transfer}{subsubsec:parameter-based-transfer-LM2SM}
& $\theta_\mathcal{S}$
& compressed weights / adapters& - \\
\hline


\multirow[c]{3}{*}{\shortstack{Transfer\\from SMs \\to LMs}} 
& \methodref{Distillation-based Transfer}{subsubsec:s2l_distillation}
& $\theta_\mathcal{L}$
& -
& logits / representations \\ \cline{2-5}

& \methodref{Generation-based Transfer}{subsubsec:s2l_generation}
& $\theta_\mathcal{L}$
& -
& synthetic data \\ \cline{2-5}

& \methodref{Parameter-based Transfer}{subsubsec:s2l_parameter}
& $\theta_\mathcal{L}, \theta_\mathcal{A}$
& -
& adapters / tunable prompts  \\
\hline

\multirow[c]{3}{*}{\shortstack{Collaborative \\Inference}}
& \methodref{Split Learning}{subsubsec:split_execution}
& $\theta_\mathcal{L}, \theta_\mathcal{S}$ 
& intermediate outputs
& intermediate outputs \\ \cline{2-5}

& \methodref{Collaborative Decoding}{subsubsec:collaborative_decoding}
& $\theta_\mathcal{S}$
& logits / tokens 
& logits / tokens \\ \cline{2-5}

& \methodref{Context-Augmented Collaboration}{subsubsec:context_augmented_collaboration}
& $\theta_\mathcal{S}, \theta_\mathcal{L}$ 
& outputs
& prompts \\
\hline
\end{tabular}
}
\label{tab:previous_techniques}
\end{table}

\subsection{Knowledge Transfer from LMs to SMs}
\label{sec:knowledge_transfer_LM2SM}

The first feasible route is to transfer the broad knowledge and capabilities of LMs to locally deployed SMs on the SM side. Depending on the form of the knowledge transferred, this route can be accomplished in three ways: 1) Distillation-based Transfer (Section~\ref{subsubsec:distillation-based-transfer-LM2SM}), which transfers teacher-induced supervision knowledge, denoted as $\mathcal{I}_{\pi_{\mathrm{KD}}}$; 2) Generation-based Transfer (Section~\ref{subsubsec:generation-based-transfer-LM2SM}), which transfers synthetic data, denoted as $\mathcal{I}_{\pi_{\mathrm{GEN}}}$; and 3) Parameter-based Transfer (Section~\ref{subsubsec:parameter-based-transfer-LM2SM}), which transfers compact parameter knowledge, denoted as $\mathcal{I}_{\pi_{\mathrm{PAR}}}$.

These three mechanisms are also associated with different levels of LM access in practice. Parameter-based methods require white-box access to LM parameters, enabling direct manipulation of LM internal weights~\cite{sunsimple,xiao2023offsite,yao2022zeroquant,xuinitializing}. Distillation-based methods mainly transfer knowledge via logits or intermediate representations and therefore typically require grey-box access, while in strictly black-box settings they can fall back to decision-level supervision~\cite{li2019fedmd,gong2021ensemble,chen2024datashunt,zhang2022ideal,sanyal2022towards}. Generation-based methods operate through the LM generation interface, making them naturally compatible with black-box LM access~\cite{ye2022zerogen,meng2022generating,liu2023retrieval}. From the privacy perspective, the LM$\rightarrow$SM transfer  primarily impacts the model-privacy budget $M_L\!\le\!\epsilon_L$, as the boundary-crossing carrier $\mathcal{I}_{\pi}$ is directly derived from $\theta_{\mathcal{L}}$ or its behavior; the data-privacy budget $M_p$ only becomes a factor when an auxiliary SM$\rightarrow$LM feedback carrier $\mathcal{J}_{\pi}$ is introduced, such as in a bidirectional knowledge transfer loop. Note that approaches employing bidirectional transfer are discussed in both relevant subsections with different emphasis.


\subsubsection{\textbf{Distillation-based Transfer}}
\label{subsubsec:distillation-based-transfer-LM2SM}

\begin{figure}
    \centering
    \includegraphics[width=\linewidth]{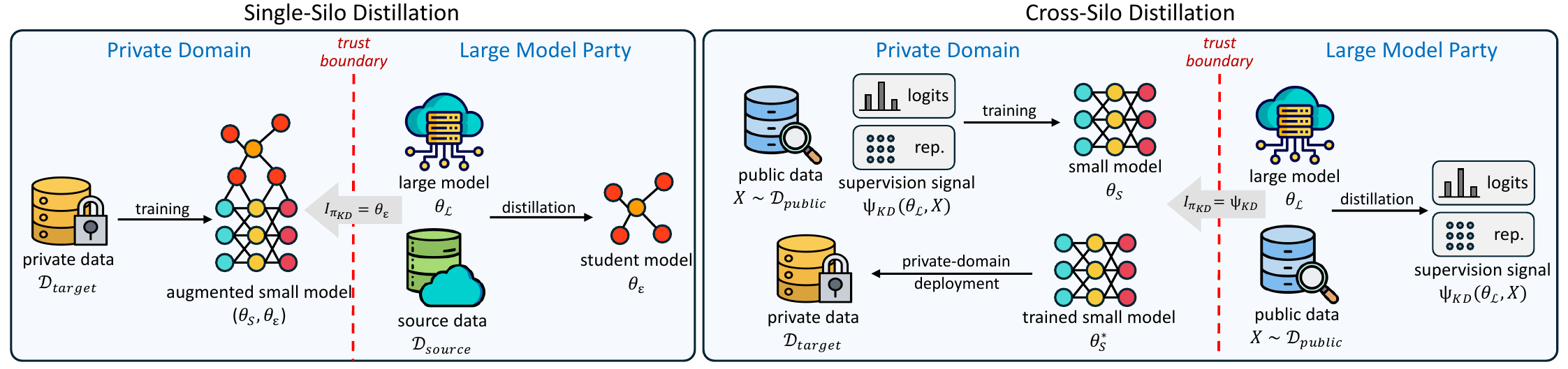}
    \caption{Distillation-based LM$\rightarrow$SM transfer.  \textit{Single-silo Distillation} transmits a compact proxy $\theta_{\mathcal{E}}$, whereas \textit{Cross-silo Distillation} transmits public-anchor supervision $\Psi_{\mathrm{KD}}(\theta_{\mathcal{L}},X)$.}
    \label{fig:LM2SM_distillation}
\end{figure}

Knowledge distillation (KD)~\cite{hinton2015distilling} is a well-established technique for transferring the knowledge and capabilities of LMs to SMs~\cite{gou2021knowledge, xu2024survey, yang2024survey, hsieh2023distilling}.
In our LM-SM collaboration setting, distillation-based transfer is distinguished from traditional Knowledge Distillation (KD) by the nature of the teacher-induced supervision signal 
$\mathcal{I}_{\pi_{\mathrm{KD}}}$ transmitted across the boundary. The defining factor is not the distillation process itself, but rather how the supervision knowledge is packaged across the boundary. Based on this packaging, we further categorize the field into two subcategories: \textit{Single-silo Distillation} and  \textit{Cross-silo Distillation}. \textit{Single-silo Distillation} transfers a compact proxy model constructed on the LM side~\cite{xiao2023offsite, peng2024fedpft}, whereas \textit{Cross-silo Distillation} transfers sample-wise supervision evaluated on the shared public dataset $\mathcal{D}_{public}$ when full LM internals are inaccessible~\cite{li2019fedmd, gong2021ensemble}.
Formally, the carrier $\mathcal{I}_{\pi_{\mathrm{KD}}}$ admits two instantiations that differ both in object type and in the data resource on which they are constructed:
\begin{equation}
\label{eq:kd_carrier_overview}
\mathcal{I}_{\pi_{\mathrm{KD}}}
=
\begin{cases}
\theta_{\mathcal{E}}, & \text{Single-silo distillation}\\[2pt]
\bigl\{\bigl(X,\,\Psi_{\mathrm{KD}}(\theta_{\mathcal{L}},X)\bigr) \,\big|\, X \in \mathcal{D}_{public}\bigr\}, & \text{Cross-silo distillation}
\end{cases},
\end{equation}
where $\theta_{\mathcal{E}}$ denotes a compact proxy distilled from $\mathcal{L}$ and $\Psi_{\mathrm{KD}}(\theta_{\mathcal{L}},X)$ is a teacher-signal extraction operator (instantiated as identity for logits, a hidden-state read-out for representations, or $\arg\max$ for hard predictions; see paragraphs below). The two cases mark a sharp division along the access regime: in the single-silo case, the LM side performs local KD based on $\mathcal{D}_{source}$ and transfers a parameter-level proxy to the SM side, whereas the cross-silo case keeps $\theta_{\mathcal{L}}$ remote and routes supervision through $\mathcal{D}_{public}$ under grey-box or black-box access. Figure~\ref{fig:LM2SM_distillation} contrasts the two cases.

\paragraph{\textbf{Single-silo Distillation}}
Single-silo Distillation packages the teacher signal as a compact proxy model constructed on the LM side and then transferred to the SM side for local adaptation.
The transferred proxy may take the form of a smaller model~\cite{guminillm, agarwal2024policy, xu2024llavadi} or a domain-specialized expert~\cite{liang2023less, liu2024ddk}, providing a stronger initialization than training from scratch~\cite{wang2024public}.
The LM side first distills its behavior into a compact proxy on the public source domain $\mathcal{D}_{source}$~\cite{xiao2023offsite, zhang2023crash, yao2025gradot, yao2025scaleot}, yielding the distilled proxy $\theta_{\mathcal{E}}$ as the transferred carrier $\mathcal{I}_{\pi_{\mathrm{KD}}}$ in Eq.~\eqref{eq:kd_carrier_overview}:
\begin{equation}
\theta_{\mathcal{E}}^{*}
=
\Psi_{\mathrm{proxy}}(\theta_{\mathcal{L}}, \mathcal{D}_{source})
=
\arg\min_{\theta_{\mathcal{E}}}
\mathbb{E}_{X\sim\mathcal{D}_{source}}
\ell_{\mathrm{KD}}\!\left(
f_{\theta_{\mathcal{E}}}(X),\, f_{\theta_{\mathcal{L}}}(X)
\right),
\end{equation}
where $\Psi_{\mathrm{proxy}}(\cdot)$ denotes a LM-side proxy construction procedure that first defines a compact proxy architecture from $\theta_{\mathcal{L}}$ through layer dropping~\cite{xiao2023offsite}, pruning~\cite{han2024amd}, or other compression mechanisms~\cite{zhang2023crash, yao2025gradot, yao2025scaleot}, and then distills the LM behavior into the resulting proxy on $\mathcal{D}_{source}$~\cite{xiao2023offsite, han2024amd}. Because $\theta_{\mathcal{E}}^{*}$ inherits its parameterization directly from $\theta_{\mathcal{L}}$ without touching $\mathcal{D}_{target}$, the proxy construction $\Psi_{\mathrm{proxy}}$ is the primary lever for satisfying the model-privacy budget $M_L\!\le\!\epsilon_L$; the data-privacy budget $M_p$ is not engaged in this direction. After the proxy is transmitted, the SM side fine-tunes the local model on $\mathcal{D}_{target}$ using the transferred proxy~\cite{xiao2023offsite, peng2024fedpft}: 
\begin{equation}
\theta_{\mathcal{S}}^{*}
=
\arg\min_{\theta_{\mathcal{S}}}
\mathbb{E}_{(X,Y)\sim\mathcal{D}_{target}}
\ell_{\mathrm{task}}\!\left(
 f_{(\theta_{\mathcal{E}}^{*},\,\theta_{\mathcal{S}})}(X)
,\,Y
\right),
\end{equation}
where $\theta_{\mathcal{S}}$ denotes the local model parameters fine-tuned on $\mathcal{D}_{target}$ using the transferred proxy.

The core challenge lies in constructing a proxy model that balances utility with compactness while preventing the leakage of proprietary LM information~\cite{xiao2023offsite,zhang2023crash,peng2024fedpft}.
A representative approach, Offsite-tuning~\cite{zhang2023crash, yu2023orchestration, xiao2023offsite} combines layer-drop and KD techniques to obtain an emulator, which is transmitted to SMs for further fine-tuning.
Building upon this paradigm, subsequent works further refine the emulator construction to enhance transfer stability and utility, including structured pruning for proxy selection~\cite{han2024amd, yao2025gradot} and dynamic layer replacement with rank compression~\cite{yao2025scaleot}.
In contrast, other approaches~\cite{peng2024fedpft, xu2025slmrec} directly obtain compact students through layer-wise compression and distillation.

\paragraph{\textbf{Cross-silo Distillation}}

Cross-silo Distillation anchors collaboration on the shared public dataset $\mathcal{D}_{public}$, on which the LM provides sample-wise supervision to the SM~\cite{li2019fedmd, gong2021ensemble, gong2022preserving, chen2024datashunt}.
The shared dataset can be either labeled or unlabeled~\cite{zhang2021parameterized, lin2020ensemble, yang2024depth}, and it serves as the public alignment anchor without exposing private-domain samples.
Concretely, the carrier is the set of sample-wise supervision pairs evaluated on $\mathcal{D}_{public}$ through the teacher-signal extraction operator $\Psi_{\mathrm{KD}}(\theta_{\mathcal{L}},X)$ introduced in Eq.~\eqref{eq:kd_carrier_overview}:
\begin{equation}
\mathcal{I}_{\pi_{\mathrm{KD}}}
= \Psi_{\mathrm{KD}}(\theta_{\mathcal{L}},X), \quad X \in \mathcal{D}_{public},
\end{equation}
where $\Psi_{\mathrm{KD}}(\theta_{\mathcal{L}},X)$ instantiates as the identity for logits~\cite{li2019fedmd, gong2021ensemble}, a hidden-state read-out for representations~\cite{chen2024datashunt}, or $\arg\max$ for hard predictions~\cite{zhang2022ideal, sanyal2022towards}, and the choice is dictated by the available LM access regime. Because the carrier touches $\theta_{\mathcal{L}}$ only through $\Psi_{\mathrm{KD}}$ on public inputs, the model-privacy exposure surface generally increases with the granularity of $\Psi_{\mathrm{KD}}$, while the data-privacy budget $M_p$ remains unengaged because $\mathcal{D}_{target}$ never exposes. The student then aligns to this supervision on the same public anchor:
\begin{equation}
\label{eq:cross_silo_kd}
\theta_{\mathcal{S}}^{*}
=
\arg\min_{\theta_{\mathcal{S}}}
\mathbb{E}_{X \sim \mathcal{D}_{public}}
 \ell_{\mathrm{KD}}\!\left(
 f_{\theta_{\mathcal{S}}}(X),\, \Psi_{\mathrm{KD}}(\theta_{\mathcal{L}},X)
 \right).
\end{equation}

When the LM provides grey-box access, this supervision can be instantiated as logits or hidden representations on $\mathcal{D}_{public}$~\cite{li2019fedmd, gong2021ensemble, chen2024datashunt}. Since no model parameters or internal architectures are exposed, this setting supports a broad range of collaboration~\cite{zhang2024fedtgp, zhang2024upload}, allowing for heterogeneous model architectures of LMs and SMs~\cite{zhang2021parameterized, fang2022robust}.

As many commercial LMs provide only black-box or API-only access, accessing soft logits or gradients becomes impossible. In such cases, the same cross-silo carrier degenerates to decision-level supervision, using LM hard predictions~\cite{zhang2022ideal, sanyal2022towards} as pseudo-labels.
While straightforward, this approach often suffers from information loss. To mitigate this, recent research explores approximating soft logits from hard decisions~\cite{wang2021zero, zhou2023bridging, woo2025synthetic} to recover richer supervision signals from the black-box interface.
However, if the teacher model was not trained on student-model domains, the effectiveness of knowledge transfer will be limited~\cite{yang2024learning, zhou2022bert, cheng2021fedgems, yumultimodal}.

\subsubsection{\textbf{Generation-based Transfer}}
\label{subsubsec:generation-based-transfer-LM2SM}
\begin{figure}
    \centering
    \includegraphics[width=\linewidth]{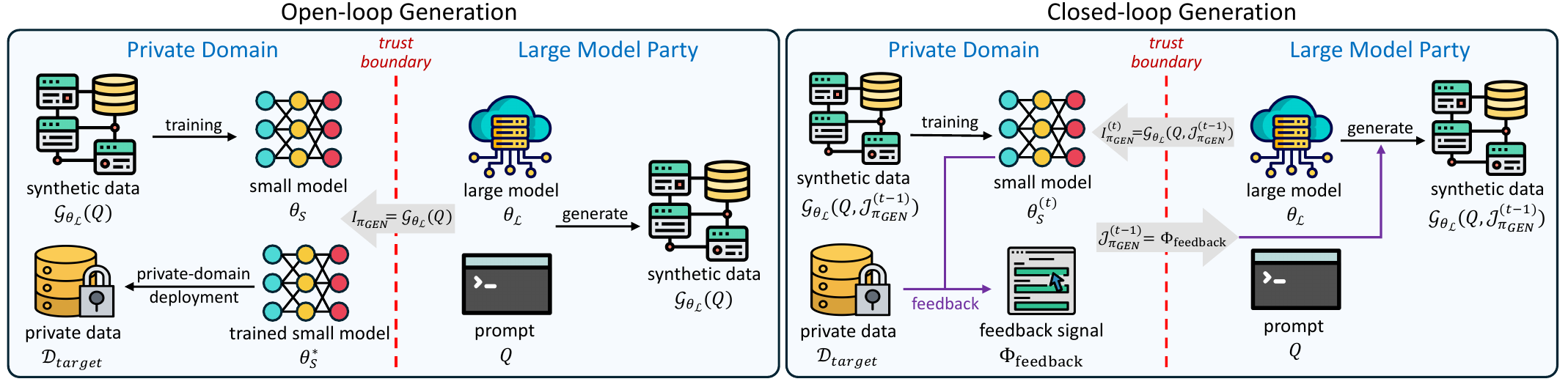}
    \caption{Generation-based LM$\rightarrow$SM transfer. The carrier is LM-generated synthetic data: \textit{Open-loop Generation} transmits $\mathcal{D}_{syn}$ once, whereas \textit{Closed-loop Generation} refreshes $\mathcal{D}_{syn}^{(t)}$ through the feedback carrier $\mathcal{J}_{\pi_{\mathrm{GEN}}}^{(t-1)}$.
    }
    \label{fig:LM2SM_generation}
\end{figure}

In generation-based transfer, the LM materializes its implicit knowledge into a synthetic dataset $\mathcal{D}_{syn}$ that serves as the transferred knowledge carrier, enabling SM adaptation without accessing private-domain data or LM internals~\cite{ye2022zerogen, meng2022generating, liu2023retrieval}. Formally, the carrier admits a unified iterative form, with the open- and closed-loop variants distinguished only by whether SM feedback is folded into the generation arguments:
\begin{equation}
\label{eq:gen_carrier_overview}
\mathcal{I}_{\pi_{\mathrm{GEN}}}^{(t)} = \mathcal{D}_{syn}^{(t)},
\qquad
\mathcal{D}_{syn}^{(t)} =
\begin{cases}
\mathcal{G}_{\theta_{\mathcal{L}}}(\mathcal{Q}), & t = 1 \;\;\text{Open-loop}\\[2pt]
\mathcal{G}_{\theta_{\mathcal{L}}}\bigl(\mathcal{Q},\, \mathcal{J}_{\pi_{\mathrm{GEN}}}^{(t-1)}\bigr), & t > 1 \;\;\text{Closed-loop}
\end{cases},
\end{equation}
where $\mathcal{G}_{\theta_{\mathcal{L}}}(\cdot)$ denotes the LM-side generation procedure, $\mathcal{Q}$ collects the prompts and task descriptors used to elicit synthesis, and $\mathcal{J}_{\pi_{\mathrm{GEN}}}^{(t-1)}$ is the SM-side feedback returned across the boundary at the previous round (formalized in the closed-loop paragraph). Because the LM is invoked only through its generation interface, this paradigm is natively compatible with black-box LMs; the carrier therefore primarily exposes the model-privacy budget $M_L$ via training-memory traces in $\mathcal{D}_{syn}^{(t)}$, while the data-privacy budget $M_p$ is engaged only in the closed-loop case through the reverse signal $\mathcal{J}_{\pi_{\mathrm{GEN}}}^{(t-1)}$. Note the open-loop case can be regarded as the degenerate instance with $\mathcal{J}_{\pi_{\mathrm{GEN}}}^{(0)}=\varnothing$. Figure~\ref{fig:LM2SM_generation} contrasts the one-shot and feedback-conditioned synthesis interfaces. 

\paragraph{\textbf{Open-loop Generation}}
In this setting, knowledge transfer is unidirectional: the SM passively learns from synthetic data generated solely from LM priors, without SM feedback. Since no private data or iterative interaction is involved, this approach is often referred to as zero-shot or offline data synthesis~\cite{zhongseeking}.
Representative approaches prompt LMs to directly generate task-specific samples or retrieval-augmented data for downstream training~\cite{ye2022zerogen, meng2022generating, liu2023retrieval}.
Recent frameworks such as PubSynth~\cite{wu2024promptpublic} and PRISM~\cite{ravichandran2025prism} further show that the LM can synthesize privacy-preserving pre-training corpora or task-and-plan demonstrations entirely from public prompts or task specifications, without any student-side feedback~\cite{wu2024promptpublic, ravichandran2025prism}.

In Eq.~\eqref{eq:gen_carrier_overview}, $\mathcal{Q}$ collects the task descriptors, prompts, and label verbalizers used for synthesis.
In open-loop generation, $\mathcal{Q}$ is typically built from public task information, such as task descriptions and labels~\cite{ye2022zerogen, meng2022generating}.  Building $\mathcal{Q}$ from private task descriptions, labels, or prompts may expose private information~\cite{chen-etal-2023-customizedText,tong2025inferdpt}.

The student is then adapted using this carrier:
\begin{equation}
\theta_{\mathcal{S}}^{*}
=
\arg\min_{\theta_{\mathcal{S}}}
\mathbb{E}_{(X,\tilde{Y})\sim \mathcal{D}_{syn}}
\ell_{\mathrm{task}}\!\left(
f_{\theta_{\mathcal{S}}}(X),\, \tilde{Y}
\right).
\end{equation}
However, relying solely on LM priors introduces risks of hallucination and domain misalignment~\cite{gao2023self, gholami2024gold, zhang2025poison}.
To mitigate this, subsequent extensions incorporate post-hoc refinement mechanisms, such as confidence-based filtering~\cite{gao2023self, li2025learning}, external grounding signals~\cite{latouche2024zero}, or out-of-distribution guidance~\cite{gholami2024gold}, thereby improving the robustness of the one-shot generation paradigm.

\paragraph{\textbf{Closed-loop Generation}}
As LMs do not have access to private data, their generated knowledge may not be fully aligned with the private domain. To further address knowledge divergence between synthetic and private data, a closed-loop generation paradigm is adopted.
Here, the SM acts as a \textit{verifier} or \textit{guide}, providing feedback signals to steer the synthesis process~\cite{ye2022progen}.
Although this process introduces a reverse signal $\mathcal{J}_{\pi_{\mathrm{GEN}}}^{(t)}$, the paradigm remains LM-to-SM transfer because the carrier ultimately used for learning is still the LM-generated synthetic dataset $\mathcal{I}_{\pi_{\mathrm{GEN}}}^{(t+1)}$.
At each round, the SM first constructs an outbound feedback signal from its current state, after which the LM regenerates the synthetic carrier conditioned on this feedback. We make this recursion explicit by introducing an SM-side feedback-extraction operator $\Phi_{\mathrm{feedback}}(\cdot)$:
\begin{equation}
\label{eq:closed_loop_gen}
\mathcal{J}_{\pi_{\mathrm{GEN}}}^{(t)}
= \Phi_{\mathrm{feedback}}\!\bigl(\theta_{\mathcal{S}}^{(t)},\, \mathcal{D}_{syn}^{(t)},\, \mathcal{D}_{target}\bigr),
\qquad
\mathcal{D}_{syn}^{(t+1)}
= \mathcal{G}_{\theta_{\mathcal{L}}}\!\bigl(\mathcal{Q},\; \mathcal{J}_{\pi_{\mathrm{GEN}}}^{(t)}\bigr),
\end{equation}
where $\Phi_{\mathrm{feedback}}(\cdot)$ extracts sample-selection, uncertainty, or error signals from the current student state and its validation behavior on the private domain, with concrete instantiations discussed below. Because $\theta_{\mathcal{S}}^{(t)}$ and the feedback rule may encode information learned from $\mathcal{D}_{target}$, the outbound feedback carrier $\mathcal{J}_{\pi_{\mathrm{GEN}}}^{(t)}$ is the channel through which the data-privacy budget $M_p(\mathcal{D}_{target}, \theta_{\mathcal{S}}, \mathcal{J}_{\pi_{\mathrm{GEN}}}^{(t)})\!\le\!\epsilon_p$ is enforced under Eq.~\eqref{eq:problem}, a constraint that does not arise in the open-loop case. The student is then retrained on the refreshed carrier:
\begin{equation}
\theta_{\mathcal{S}}^{(t+1)}
=
\arg\min_{\theta_{\mathcal{S}}}
\mathbb{E}_{(X,\tilde{Y})\sim \mathcal{D}_{syn}^{(t+1)}}
\ell_{\mathrm{task}}\!\left(
f_{\theta_{\mathcal{S}}}(X),\, \tilde{Y}
\right).
\end{equation}
The recursion iterates until the feedback-induced shift on $\mathcal{D}_{syn}^{(t)}$ falls below a threshold or the LM invocation budget $\epsilon_e^{(query)}$ is exhausted, making the iteration count $T$ a direct knob trading query efficiency for distributional alignment.
Instead of static generation, these strategies iteratively refine the data by leveraging SM-side internal states to bridge the domain gap. Key feedback mechanisms include sample-influence analysis for subset selection~\cite{ye2022progen, zou2024fusegen}, identifying domain-specific error patterns~\cite{deng2023mutual}, uncertainty-aware active learning~\cite{liu2024evolving}, or utilizing output-level behavior signals~\cite{desalvo2024no}.
Recent systems instantiate this loop by explicitly aligning generation with student influence scores or cognitive-capacity-aware critique signals, as in Montessori-Instruct and CRV~\cite{li2025montessori, cai2025cognitive}.


\subsubsection{\textbf{Parameter-based Transfer}}
\label{subsubsec:parameter-based-transfer-LM2SM}
\begin{figure}
    \centering
    \includegraphics[width=\linewidth]{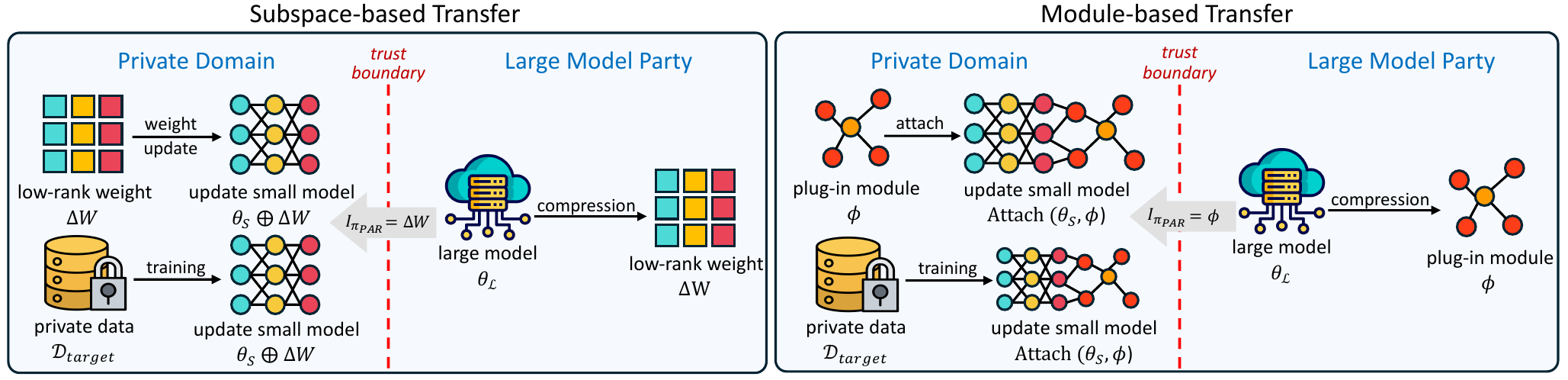}
    \caption{Parameter-based LM$\rightarrow$SM transfer. The carrier is LM-derived parameter knowledge: \textit{Subspace-based Transfer} injects $\Delta\theta$ into existing SM weights, whereas \textit{Module-based Transfer} attaches a self-contained module $\phi$.}

    \label{fig:LM2SM_parameter}
\end{figure}

Parameter-based transfer selectively transfers compact parametric knowledge from a more knowledgeable teacher model to a student model. This method is particularly relevant in scenarios where direct sharing of full models is restricted due to copyright protection or privacy concerns.
In practice, such parametric carriers are often obtained via model compression or parameter selection techniques, such as pruning~\cite{sunsimple}, layer-drop~\cite{xiao2023offsite, ling2024slimgpt}, quantization~\cite{yao2022zeroquant, huostquant}, and weight selection~\cite{xuinitializing}, and can be further combined with knowledge distillation to improve transfer effectiveness.
Depending on how task-specific knowledge is encoded and transferred, existing approaches can be broadly categorized into \textit{subspace-based transfer} and \textit{module-based transfer}~\cite{hu2021lora,tang2024direct,liao2024instance}.
Formally, the LM extracts a compact parametric carrier $\mathcal{I}_{\pi_{\mathrm{PAR}}}=\Delta\theta$ and transfers it to the SM, where it is incorporated into the student parameterization through a generic interface $\mathrm{Inject}(\cdot)$:
\begin{equation}
\label{eq:par_carrier}
\mathcal{I}_{\pi_{\mathrm{PAR}}} = \Delta\theta = \Phi_{\mathrm{PAR}}(\theta_{\mathcal{L}}),
\end{equation}
\begin{equation}
\label{eq:par_inject}
\theta_{\mathcal{S}}^{+}
= \mathrm{Inject}(\theta_{\mathcal{S}}, \Delta\theta) =
\begin{cases}
\theta_{\mathcal{S}} \oplus \Delta\theta, & \text{Subspace-based transfer}\\[2pt]
\mathrm{Attach}(\theta_{\mathcal{S}}, \phi),\; \Delta\theta := \phi, & \text{Module-based transfer}
\end{cases},
\end{equation}
where $\Phi_{\mathrm{PAR}}(\cdot)$ denotes the LM-side parameter extraction procedure, the operator $\oplus$ realizes an additive update inside existing weight matrices ($W \leftarrow W + \Delta\theta$, introducing no new parameters), and $\mathrm{Attach}(\cdot)$ augments the student with a self-contained module $\phi$ as a new architectural component. The adapted student is then optimized on the target-domain task:
\begin{equation}
\label{eq:par_adaptation}
\theta_{\mathcal{S}}^{*}
=
\arg\min_{\theta_{\mathcal{S}}^{+}}
\mathbb{E}_{(X,Y)\sim\mathcal{D}_{target}}
\ell_{\mathrm{task}}\!\left(
f_{\theta_{\mathcal{S}}^{+}}(X),\, Y
\right).
\end{equation}
Because $\Delta\theta$ is read off $\theta_{\mathcal{L}}$ directly, this paradigm presupposes white-box access, which is the strictest access requirement among the three LM$\rightarrow$SM carriers. It is consequently the route that most directly engages the model-privacy budget $M_L\!\le\!\epsilon_L$. The two paragraphs below specify the structural form of $\Delta\theta$ for each case, and the adaptation step inherits Eq.~\eqref{eq:par_adaptation}. Figure~\ref{fig:LM2SM_parameter} contrasts the two injection mechanisms.

\paragraph{\textbf{Subspace-based Transfer}}
Instead of modifying the entire parameter space, this line of work identifies task-aware low-rank subspaces or intrinsic dimensions within the LM to serve as lightweight carriers. In this case, the transferred knowledge carrier modifies the student through lightweight updates inside the existing parameter space.
Often implemented via low-rank parameterizations (e.g., LoRA~\cite{hu2021lora}), task-specific knowledge is compressed into compact update matrices.
Concretely, the subspace branch instantiates $\Delta\theta$ in Eq.~\eqref{eq:par_carrier} as a low-rank additive update to an existing weight matrix $W \in \mathbb{R}^{d\times k}$:
\begin{equation}
\label{eq:lora_subspace}
\Delta\theta = \Delta W = AB^{\top}, \qquad \mathrm{rank}(\Delta W) \le r,
\end{equation}
where $A \in \mathbb{R}^{d\times r}$, $B \in \mathbb{R}^{k\times r}$ span a task-aware low-rank subspace~\cite{hu2021lora}. The rank restriction directly contracts the carrier volume from $\mathcal{O}(d\!\cdot\!k)$ to $\mathcal{O}((d+k)\!\cdot\!r)$ with $r\!\ll\!\min(d,k)$, providing a quantitative lever on the communication sub-budget $\epsilon_e^{(comm)}$. The student applies the transferred update to its existing parameters via $\theta_{\mathcal{S}}^{+}=\theta_{\mathcal{S}}\oplus\Delta\theta$ (the subspace branch of Eq.~\eqref{eq:par_inject}) and is then optimized through Eq.~\eqref{eq:par_adaptation}.
These subspaces are extracted either through parameter importance detection to isolate task-relevant directions~\cite{dong2025mlas, zhongseeking} or via task-conditioned generation of low-rank updates~\cite{shao2025context, xiao2025loragen}, enabling the SM to inherit high-dimensional capabilities through minimal parametric shifts.

\paragraph{\textbf{Module-based Transfer}}
Module-based transfer defines the carrier as a self-contained parameter set $\phi$ instantiated as a new architectural component \textit{attached alongside} the frozen student weights, rather than folded into them. This is the key distinction from subspace-based transfer, where $\Delta\theta$ is applied as an additive update to existing weight matrices ($W \leftarrow W + \Delta\theta$) and introduces no new parameters. Concretely, existing methods encapsulate task knowledge in explicit lightweight or frozen plug-in modules~\cite{tang2024direct, kim2025pifi} or in adapter-generation rules that produce such modules conditioned on task information~\cite{liao2024instance}.
For module-based transfer, the carrier is instantiated as a self-contained parameter module, i.e., $\Delta\theta=\phi$.
Accordingly, the module branch of Eq.~\eqref{eq:par_inject} gives $\theta_{\mathcal{S}}^{+}=\mathrm{Attach}(\theta_{\mathcal{S}}, \phi)$, which augments the student with $\phi$ as a new component while leaving the original weights untouched~\cite{tang2024direct, liao2024instance}. While architectural separation allows the carrier to be independently composable, it increases the parameter footprint compared to subspace-based transfer. This imposes a more rigid constraint on the local computational budget $\epsilon_e^{(comp)}$. The adapted student is then optimized through Eq.~\eqref{eq:par_adaptation}.
These parametric carriers can be composed, injected, or generated for the student model, enabling selective transfer of task-specific behaviors with minimal parameter overhead.

\subsection{Knowledge Transfer from SMs to LMs}
\label{sec:knowledge_transfer_SM2LM}

In the reverse direction, SMs trained on the private target domain $\mathcal{D}_{target}$ act as active domain experts, encapsulating local knowledge into outbound carriers $\mathcal{J}_{\pi}$ that update or augment the LM without exposing raw private data.  This direction of knowledge transfer is also realized in three ways: 1) Distillation-based Transfer (Section~\ref{subsubsec:s2l_distillation}), which transfers supervision signals $\mathcal{J}_{\pi_{\mathrm{KD}}}$; 2) Generation-based Transfer (Section~\ref{subsubsec:s2l_generation}), which transfers synthetic data or synthesis guidance $\mathcal{J}_{\pi_{\mathrm{GEN}}}$; and 3) Parameter-based Transfer (Section~\ref{subsubsec:s2l_parameter}), which transfers compact tunable parameters $\mathcal{J}_{\pi_{\mathrm{PAR}}}$.
The three mechanisms differ in how they encode private-domain knowledge while satisfying the data-privacy budget in Eq.~\eqref{eq:problem}: anchoring on public samples, replacing data with synthetic proxies, or compressing knowledge into compact parameter modules. The precise carrier form and the corresponding $M_p$ exposure surface are formalized in each subsection below.

\subsubsection{\textbf{Distillation-based Transfer}}
\label{subsubsec:s2l_distillation}
\begin{figure}
    \centering
    \includegraphics[width=\linewidth]{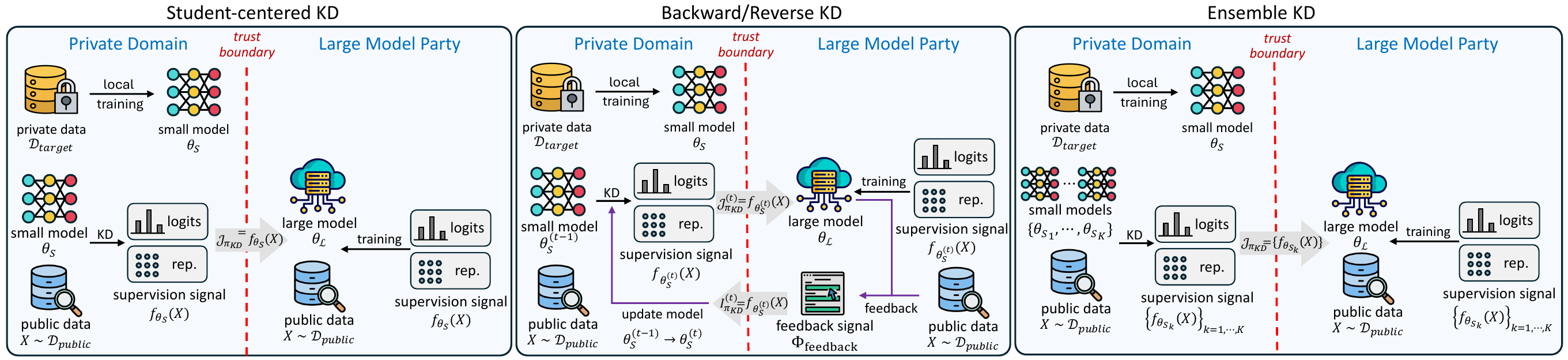}
    \caption{Distillation-based SM$\rightarrow$LM transfer. SMs are trained locally on private data, while the boundary-crossing carrier is SM-side supervision on public anchors: \textit{Student-centered KD} transmits one SM signal, \textit{Backward/Reverse KD} exchanges reciprocal KD carriers, and \textit{Ensemble KD} fuses signals from multiple SMs.}
    \label{fig:SM2LM_distillation}
\end{figure}

In distillation-based transfer, knowledge is migrated from domain-specific Small Models (SMs) to Large Models (LMs) via distilled supervision signals $\mathcal{J}_{\pi_{\mathrm{KD}}}$ rather than raw data. Based on the structural configuration of this knowledge carrier, we identify three primary variants: \textit{Student-centered KD}, which directly transmits SM knowledge to the LM ~\cite{zhou2022bert,liu2021learning,pham2021meta}; \textit{Backward/Reverse KD}, which facilitates reciprocal LM--SM co-distillation~\cite{zhang2021dual,li2022shadow,nasser2024reverse}; and \textit{Ensemble KD}, which fuses supervision from multiple SMs into a unified teaching signal~\cite{cheng2021fedgems,yumultimodal,fan2025fedmkt}.
Formally, the SM constructs an outbound distilled carrier by evaluating its parameters on shared public anchors $X \sim \mathcal{D}_{public}$:
\begin{equation}
\label{eq:s2l_kd_carrier_overview}
\mathcal{J}_{\pi_{\mathrm{KD}}}(X)
=
\Phi_{\mathrm{KD}}(\theta_{\mathcal{S}}; X)
=
\begin{cases}
f_{\theta_{\mathcal{S}}}(X), & \text{Student-centered KD}\\[2pt]
f_{\theta_{\mathcal{S}}^{(t)}}(X),\ t=1,2,\dots, & \text{Backward/reverse KD}\\[2pt]
\mathrm{Fuse}\bigl(\{f_{\theta_{\mathcal{S}_k}}(X)\}_{k=1}^{K}\bigr), & \text{Ensemble KD}
\end{cases},
\end{equation}
where $\Phi_{\mathrm{KD}}(\,\cdot\,;X)$ denotes the SM-side evaluation operator. In the backward/reverse KD, the SM model parameters are updated $\theta_{\mathcal{S}}^{(t)}$ at each iteration $t$ with the reciprocal LM-side knowledge carrier $\mathcal{I}_{\pi_{\mathrm{KD}}}^{(t)}(X)$, detailed in Sec.~\ref{subsubsec:distillation-based-transfer-LM2SM}. 
Although using public data reduce private data $\mathcal{D}_{target}$'s exposure, the transmitted knowledge carrier may still reveal private-domain information and must satisfy $M_p(\mathcal{D}_{target}, \theta_{\mathcal{S}}, \mathcal{J}_{\pi_{\mathrm{KD}}}) \le \epsilon_p$. Figure~\ref{fig:SM2LM_distillation} illustrates the three carrier configurations.

\paragraph{\textbf{Student-centered KD}}
Student-centered KD allows the teacher model to be unfrozen and trained to adapt to the specific domain of the student~\cite{zhou2022bert, liu2021learning, pham2021meta}.
In this setting, SMs serve as proxy annotators, providing supervision signals to guide the adaptation of LMs.
In this case, the carrier is constructed by a single domain-specific SM on shared public samples:
\begin{equation}
\mathcal{J}_{\pi_{\mathrm{KD}}}(X)
=
f_{\theta_{\mathcal{S}}}(X),
\qquad X \sim \mathcal{D}_{public}.
\end{equation}
The LM then adapts to this carrier as follows:
\begin{equation}
\label{eq:student_centered_kd}
\theta_{\mathcal{L}}^{*}
=
\arg\min_{\theta_{\mathcal{L}}}
\mathbb{E}_{X\sim\mathcal{D}_{public}}
\ell_{\mathrm{KD}}\!\left(
f_{\theta_{\mathcal{L}}}(X), \mathcal{J}_{\pi_{\mathrm{KD}}}(X)
\right),
\end{equation}
The carrier reveals only the student input--output mapping at public anchors $\mathcal{D}_{public}$, so $M_p$ is constrained by the degree to which an adversary can indirectly infer knowledge of $\mathcal{D}_{target}$ from SM-side signals on $\mathcal{D}_{public}$.
 The outbound carrier $\mathcal{J}_{\pi_{\mathrm{KD}}}(X)$ has at least two practical instantiations. Under black-box access, the SM transmits only a decision-level label carrier~\cite{zhang2022ideal}:
\begin{equation}
\mathcal{J}_{\pi_{\mathrm{KD}}}^{\mathrm{label}}(X)
=
\hat{Y}_{\mathcal{S}}
=
\arg\max f_{\theta_{\mathcal{S}}}(X),
\qquad
\theta_{\mathcal{L}}^{*}
=
\arg\min_{\theta_{\mathcal{L}}}
\mathbb{E}_{X\sim\mathcal{D}_{public}}
\ell_{\mathrm{CE}}\!\left(
f_{\theta_{\mathcal{L}}}(X),\hat{Y}_{\mathcal{S}}
\right).
\end{equation}
Under grey-box access, the SM instead transmits an uncertainty-aware carrier $\alpha_{\mathcal{S}}(X)$ that includes its predicted class distribution and an uncertainty estimate~\cite{xiang2025evidential,guo2024lud}:
\begin{equation}
\mathcal{J}_{\pi_{\mathrm{KD}}}^{\mathrm{uncertainty}}(X)
=
\alpha_{\mathcal{S}}(X),
\qquad
\theta_{\mathcal{L}}^{*}
=
\arg\min_{\theta_{\mathcal{L}}}
\mathbb{E}_{X\sim\mathcal{D}_{public}}
\ell_{\mathrm{KD}}\!\left(
f_{\theta_{\mathcal{L}}}(X),\alpha_{\mathcal{S}}(X)
\right).
\end{equation}
This aligns with weak-to-strong generalization, where decisions from weaker models are used to supervise stronger models~\cite{burns2024weak, chen2024datashunt}.

\paragraph{\textbf{Backward/Reverse KD}}
Backward or Reverse KD~\cite{zhang2021dual, li2022shadow, nasser2024reverse} inverts the distillation flow, utilizing shared logits and representations to facilitate knowledge transfer from the student back to the teacher, often enabling a bidirectional loop~\cite{zhang2021dual, li2022shadow}.
Unlike student-centered KD, this paradigm extends the one-way transfer above into bidirectional or online co-distillation, where the LM and SM construct reciprocal carriers on shared samples and update each other jointly. The step-$t$ carrier construction is:
\begin{equation}
\mathcal{J}_{\pi_{\mathrm{KD}}}^{(t)}(X)
=
f_{\theta_{\mathcal{S}}^{(t)}}(X),
\qquad
\mathcal{I}_{\pi_{\mathrm{KD}}}^{(t)}(X)
=
f_{\theta_{\mathcal{L}}^{(t)}}(X).
\end{equation}
The coupled SM adaptation then follows:
\begin{equation}
\left\{
\begin{aligned}
\theta_{\mathcal{L}}^{(t+1)}
&=
\arg\min_{\theta_{\mathcal{L}}}
\mathbb{E}_{X\sim\mathcal{D}_{public}}
\ell_{\mathrm{KD}}\!\left(
f_{\theta_{\mathcal{L}}}(X), \mathcal{J}_{\pi_{\mathrm{KD}}}^{(t)}(X)
\right), \\
\theta_{\mathcal{S}}^{(t+1)}
&=
\arg\min_{\theta_{\mathcal{S}}}
\mathbb{E}_{X\sim\mathcal{D}_{public}}
\ell_{\mathrm{KD}}\!\left(
f_{\theta_{\mathcal{S}}}(X), \mathcal{I}_{\pi_{\mathrm{KD}}}^{(t)}(X)
\right),
\end{aligned}
\right.,
\end{equation}
where $\mathcal{J}_{\pi_{\mathrm{KD}}}^{(t)}(X)$ denotes the student-side supervision used to update the LM at iteration $t$, $\mathcal{I}_{\pi_{\mathrm{KD}}}^{(t)}(X)$ denotes the reciprocal supervision signal returned by the LM to the student (in the sense of the LM-side carrier introduced in Sec.~\ref{subsubsec:distillation-based-transfer-LM2SM}). Therefore student-centered KD can be viewed as a special case of this bidirectional formulation~\cite{zhang2021dual, li2022shadow, nasser2024reverse}.
Since both carrier directions are exchanged repeatedly on $\mathcal{D}_{public}$, the relevant privacy accounting is cumulative over the interaction trajectory: the SM$\rightarrow$LM exposure accumulates over the transmitted carrier sequence $\{\mathcal{J}_{\pi_{\mathrm{KD}}}^{(t)}\}_{t=1}^{T}$, and the reciprocal LM$\rightarrow$SM exposure accumulates over $\{\mathcal{I}_{\pi_{\mathrm{KD}}}^{(t)}\}_{t=1}^{T}$, rather than arising from a single-iteration transmission.
Thus, the cumulative budgets must bound the summed per-round exposures, i.e., $\sum_{t=1}^{T} M_p(\mathcal{D}_{target},\theta_{\mathcal{S}}^{(t)},\mathcal{J}_{\pi_{\mathrm{KD}}}^{(t)})\le\epsilon_p$ for SM$\rightarrow$LM carriers and $\sum_{t=1}^{T} M_L(\theta_{\mathcal{L}}^{(t)},\mathcal{I}_{\pi_{\mathrm{KD}}}^{(t)})\le\epsilon_L$ for LM$\rightarrow$SM carriers.

This makes privacy budgeting more challenging than in the one-shot student-centered setting.
Moving beyond simple prediction consistency, recent work explores diverse interaction mechanisms, including inference-time in-context representation alignment~\cite{qin2025beyond}, task-specific semantic exchange~\cite{wu2025bidirectional}, and explicit bidirectional loss designs for stable optimization~\cite{li2025bild}.

\paragraph{\textbf{Ensemble KD}} To overcome the limited capability of any single small model, \textit{Ensemble Knowledge Distillation} leverages multiple SMs as complementary domain experts~\cite{cheng2021fedgems,yumultimodal,fan2025fedmkt}. Here, the knowledge carrier is no longer provided by a single SM but is instead fused from multiple local experts, and the key challenge is to combine heterogeneous and potentially biased supervision into a stable teaching signal for the LM. Here we formulate the fusion as : 
\begin{equation}
\mathcal{J}_{\pi_{\mathrm{KD}}}^{(k)}(X) = f_{\theta_{\mathcal{S}_k}}(X),\ k=1,\dots,K,
\qquad
\bar{\mathcal{J}}_{\pi_{\mathrm{KD}}}(X) = \mathrm{Fuse}\!\left(\{\mathcal{J}_{\pi_{\mathrm{KD}}}^{(k)}(X)\}_{k=1}^{K}\right).
\end{equation}
The LM then adapts to this fused carrier through:
\begin{equation}
\label{eq:ensemble_kd}
\theta_{\mathcal{L}}^{*}
=
\arg\min_{\theta_{\mathcal{L}}}
\mathbb{E}_{X\sim\mathcal{D}_{public}}
\ell_{\mathrm{KD}}\!\left(
f_{\theta_{\mathcal{L}}}(X),
\bar{\mathcal{J}}_{\pi_{\mathrm{KD}}}(X)
\right),
\end{equation}
where $\mathcal{J}_{\pi_{\mathrm{KD}}}^{(k)}(X)$ denotes the supervision signal from the $k$-th small model, $K$ is the number of participating local experts. $\bar{\mathcal{J}}_{\pi_{\mathrm{KD}}}(X)$ denotes the fused teaching signal. Eq.~\eqref{eq:ensemble_fuse} below concretizes the abstract $\mathrm{Fuse}(\cdot)$ operator as a weighted aggregation, the canonical instantiation in current work:
\begin{equation}
\label{eq:ensemble_fuse}
\bar{\mathcal{J}}_{\pi_{\mathrm{KD}}}(X) = \sum_{k=1}^{K} w_k \cdot \mathcal{J}_{\pi_{\mathrm{KD}}}^{(k)}(X),
\end{equation}
where $w_k$ denotes the weight assigned to supervision from the $k$-th model.
The effectiveness of ensemble KD depends on two interrelated factors: the \textit{diversity} of supervision signals and the \textit{quality} of the fusion rule.
Diversity arises naturally from heterogeneous private domains $\{\mathcal{D}_{target}^{(k)}\}$: each SM encodes complementary distributional information, so the fused carrier $\bar{\mathcal{J}}_{\pi_{\mathrm{KD}}}(X)$ can in principle convey richer domain knowledge than any single-SM signal.
However, this same heterogeneity also introduces signal conflicts and quality disparities: SMs trained on non-IID private data may produce supervision that is mutually contradictory or domain-biased, and naive averaging in Eq.~\eqref{eq:ensemble_fuse} can degrade LM adaptation rather than improve it~\cite{cheng2021fedgems,yumultimodal,fan2025fedmkt}.

Existing approaches address this tension through three representative instantiations of $w_k$.
\textbf{(i)} \textit{Selective filtering}~\cite{cheng2021fedgems} discards low-quality or outlier supervision before aggregation, ensuring that only sufficiently reliable carriers contribute to the fused signal.
\textbf{(ii)} \textit{Confidence-based weighting}~\cite{yumultimodal} assigns higher weight to SMs that express high prediction confidence on the public sample $X$, reflecting their domain reliability on the shared evaluation anchor.
\textbf{(iii)} \textit{Alignment-aware mutual transfer}~\cite{fan2025fedmkt} dynamically adjusts $w_k$ based on cross-model representation alignment, reducing the influence of domain-shifted or structurally incompatible supervision signals.
A persistent open challenge is that computing reliable fusion weights without access to a shared labeled validation set is difficult, since the private nature of $\mathcal{D}_{target}^{(k)}$ prevents direct quality assessment at the central LM side.

Because each SM-side knowledge carrier is independently transmitted across $K$ private boundaries, the fusion rule also determines how privacy exposure is allocated across clients; under compositional leakage measures such as Differential Privacy, the aggregate exposure can be upper-bounded by the accumulated per-client transmissions.

The privacy footprint also escalates across the three variants: Student-centered KD transmits a one-shot snapshot of a single SM on $\mathcal{D}_{public}$; backward/reverse KD additionally exposes the LM-side reciprocal carrier and thus impacts both $M_p$ and $M_L$; ensemble KD aggregates $K$ snapshots and relies on the fusion rule to simultaneously control signal quality and allocate per-client privacy budget.

\subsubsection{\textbf{Generation-based Transfer}}
\label{subsubsec:s2l_generation}
In generation-based transfer, SMs transfer private-domain knowledge to LMs through synthetic data or synthesis guidance rather than raw private data. The transmitted carrier is used to adapt or augment the LM~\cite{kurakin2023harnessing, yu2024privacy}.
A key distinction among existing methods lies in the \textit{source of synthesis}: \textit{SM Generation} constructs the synthetic carrier on the private side~\cite{li2022federated,cheng2023gfl,kim2022stable,song2023federated,zhou2020distilled}, whereas \textit{LM Generation} transmits compact SM guidance that steers LM-side synthesis~\cite{lindifferentially,xie2024differentially,ruida2023let,zhang2024upload,deng2023mutual}.
Formally, the boundary-crossing carrier $\mathcal{J}_{\pi_{\mathrm{GEN}}}$ takes one of two forms:
\begin{equation}
\label{eq:s2l_gen_carrier_overview}
\mathcal{J}_{\pi_{\mathrm{GEN}}} =
\begin{cases}
\mathcal{D}_{syn}\ \text{or}\ g_{\mathcal{S}}, & \text{SM generation}\\[2pt]
C_{\mathrm{priv}}=\Phi_{\mathrm{guide}}(\theta_{\mathcal{S}},\mathcal{D}_{target}), & \text{LM generation}
\end{cases},
\end{equation}
In SM Generation, the transferred carrier is either the synthesized dataset $\mathcal{D}_{syn}$ or the private-side generator model $g_{\mathcal{S}}$ that produces or supports it~\cite{li2022federated,cheng2023gfl,zhou2020distilled,song2023federated,kim2022stable}, with $g_{\mathcal{S}}$ trained from $\mathcal{D}_{target}$. In LM Generation, the transferred carrier is the compact guidance signal $C_{\mathrm{priv}}=\Phi_{\mathrm{guide}}(\theta_{\mathcal{S}},\mathcal{D}_{target})$, which steers LM-side synthesis or supports direct in-context use.

The synthetic dataset consumed by the LM is constructed differently in the two forms:
\begin{equation}
\label{eq:s2l_gen_construction}
\begin{cases}
\mathcal{D}_{syn}=g_{\mathcal{S}}(\mathcal{D}_{target}), & \text{SM generation}\\[2pt]
\mathcal{D}_{syn}^{(t)}=\mathcal{G}_{\theta_{\mathcal{L}}}^{(t)}(\mathcal{Q}_{\mathrm{pub}}, C_{\mathrm{priv}}),\quad t=1,\ldots,T, & \text{LM generation}
\end{cases},
\end{equation}
where $g_{\mathcal{S}}(\mathcal{D}_{target})$ denotes the synthetic dataset that the SM-side generator derives from $\mathcal{D}_{target}$, $\mathcal{G}_{\theta_{\mathcal{L}}}^{(t)}$ is the LM-side synthesis procedure at round $t$ introduced in Section~\ref{subsubsec:generation-based-transfer-LM2SM}, and $\mathcal{Q}_{\mathrm{pub}}$ collects the public prompts or task descriptors used by the LM.

\begin{figure}
    \centering
    \includegraphics[width=\linewidth]{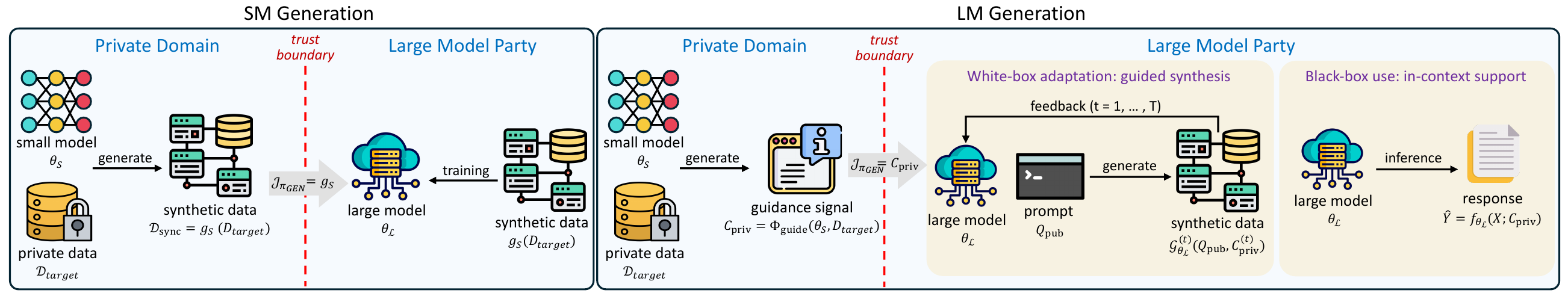}
    \caption{Generation-based SM$\rightarrow$LM transfer. The carrier differs by synthesis source: \textit{SM Generation} transmits a private-side generator model $g_{\mathcal{S}}$ or its synthesized dataset $\mathcal{D}_{syn}$. \textit{LM Generation} transmits compact guidance $C_{\mathrm{priv}}$ for iterative LM-side synthesis under white-box adaptation or direct in-context support under black-box use.}
    \label{fig:SM2LM_generation}
\end{figure}

Therefore, SM Generation and LM Generation differ in both synthesis location and carrier form. SM Generation transmits a private-side generator model or its synthesized dataset, whereas LM Generation transmits compact guidance for LM-side synthesis or direct in-context support.
Figure~\ref{fig:SM2LM_generation} contrasts the two synthesis processes.

\paragraph{\textbf{SM Generation}}
In SM Generation, the private side constructs the transferred carrier locally, matching the SM-generation cases in Eqs.~\eqref{eq:s2l_gen_carrier_overview} and \eqref{eq:s2l_gen_construction}. Representative instantiations include local generative synthesis~\cite{li2022federated,cheng2023gfl}, dataset distillation or condensation~\cite{song2023federated,zhou2020distilled,wang2018dataset,zhao2021dataset}, mixup-based augmentation~\cite{zhang2018mixup}, and privacy-preserving text generation~\cite{kurakin2023harnessing,yu2024privacy}. These methods either transmit the synthesized dataset $\mathcal{D}_{syn}$ directly or transmit a private-side generator model $g_{\mathcal{S}}$ that supports server-side synthesis or stabilization~\cite{kim2022stable}. Both carriers are derived from $\mathcal{D}_{target}$ and therefore expose private data under $M_p$, which DP-based generators bound with explicit privacy guarantees~\cite{xie2024differentially}.
When the synthesized dataset is transmitted to the LM, the LM can be adapted by optimizing on the received synthetic examples:
\begin{equation}
\label{eq:gen_s2l}
\theta_{\mathcal{L}}^{*}
=
\arg\min_{\theta_{\mathcal{L}}}
\mathbb{E}_{(X,\tilde{Y})\sim\mathcal{D}_{syn}}
 \ell_{\mathrm{task}}\!\left(
f_{\theta_{\mathcal{L}}}(X),\, \tilde{Y}
\right),
\end{equation}
where $\tilde{Y}$ denotes the synthetic label associated with $X$.
Recent variants further compress the transmitted carrier through one-shot synthetic distillates~\cite{zhang2024one} or graph-domain condensation that preserves relational structure in synthetic graphs~\cite{yan2025federated}, reducing the communication cost of carrier transfer.

\paragraph{\textbf{LM Generation}}
In LM Generation, compact guidance $\mathcal{J}_{\pi_{\mathrm{GEN}}}=C_{\mathrm{priv}}$ steers LM-side synthesis, matching the LM-generation cases in Eqs.~\eqref{eq:s2l_gen_carrier_overview} and \eqref{eq:s2l_gen_construction}; the same guidance can also be consumed directly as in-context support. Under white-box adaptation, the LM may refine $\mathcal{D}_{syn}^{(t)}$ across multiple synthesis rounds using feedback from previously generated samples or task-specific guidance, and the resulting dataset can be used in the adaptation objective in Eq.~\eqref{eq:gen_s2l}~\cite{deng2023mutual,zou2024fusegen}. Under black-box use, the returned guidance signal is consumed directly as in-context support while $\theta_{\mathcal{L}}$ remains frozen~\cite{ruida2023let,lindifferentially,xie2024differentially}:
\begin{equation}
\hat{Y}
=
f_{\theta_{\mathcal{L}}}\!\left(X;\, C_{\mathrm{priv}}\right).
\end{equation}
The guidance can take various forms, including in-context samples under privacy protection~\cite{lindifferentially,xie2024differentially,ruida2023let}, class-level prototypes~\cite{zhang2024upload,tan2022fedproto,tan2022federated,zhang2024fedtgp}, tunable prompts~\cite{liu2023retrieval}, and reward signals~\cite{deng2023mutual}. These forms differ in how directly they expose private data: instance-level guidance such as in-context samples reveals more than aggregated carriers such as prototypes or scalar rewards. 
Beyond their privacy profile, these guidance forms also narrow the distribution gap between public-prompt synthesis and the private domain, for example through adaptive DP prompts~\cite{gaodata} or constraint-based signals~\cite{wu2024promptpublic}.

\subsubsection{\textbf{Parameter-based Transfer}}
\label{subsubsec:s2l_parameter}
\begin{figure}
    \centering
    \includegraphics[width=\linewidth]{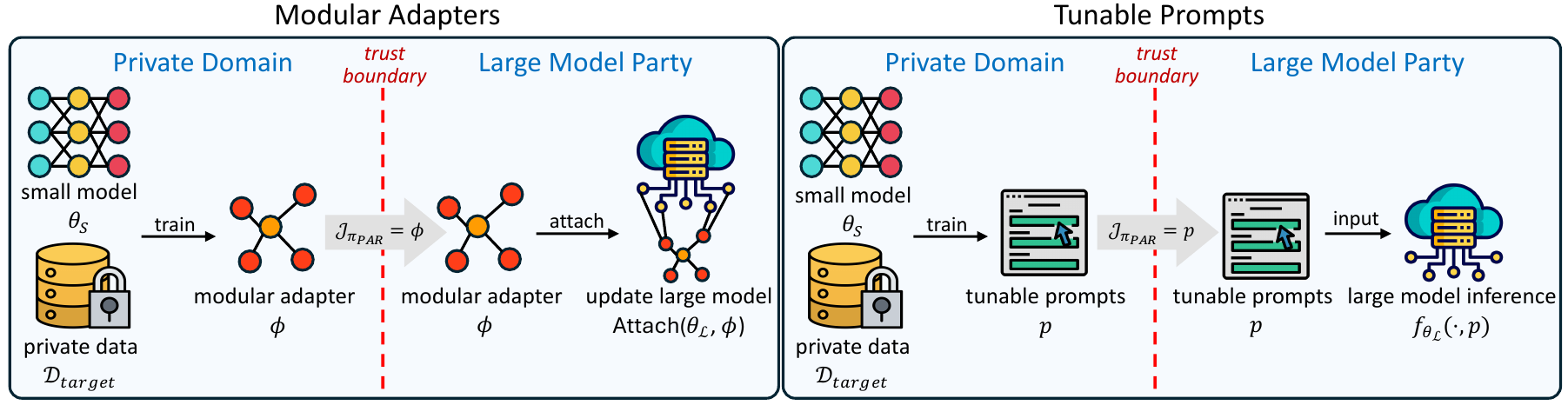}
    \caption{Parameter-based SM$\rightarrow$LM transfer. The carrier is private-trained parameter knowledge: \textit{Modular Adapters} attach a module $\phi$ inside the LM, whereas \textit{Tunable Prompts} expose a prompt vector $\mathbf{p}$ through the input interface.}

    \label{fig:SM2LM_parameter}
\end{figure}

In parameter-based transfer, SMs optimize compact parameter carriers $\mathcal{J}_{\pi_{\mathrm{PAR}}}$ on the private domain $\mathcal{D}_{target}$ and transfer them to the LM, enabling domain adaptation without modifying its pre-trained weights~\cite{zhang2023fedpetuning}. Two carrier types are used: explicit adapter modules~\cite{xiao2023offsite,jin2023parameter,zhang2024federated} and prompt-level vectors~\cite{li-liang-2021-prefix,dong2023tunable,BlackVIP_2023_CVPR,zhao2023fedprompt}.
The applicable carrier form depends on the LM access setting. Modular adapters require white-box access, since they are inserted into internal LM layers. Tunable prompts work under API-only access, since they operate through the input interface and leave the LM frozen.
Formally, the SM learns a compact outbound carrier $\mathcal{J}_{\pi_{\mathrm{PAR}}}$ from the private domain and supplies it to the LM, but the way the carrier is applied differs sharply across the two variants:
\begin{equation}
\label{eq:s2l_par_overview}
\begin{aligned}
\mathcal{J}_{\pi_{\mathrm{PAR}}}
&=
\Phi_{\mathrm{PAR}}(\theta_{\mathcal{S}}, \mathcal{D}_{target})
=
\begin{cases}
\phi, & \text{Modular adapters}, \\[2pt]
\mathbf{p}, & \text{Tunable prompts}
\end{cases},\\
f_{\theta_{\mathcal{L}}}^{+}(\cdot)
&=
\begin{cases}
f_{\mathrm{Attach}(\theta_{\mathcal{L}},\,\phi)}(\cdot), & \text{Modular adapters}, \\[2pt]
f_{\theta_{\mathcal{L}}}(\cdot;\mathbf{p}),\ \theta_{\mathcal{L}}\ \text{frozen}, & \text{Tunable prompts}
\end{cases}.
\end{aligned}
\end{equation}
where $\Phi_{\mathrm{PAR}}(\cdot)$ denotes the procedure of learning the SM-side knowledge carrier from $\mathcal{D}_{target}$.
The compactness of $\phi$ and $\mathbf{p}$ also yields a tight communication budget $\epsilon_e^{(comm)}$, making this paradigm a favorable choice when both privacy and communication budget are simultaneously constrained. Figure~\ref{fig:SM2LM_parameter} contrasts the two carrier forms.

\paragraph{\textbf{Modular Adapters}}
In Modular Adapters, the outbound carrier $\mathcal{J}_{\pi_{\mathrm{PAR}}} = \phi$ is a self-contained parameter module trained on $\mathcal{D}_{target}$ and inserted into the LM without modifying its pre-trained weights. Representative implementations include task-specific adapter modules~\cite{xiao2023offsite, jin2023parameter} and personalized LoRA variants in distributed settings~\cite{zhang2024federated, zhang2023fedpetuning}. To improve utility under limited parameter budgets, mixtures of LoRA experts increase representational capacity~\cite{wumixture}, while weight-decomposed adapter formulations enhance stability and efficiency of parameter injection~\cite{liu2024dora}.
A naive learning step of the knowledge carrier is:
\begin{equation}
\label{eq:adapter_local}
\phi
=
\Phi_{\mathrm{adapter}}(\theta_{\mathcal{S}}, \mathcal{D}_{target}),
\end{equation}
where $\Phi_{\mathrm{adapter}}(\cdot)$ denotes the SM-side procedure that compiles target-domain knowledge into an LM-compatible adapter module, so that $\mathcal{J}_{\pi_{\mathrm{PAR}}}=\phi$. The carrier is then applied via the first case of Eq.~\eqref{eq:s2l_par_overview}, attaching $\phi$ into the LM weight space.

\paragraph{\textbf{Tunable Prompts}}
Tunable prompts~\cite{li-liang-2021-prefix} are prefixed parameters that transmit domain knowledge from SMs to LMs, instantiated as knowledge messengers under both white-box and black-box access settings~\cite{dong2023tunable,BlackVIP_2023_CVPR}, with derivative-free optimization frameworks enabling alignment under strictly black-box LM access without gradient access~\cite{cheng2024black, chen2024instructzero}. In federated settings, this property additionally offers a communication-efficient and privacy-preserving transfer channel~\cite{zhao2023fedprompt}.
Here, the outbound carrier $\mathcal{J}_{\pi_{\mathrm{PAR}}}=\mathbf{p}$ is a prompt-level  vector learned on the private side:
\begin{equation}
\label{eq:prompt_local}
\mathbf{p}
=
\Phi_{\mathrm{prompt}}(\theta_{\mathcal{S}}, \mathcal{D}_{target}),
\end{equation}
where $\Phi_{\mathrm{prompt}}(\cdot)$ compiles target-domain knowledge into a prompt carrier. The carrier is then applied via the second case of Eq.~\eqref{eq:s2l_par_overview}: $\theta_{\mathcal{L}}$ is held frozen and $\mathbf{p}$ is prepended at the input interface, so that downstream task solving reduces to inference through $f_{\theta_{\mathcal{L}}}(\cdot;\mathbf{p})$ without further weight updates.

\subsection{Cross-silo Collaborative Inference}
\label{sec:cross-silo-collaboration}
While the aforementioned works result in either fine-tuned LMs or enhanced SMs, which are able to perform inference independently once trained, this section discusses collaborative approaches requiring both LMs and SMs at inference time. One common characteristic of the approaches in this section is that they all require LM-SM knowledge transfer and collaboration at inference time (Table \ref{tab:previous_techniques}). Unlike the training-time transfer mechanisms discussed in the preceding sections, cross-silo collaborative inference requires \textit{both} LMs and SMs to collaborate at inference time as well, with task-relevant knowledge exchanged through SM-side outbound carriers $\mathcal{J}_{\pi}$ and, when applicable, LM-side returned carriers $\mathcal{I}_{\pi}$ at each interaction step. Following the preceding sections, this section also categorizes these approaches by the \textbf{granularity of the transferred carriers} (Table~\ref{tab:previous_techniques}):
1) \textbf{Split Learning} (Section~\ref{subsubsec:split_execution}), which exchanges intermediate hidden-state carriers $\mathcal{J}_{\pi}^{\mathrm{hidden}}$ between a local head and a remote tail~\cite{vepakomma2018split,li2024introducing,gu2025vflair};
2) \textbf{Collaborative Decoding} (Section~\ref{subsubsec:collaborative_decoding}), which exchanges step-wise token-level carriers $\mathcal{J}_{\pi}^{(t)}$ during auto-regressive generation~\cite{zhang2024fast,leviathan2023fast,li2023contrastive,liu2024tuning};
3) \textbf{Context-Augmented Collaboration} (Section~\ref{subsubsec:context_augmented_collaboration}), which exchanges output-level contextual request and response carriers $\mathcal{I}_{\pi}^{\mathrm{request}}$ and $\mathcal{J}_{\pi}^{\mathrm{response}}$, such as queries, retrieved evidence, tool outputs, summaries, or decisions~\cite{gao2023retrieval,shen2023hugginggpt,wu2024autogen,xi2025rise}.

These three paradigms differ in the granularity of the carrier and the resulting level of LM access they require. Split execution exposes internal intermediate hidden states and therefore requires white-box access to the LM~\cite{vepakomma2018split,gu2025vflair}. Collaborative decoding operates at the prediction level and corresponds to grey-box access~\cite{zhang2024cogenesis,thareja2026dp}. Context-augmented collaboration exchanges only semantic outputs and therefore requires only black-box access~\cite{gao2023retrieval,yi2025ecoagent,wu2024autogen}. 

\begin{figure}
    \centering
    \includegraphics[width=\linewidth]{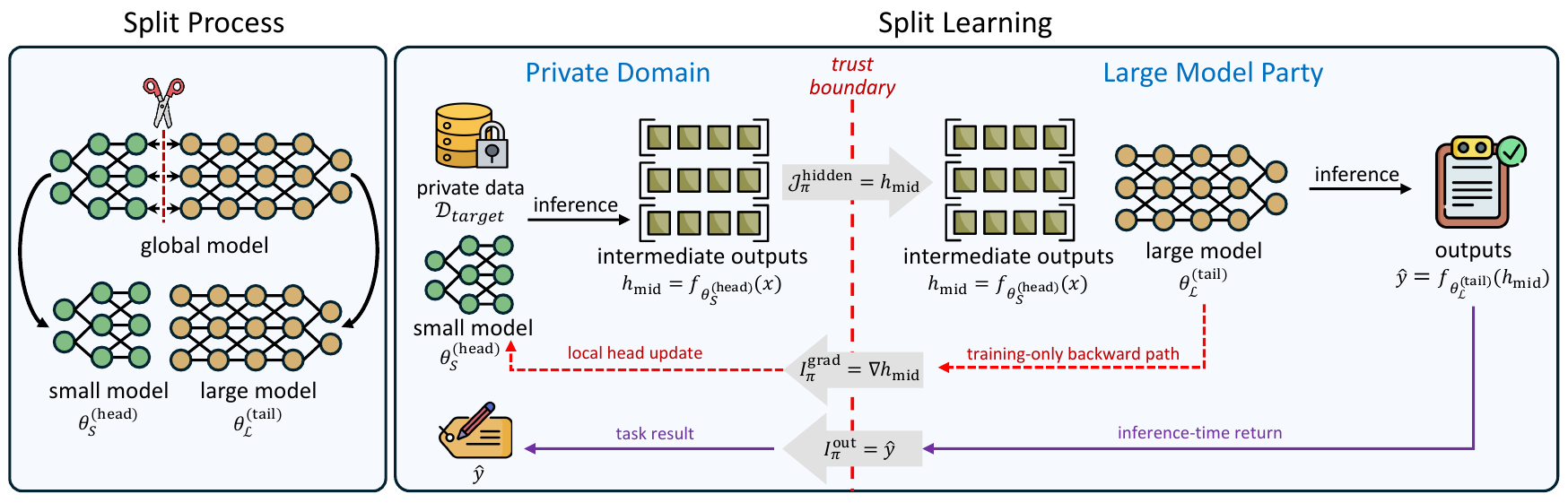}
    \caption{Split learning for cross-silo collaborative inference. The SM transmits the hidden-state carrier $\mathcal{J}_{\pi}^{\mathrm{hidden}}=h_{\mathrm{mid}}$ to the LM tail, whereas the LM returns $\mathcal{I}_{\pi}^{\mathrm{out}}=\hat{y}$ at inference time and $\mathcal{I}_{\pi}^{\mathrm{grad}}=\nabla h_{\mathrm{mid}}$ only during training-time backpropagation.}
    \label{fig:collaborative_split}
\end{figure}

\subsubsection{\textbf{Split Learning}}
\label{subsubsec:split_execution}
Split learning partitions an LM across parties at one or more cut layers~\cite{vepakomma2018split,shen2023split,gu2025vflair}. In a two-part split, the input-side module is deployed on the SM side and transmits a cut-layer representation to the LM-side tail; in a three-part split, the SM side also contains the output-side task head, while the LM side executes the middle layers~\cite{vepakomma2018split,li2024introducing,fedbert,gu2025vflair}. Vertical federated learning (VFL) is a related setting in which parties hold different feature subsets for the same samples and jointly train a model without exposing raw data or model parameters~\cite{liu2024vertical,kairouz2021advances}. In LM--SM collaboration, this exposes intermediate representations rather than raw private inputs or full LM parameters. During inference, the LM-side tail computes subsequent layers and returns the task output; during training or split fine-tuning, it also returns cut-layer gradients for local updates~\cite{vepakomma2018split,shen2023split,mudvari2024splitllm,chen2024adaptive,gu2025vflair}. Accordingly, split execution defines the SM$\rightarrow$LM carrier as an intermediate hidden-state carrier $\mathcal{J}_{\pi}^{\mathrm{hidden}}$ produced at a designated cut layer:
\begin{equation}
\label{eq:split_forward}
\mathcal{J}_{\pi}^{\mathrm{hidden}}(x)
=
h_{\mathrm{mid}}
=
f_{\theta_{\mathcal{S}}^{(\mathrm{head})}}(x).
\end{equation}
where $h_{\mathrm{mid}}$ is the cut-layer hidden state produced by the local head, and $\mathcal{J}_{\pi}^{\mathrm{hidden}}(x)$ denotes the corresponding SM$\rightarrow$LM carrier. In reverse, the LM$\rightarrow$SM carrier is phase-dependent:
\begin{equation}
\label{eq:split_return}
\mathcal{I}_{\pi}(x)
=
\begin{cases}
\mathcal{I}_{\pi}^{\mathrm{out}}(x)=\hat{y}=f_{\theta_{\mathcal{L}}^{(\mathrm{tail})}}(h_{\mathrm{mid}}), & \text{inference time},\\
\mathcal{I}_{\pi}^{\mathrm{grad}}(x)=\nabla h_{\mathrm{mid}}=\dfrac{\partial \ell_{\mathrm{task}}}{\partial h_{\mathrm{mid}}}, & \text{training time}.
\end{cases}
\end{equation}
where $\mathcal{I}_{\pi}^{\mathrm{out}}$ is the task output during inference, while $\mathcal{I}_{\pi}^{\mathrm{grad}}$ is the backpropagated gradient carrier that supports local head updates during training-time split learning or split fine-tuning~\cite{vepakomma2018split, liu2024vertical, mudvari2024splitllm, chen2024adaptive}.
The carriers exchanged in split learning jointly determine its privacy exposure and efficiency cost~\cite{vepakomma2018split,liu2024vertical,li2024introducing}.
Under the privacy constraints in Eq.~\eqref{eq:problem}, this corresponds to a regime that simultaneously activates $M_p(\mathcal{J}_{\pi}^{\mathrm{hidden}}) \le \epsilon_p$ and $M_L(\mathcal{I}_{\pi}) \le \epsilon_L$.
Figure~\ref{fig:collaborative_split} illustrates the split learning process. 

The repeated exchange of intermediate outputs introduces communication overhead and latency, which often becomes the dominant bottleneck under resource-constrained deployments~\cite{mudvari2024splitllm,chen2024adaptive} (i.e., a persistent draw on the communication sub-budget $\epsilon_e^{(comm)}$ at every step).
In practice, despite these challenges, split execution enables LMs to be deployed in heterogeneous systems where local resources are insufficient. Representative implementations include client--server collaborative inference~\cite{mudvari2024splitllm} and adaptive layer splitting for wireless edge environments~\cite{chen2024adaptive}.
To further improve the efficiency of the split training pipeline, PEFT methods~\cite{wang2023privatelora}, local update and data pruning~\cite{cao2024sfprompt} are incorporated.
In addition, previous works have also proposed various frameworks for three-part split learning by keeping both input-side and output-side components local while offloading the LM body, reducing local computation without exposing raw inputs or task supervision~\cite{gu2025vflair, chen2024unveiling, lin2024splitlora, shen2023split}.

\begin{figure}
    \centering
    \includegraphics[width=\linewidth]{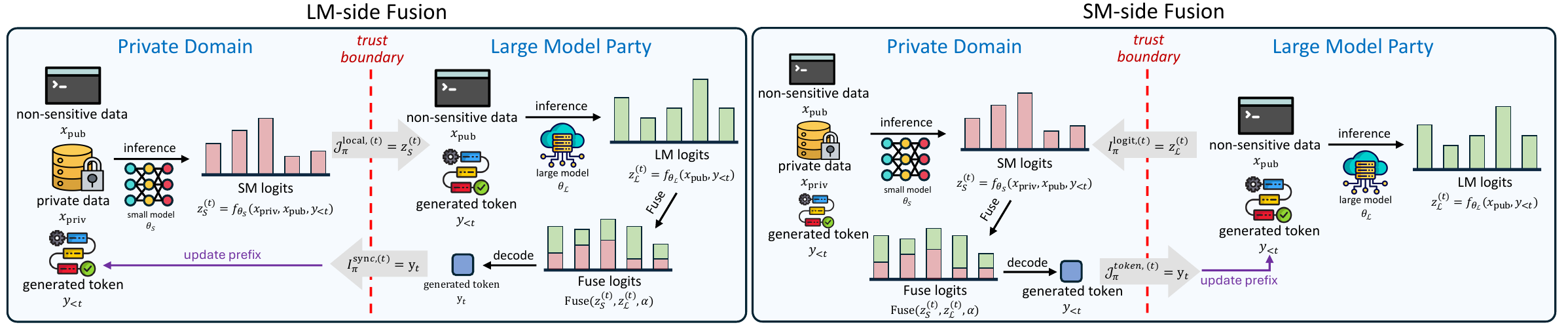}
    \caption{Collaborative decoding for cross-silo collaborative inference. \textit{LM-side fusion} transmits SM-side logits to the LM, whereas \textit{SM-side fusion} transmits LM-side logits to the SM and synchronizes the decoded token.}
    \label{fig:collaborative_decoding}
\end{figure}

\subsubsection{\textbf{Collaborative Decoding}}
\label{subsubsec:collaborative_decoding}
Collaborative Decoding~\cite{zhang2024fast} leverages the capabilities of both a LM and a SM to improve the effectiveness and efficiency at decoding phase. Different collaborative decoding approaches focus on different objectives, \eg speculative decoding~\cite{xia2024unlocking,chen2024cascade,leviathan2023fast,kim2024speculative,sun2024spectr,liu2024online,narasimhanfaster,bachmannjudge} to improve inference efficiency, and contrastive decoding~\cite{li2023contrastive} to improve generation for LMs. To achieve the adaptation of LMs to private domains while avoiding fine-tuning of LMs, proxy or emulator tuning~\cite{liu2024tuning,mitchell2024emulator,xu2024small,he2024cpt,ormazabal2023comblm,ronaghi2026training,lv2026specsteer} leverages the fine-tuning of a smaller, more efficient model to emulate the behavior of a larger, more complex model. While most works in this direction focus on inference-time efficiency and assume both models decode from the same contextual input $x$, they overlook privacy constraints. We therefore focus below on privacy-sensitive fusion under restricted LM context. By contrast, recent studies~\cite{zhang2024cogenesis,thareja2026dp,bhusal2026privacy} have begun to explore how to handle private context in such LM--SM collaborative decoding systems. Here we formulate this problem as follows. We assume the original data is composed of both private data $x_{\mathrm{priv}}$ and non-sensitive data $x_{\mathrm{pub}}$, that is $x = (x_{\mathrm{priv}}, x_{\mathrm{pub}})$ .

In the privacy-sensitive setting, the LM is restricted to the non-sensitive context $x_{\mathrm{pub}}$, while the SM observes both $x_{\mathrm{priv}}$ and $x_{\mathrm{pub}}$. At decoding step $t$, the SM forms private-context next-token logits and the LM forms public-context next-token logits:
\begin{equation}
\label{eq:collab_decoding_scores}
z_{\mathcal{S}}^{(t)}
=
f_{\theta_{\mathcal{S}}}(x_{\mathrm{priv}}, x_{\mathrm{pub}}, y_{<t}),
\qquad
z_{\mathcal{L}}^{(t)}
=
f_{\theta_{\mathcal{L}}}(x_{\mathrm{pub}}, y_{<t}).
\end{equation}
where $z_{\mathcal{S}}^{(t)}$ and $z_{\mathcal{L}}^{(t)}$ denote the SM-side and LM-side next-token logits at decoding step $t$, respectively, and $y_{<t}$ denotes the prefix of generated tokens before decoding step $t$.
The next token is then sampled from the fused logits:
\begin{equation}
\label{eq:collab_decoding_fusion}
y_t
=
\operatorname{Decode}\!\left(
\operatorname{Fuse}(z_{\mathcal{S}}^{(t)}, z_{\mathcal{L}}^{(t)}; \alpha_t);
\Omega_t
\right),
\end{equation}
where $\operatorname{Fuse}(\cdot)$ denotes a method-specific logit-combination rule and $\alpha_t$ controls the fusion weight.
We distinguish two fusion strategies by where fusion is performed. In \emph{LM-side fusion}, the SM transmits $z_{\mathcal{S}}^{(t)}$ to the LM; the LM fuses it with $z_{\mathcal{L}}^{(t)}$ and returns the sampled token $y_t$~\cite{fan2024giant,zhang2024cogenesis,thareja2026dp,bhusal2026privacy}. In \emph{SM-side fusion}, the LM transmits $z_{\mathcal{L}}^{(t)}$ to the SM; the SM performs fusion locally and returns the sampled token $y_t$ to the LM~\cite{lv2025costeer,lv2026specsteer}:
\begin{equation}
\label{eq:collab_decoding_carriers}
\mathcal{J}_{\pi}^{(t)}
=
\begin{cases}
\mathcal{J}_{\pi}^{\mathrm{logit},(t)}=z_{\mathcal{S}}^{(t)}, & \text{LM-side fusion},\\
\mathcal{J}_{\pi}^{\mathrm{token},(t)}=y_t, & \text{SM-side fusion},
\end{cases}
\qquad
\mathcal{I}_{\pi}^{(t)}
=
\begin{cases}
\mathcal{I}_{\pi}^{\mathrm{token},(t)}=y_t, & \text{LM-side fusion},\\
\mathcal{I}_{\pi}^{\mathrm{logit},(t)}=z_{\mathcal{L}}^{(t)}, & \text{SM-side fusion}.
\end{cases}
\end{equation}
Figure~\ref{fig:collaborative_decoding} illustrates the two fusion strategies.
In LM-side fusion, the transmitted SM logits preserve vocabulary-level private-context information. Methods therefore process this logit carrier before or during transmission: adaptive weighting adjusts the contribution of SM logits~\cite{fan2024giant,zhang2024cogenesis}, DP-based fusion bounds leakage from the transmitted logits~\cite{thareja2026dp,bhusal2026privacy}, and sparse logit transmission reduces the amount of logit information crossing the boundary~\cite{zhang2024cogenesis}.
In SM-side fusion, private context stays on the SM side. The LM transmits $z_{\mathcal{L}}^{(t)}$ to the SM, and the SM combines it with $z_{\mathcal{S}}^{(t)}$ and returns only the sampled token $y_t$~\cite{lv2025costeer,lv2026specsteer}. Since only the sampled token crosses the boundary, no vocabulary-level logit information is exposed, though the shared autoregressive prefix can still reveal private-context influence~\cite{morris2024language, thareja2026dp, debenedetti2024privacy}.

The two fusion strategies also differ in efficiency: LM-side fusion transmits a vocabulary-sized logit vector at each decoding step, whereas SM-side fusion returns only a single token but still requires an LM scoring call per step~\cite{zhang2024fast,leviathan2023fast}.

\subsubsection{\textbf{Context-Augmented Collaboration}}
\label{subsubsec:context_augmented_collaboration}
In this section, we investigate frameworks ~\cite{gao2023retrieval, shen2023hugginggpt, xi2025rise} that coordinate LMs and SMs through \textit{output-level contextual carriers}, rather than intermediate activations~\cite{vepakomma2018split, wang2023privatelora, fedbert} or token-level decoding signals~\cite{leviathan2023fast, li2023contrastive, liu2024tuning}. In this framework, the LM constructs a contextual request carrier $\mathcal{I}_{\pi}^{\mathrm{request},(t)}$ from non-sensitive context and an LM-visible interaction state, while the SM returns a contextual response carrier $\mathcal{J}_{\pi}^{\mathrm{response},(t)}$ that may be derived from local private information but is returned only as an external output without exposing SM internals. Formally, at interaction step $t$:
\begin{equation}
\label{eq:context_request_response}
\mathcal{I}_{\pi}^{\mathrm{request},(t)}
=
\Psi_{\mathrm{request}}^{(t)}(x_{\mathrm{pub}},h^{(t)}),
\qquad
\mathcal{J}_{\pi}^{\mathrm{response},(t)}
=
\Phi_{\mathrm{response}}^{(t)}\!\left(
\mathcal{I}_{\pi}^{\mathrm{request},(t)},
\theta_{\mathcal{S}},
x_{\mathrm{priv}},
h^{(t)}
\right),
\end{equation}
\begin{equation}
\label{eq:context_update}
h^{(t+1)}, \hat{y}^{(t+1)}
=
\mathrm{Update}_{\theta_{\mathcal{L}}}\!\left(
x_{\mathrm{pub}},
h^{(t)},
\mathcal{J}_{\pi}^{\mathrm{response},(t)}
\right),
\end{equation}
where $\mathcal{I}_{\pi}^{\mathrm{request},(t)}$ is the LM$\rightarrow$SM contextual request carrier, $\mathcal{J}_{\pi}^{\mathrm{response},(t)}$ is the SM$\rightarrow$LM contextual response carrier, and $h^{(t)}$ denotes the LM-visible task state accumulated before step $t$. The operator $\Psi_{\mathrm{request}}^{(t)}(\cdot)$ constructs requests such as queries, prompts, subtasks, or draft answers~\cite{shen2023hugginggpt, shinn2023reflexion}; $\Phi_{\mathrm{response}}^{(t)}(\cdot)$ constructs responses such as retrieved evidence, tool outputs, summaries, critiques, or decisions~\cite{gao2023retrieval, liang2024taskmatrix, wu2024autogen}; and $\mathrm{Update}_{\theta_{\mathcal{L}}}(\cdot)$ denotes the LM-side state update or answer-generation step after the response carrier is returned. If either the request or response construction transmits information derived from $x_{\mathrm{priv}}$ or private memory, that transmitted carrier must be counted under the data-leakage budget $M_p$.

\begin{figure}
    \centering
    \includegraphics[width=\linewidth]{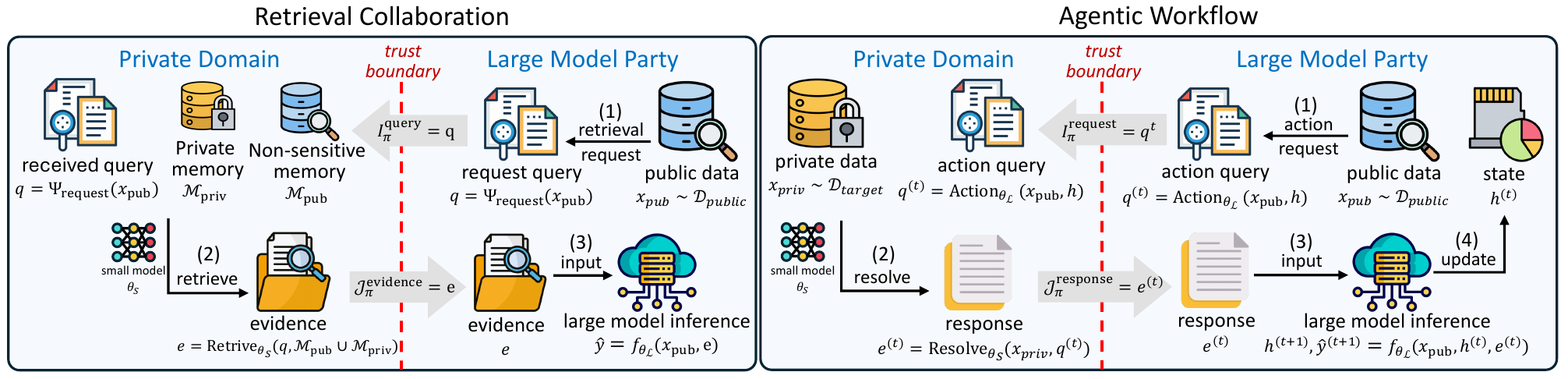}
    \caption{Context-augmented collaboration for cross-silo collaborative inference. The carrier is output-level context: \textit{Retrieval Collaboration} exchanges query and evidence once, whereas \textit{Agentic Workflow} iterates request and response carriers.}
    \label{fig:collaborative_context}
\end{figure}

In the following, we examine two key examples for this setting, i.e., Eqs.~\eqref{eq:context_request_response}--\eqref{eq:context_update}: Retrieval collaboration and Agentic collaboration. 
Figure~\ref{fig:collaborative_context} illustrates the two context-augmented collaboration patterns under this context-based collaborative paradigm.

\paragraph{\textbf{Retrieval Collaboration}}
Retrieval-augmented generation (RAG) \cite{gao2023retrieval} has drawn considerable attention for helping LMs use external knowledge sources. A small model can be employed as a retriever, either trained end-to-end \cite{NEURIPS2020_retrieval, RAG-end2end} with the generative model or tuned with a frozen black-box LM \cite{shi-etal-2024-replug,li2025blade}. In privacy-aware settings, an SM-side retriever maps an LM request to task-relevant evidence from an SM-side retrieval memory partitioned into public and private components~\cite{rag-privacy}, returning it to the LM as an output-level contextual carrier ~\cite{gao2023retrieval}. Formally:
\begin{equation}
\label{eq:retrieval_carrier}
\mathcal{I}_{\pi}^{\mathrm{query}}
=
q
=
\Psi_{\mathrm{request}}(x_{\mathrm{pub}}),
\qquad
\mathcal{J}_{\pi}^{\mathrm{evidence}}
=
e
=
\mathrm{Retrieve}_{\theta_{\mathcal{S}}}\!\left(
q,
\mathcal{M}_{\mathrm{pub}}\cup\mathcal{M}_{\mathrm{priv}}
\right),
\end{equation}
\begin{equation}
\label{eq:retrieval_solve}
\hat{y}
=
f_{\theta_{\mathcal{L}}}\!\left(
x_{\mathrm{pub}},
e
\right),
\end{equation}
where $q$ denotes the LM$\rightarrow$SM contextual retrieval request carrier, $\mathcal{M}_{\mathrm{pub}}$ and $\mathcal{M}_{\mathrm{priv}}$ denote the non-sensitive and private SM-side memories, respectively, and $e$ denotes the returned SM$\rightarrow$LM evidence carrier, through which retrieval over $\mathcal{M}_{\mathrm{priv}}$ exposes private context~\cite{rag-privacy}. $\hat{y}$ denotes the final LM response produced by conditioning on the public context and the returned evidence. A small model can be employed as a retriever, trained either end-to-end~\cite{NEURIPS2020_retrieval, RAG-end2end} with the generative model or tuned with a frozen black-box LM~\cite{shi-etal-2024-replug}. Further improvements target query construction through instruction-aware or query-aware retrieval~\cite{lin2023ra, li2025unirag}, and retrieval feedback through adaptive or self-reflective strategies that decide when and what to retrieve~\cite{asaiself, yao2025seakr}.

These studies did not consider the privacy risks~\cite{rag-privacy} of RAG. On the request side, under the canonical formulation above, the query $q$ is privacy-preserving only when $\Psi_{\mathrm{request}}$ is restricted to $x_{\mathrm{pub}}$; if the request is initiated or refined using private context, the request carrier itself may expose sensitive information~\cite{zeng2025mitigating,cheng2025remoterag}. Recent methods mitigate this leakage by perturbing or synthesizing retrieval-side representations before cloud retrieval~\cite{zeng2025mitigating, cheng2025remoterag}. On the response side, since evidence grounded in $\mathcal{M}_{\mathrm{priv}}$ is derived from $\mathcal{D}_{target}$ or other private resources~\cite{rag-privacy}, the evidence carrier $e$ becomes the primary $M_p$ exposure surface, and balancing evidence utility against context exposure at both stages remains an open challenge.

\paragraph{\textbf{Agentic Workflow}}

Multi-agent collaboration coordinates LM agents through role specialization, planning, tool use, and iterative communication~\cite{wu2024autogen,hong2023metagpt,qian2024chatdev,xi2025rise,plaat2025agentic}. In LM--SM agent collaboration, the LM often acts as a planner or coordinator, while SM or local agents execute privacy-sensitive, device-specific, or low-cost subtasks under local context~\cite{shen2023hugginggpt,liang2024taskmatrix,wang2024mobile,fancore,yi2025ecoagent,chen2025smurfs}. We therefore model this setting as a multi-round exchange of action requests and responses across the trust boundary.
Agentic collaboration repeatedly applies the request--response--update (Eqs.~\eqref{eq:context_request_response}--\eqref{eq:context_update}) in a loop: the LM-visible state $h^{(t)}$ evolves across $T$ steps, and each returned carrier $\mathcal{J}_{\pi}^{\mathrm{response},(t)}$ may encode a plan execution result, tool output, summary, critique, or decision~\cite{wu2024autogen, xi2025rise}.  This multi-round instantiation can be summarized as:
\begin{equation}
\label{eq:agentic_workflow_carrier}
\mathcal{I}_{\pi}^{\mathrm{request},(t)}
=
\mathrm{Action}_{\theta_{\mathcal{L}}}\!\left(x_{\mathrm{pub}},h^{(t)}\right),
\qquad
\mathcal{J}_{\pi}^{\mathrm{response},(t)}
=
\mathrm{Resolve}_{\theta_{\mathcal{S}}}\!\left(
\mathcal{I}_{\pi}^{\mathrm{request},(t)},x_{\mathrm{priv}}
\right),
\end{equation}
\begin{equation}
\label{eq:agentic_workflow_overview}
h^{(t+1)},\hat{y}^{(t+1)}
=
\mathcal{f}_{\theta_{\mathcal{L}}}\!\left(
x_{\mathrm{pub}},
h^{(t)},
\mathcal{J}_{\pi}^{\mathrm{response},(t)}
\right),
\qquad t=0,\ldots,T-1 .
\end{equation}
Here, $\mathcal{I}_{\pi}^{\mathrm{request},(t)}$ is the LM$\rightarrow$SM contextual action request, while $\mathcal{J}_{\pi}^{\mathrm{response},(t)}$ is the SM$\rightarrow$LM bounded contextual response. $\mathrm{Action}_{\theta_{\mathcal{L}}}(\cdot)$ emits the next action request, and $\mathrm{Resolve}_{\theta_{\mathcal{S}}}(\cdot)$ performs private-side execution or evaluation before returning the response carrier. Existing agentic systems can then be organized by the \textbf{semantic content} of the two carriers. 
\textit{Hierarchical planning} uses carriers that primarily encode task decomposition and execution state. In this pattern, $\mathrm{Action}_{\theta_{\mathcal{L}}}$ specializes to planning: the request carrier is a plan or subtask $p^{(t)}$ constructed from $x_{\mathrm{pub}}$ and $h^{(t)}$. The private-side resolver then executes the subtask with local context and returns an execution summary or tool result $m^{(t)}$ as the response carrier. The LM therefore acts as planner or coordinator, while SMs resolve privacy-sensitive or routine subtasks locally~\cite{ahn2022can, shen2023hugginggpt, liang2024taskmatrix, wang2024mobile, wangvoyager, zhang2024chain, hong2023metagpt, qian2024chatdev, parmar2025plangen, chen2025smurfs, fancore, yi2025ecoagent}. Efficiency improves when local executors handle low-uncertainty subtasks without invoking the full LM planner~\cite{chen2023frugalgpt, chen2025smurfs, yi2025ecoagent}; privacy is strengthened when the LM plans conditioned on $x_{\mathrm{pub}}$ only and the SM executes locally using $x_{\mathrm{priv}}$, so that only the execution summary crosses the boundary~\cite{yi2025ecoagent, fancore}.

\textit{Intelligence fusion} instead uses carriers that primarily encode evaluation and correction signals. In this pattern, $\mathrm{Action}_{\theta_{\mathcal{L}}}$ specializes to drafting: the request carrier is an LM draft $\hat{y}_{\mathcal{L}}^{(t)}$ generated from the public context and current state. The private-side resolver assesses this draft against local context and returns feedback $r^{(t)}$ as the response carrier. The LM then refines its solution through the generic update step in Eqs.~\eqref{eq:agentic_workflow_carrier}--\eqref{eq:agentic_workflow_overview}, allowing SMs to serve as specialized judges or feedback providers while the LM synthesizes the final solution~\cite{liang2024encouraging, shinn2023reflexion, jiang2023llm, wangmixture, chen2024reconcile}. Since $r^{(t)}$ is derived from private-side evaluation, it can reveal private domain characteristics even without exposing $x_{\mathrm{priv}}$ directly.

Privacy-aware extensions of agentic LM--SM collaboration have gained increasing attention as agent systems are deployed in sensitive domains. The core challenge is that returned carriers, whether retrieved evidence, execution summaries, tool results, critiques, or decisions, may individually appear innocuous yet collectively reveal private domain characteristics through their temporal sequence~\cite{wang2025unveiling}. Tool-using agents face an additional exposure point: each tool call and its returned result constitutes an explicit carrier exchange that may encode $x_{\mathrm{priv}}$ in the tool input or expose sensitive content in the tool output~\cite{zhang2024privacyasst}. Federated multi-agent settings address carrier-level leakage by coordinating across private silos through selective disclosure protocols that restrict carrier content to task-relevant abstractions, so that no raw private data crosses organizational boundaries~\cite{shi2025federatedmas, chen2025fed}. Providing formal leakage bounds on the full response-carrier trajectory $M_p(\mathcal{D}_{target}, \theta_{\mathcal{S}}, \{\mathcal{J}_{\pi}^{\mathrm{response},(t)}\}_{t=0}^{T}) \le \epsilon_p$ remains an open problem, since the sequential nature of agentic interaction creates a compositional leakage risk that per-step privacy analyses do not capture.

 Among the three inference paradigms in Section~\ref{sec:cross-silo-collaboration}, context-augmented collaboration requires only black-box access to the LM and exchanges output-level semantic carriers, making it the most naturally compatible with privacy-sensitive deployment. However, it is not privacy-free: retrieval evidence, execution summaries, tool outputs, and feedback trajectories are all $\mathcal{J}_{\pi}$ exposure surfaces whose leakage depends on the content and composition of returned responses~\cite{rag-privacy,zeng2025mitigating,cheng2025remoterag,wang2025unveiling,zhang2024privacyasst}. The central open problem is therefore not only single-message sanitization, but trajectory-level accounting for multi-round contextual collaboration.

\subsection{Multi-party Knowledge Transfer}
\label{sec:multi-party}
While previous sections focus on knowledge transfer techniques between a single LM and a SM, this section discusses collaborative approaches considering either multiple SMs or multiple LMs. Specifically, federated learning and LM fusion are representative techniques for tackling each of these scenarios, respectively.
In this setting, collaboration expands along the participation dimension and naturally leads to two representative scenarios: federated LM--SM collaboration (multiple SMs to one LM) and LM ensemble/fusion (multiple LMs to one SM). This setting generalizes the two-party formulation in Eq.~\eqref{eq:problem} to $K$ data parties, each holding a private domain $\mathcal{D}_{target}^{(k)}$ and a local model $\mathcal{S}_k$. The multi-party collaborative objective becomes:
\begin{equation}
\label{eq:multi_party}
\begin{aligned}
\max_{\theta_{\pi}} \sum_{k=1}^{K} w_k \,\mathcal{F}\!\left(\theta_{\mathcal{L}},\, \mathcal{D}_{target}^{(k)},\, \theta_{\mathcal{S}_k},\, \pi\right),
\quad \text{s.t.} \quad \forall\,k:\;&
M_p\!\left(\mathcal{D}_{target}^{(k)},\theta_{\mathcal{S}_k},\mathcal{J}_{\pi}^{(k)}\right) \le \epsilon_p^{(k)},\\
&
M_L\!\left(\theta_{\mathcal{L}},\mathcal{I}_{\pi}^{(k)}\right) \le \epsilon_L^{(k)},\quad
M_r\!\left(\mathcal{I}_{\pi}^{(k)},\mathcal{J}_{\pi}^{(k)},\pi\right) \le \epsilon_r^{(k)},\\
&
M_e\!\left(\mathcal{D}_{target}^{(k)},\theta_{\mathcal{S}_k},\pi\right) \preceq \boldsymbol{\epsilon}_e^{(k)} .
\end{aligned}
\end{equation}
where $w_k$ denotes the participation weight of the $k$-th data party and the per-client privacy, integrity, and efficiency constraints generalize those in Eq.~\eqref{eq:problem}. $\mathcal{J}_{\pi}^{(k)}$ and $\mathcal{I}_{\pi}^{(k)}$ denote the outbound and returned carriers associated with the $k$-th data party, respectively. The homogeneous-budget case is recovered by setting $\epsilon_p^{(k)}=\epsilon_p$, $\epsilon_L^{(k)}=\epsilon_L$, $\epsilon_r^{(k)}=\epsilon_r$, and $\boldsymbol{\epsilon}_e^{(k)}=\boldsymbol{\epsilon}_e$ for all $k$. Compared with two-party collaboration, the central challenge shifts to aggregating heterogeneous carriers from multiple private domains under communication, privacy, and integrity constraints. We next discuss federated LM--SM collaboration and LM ensemble/fusion under this unified view.

\paragraph{\textbf{Federated LM--SM Collaboration}}
Federated Learning (FL)~\cite{mcmahan2017communication,yang2019federated,ciucanu2022samba} enables collaborative training between multiple decentralized data parties. One of the most difficult problems in federated learning is dealing with data heterogeneity and knowledge bias across different data parties.
In LM--SM federated collaboration, these difficulties are addressed by exchanging transferable carriers rather than raw private data. Across parameter-based, distillation-based, and generation-based instantiations, the server-side update at each communication round can be unified as:
\begin{equation}
\label{eq:fl_aggregate}
\theta_{\mathcal{L}}^{(t+1)}
=
\mathrm{Fuse}\!\left(\theta_{\mathcal{L}}^{(t)},\; \left\{\mathcal{J}_{\pi}^{(k,t)}\right\}_{k=1}^{K}\right),
\end{equation}
where $\mathcal{J}_{\pi}^{(k,t)}$ denotes the carrier uploaded by the $k$-th client at round $t$, and $\mathrm{Fuse}(\cdot)$ denotes an aggregation rule instantiated by the selected transfer mechanism. Concretely, client carriers may be parameter deltas such as adapters or soft prompts~\cite{fan2023fatellm, zhang2023fedpetuning, dong2023tunable}, distilled supervision such as logits on a public set~\cite{cheng2021fedgems, fan2025fedmkt, yumultimodal}, synthetic data or generated feedback~\cite{deng2023mutual}, prototype-level statistics used to stabilize heterogeneous aggregation~\cite{jin2025fedcpd, wu2024global}, and, in recent settings, reasoning-level signals under privacy constraints~\cite{chen2025fedcot}. Each uploaded carrier contributes to the corresponding per-client $M_p^{(k)}$ budget, so aggregation quality and privacy accounting are coupled rather than separable design choices.

Existing federated LM--SM approaches can be organized by the transfer mechanism employed. Parameter-based transfer aggregates knowledge from multiple SMs to LMs through lightweight adapters or soft prompts~\cite{fan2023fatellm, kuang2023federatedscopellm, ye2024openfedllm, dong2023tunable, lu2023zoopflexploringblackboxfoundation}, addressing data heterogeneity and knowledge bias. Distillation-based techniques establish bi-directional knowledge transfer between LMs and SMs through logit or representation fusion~\cite{cheng2021fedgems, yumultimodal, fan2025fedmkt}. Generation-based approaches enable mutual enhancement through synthetic data or feedback exchange~\cite{deng2023mutual}. Additionally, offsite tuning has been extended to federated settings~\cite{wu2024fedbiot, chua2024fedpeat} to support efficient LM-to-SM knowledge transfer.

A key challenge in federated LM--SM collaboration is handling system and model heterogeneity during carrier aggregation. Recent works address this through specialized aggregation operators: FLoRA~\cite{wang2024flora} uses stacking-based aggregation to support heterogeneous LoRA ranks; FedEx-LoRA~\cite{singhalfedex} introduces residual correction to reduce aggregation error; and HLoRA~\cite{liu2025hlora} optimizes convergence under heterogeneous resource budgets. These advances indicate that stable global gains depend not only on whether clients upload useful local knowledge, but critically on whether the aggregation operator can align heterogeneous update spaces across diverse client configurations.

\paragraph{\textbf{LM Ensemble and Fusion}}
Since different LMs of various architectures trained with distinct datasets should have different angles considering the same task, transferring collective knowledge from multiple LMs to a SM presents an interesting opportunity. Previous works focus primarily on the exploration of knowledge transfer from LMs to SMs, using distillation-based~\cite{wan2024knowledge} or parameter-based~\cite{wan2024fusechat, jin2023dataless, zhang2023composing} approaches.
This line can be formulated as a multi-source fusion problem, where the student learns from an aggregated supervision signal constructed from multiple LMs. Formally, the LM ensemble objective can be written as:
\begin{equation}
\label{eq:lm_fusion}
\theta_{\mathcal{S}}^{*}
=
\arg\min_{\theta_{\mathcal{S}}}
\mathbb{E}_{X\sim\mathcal{D}}
\ell_{\mathrm{task}}\!\left(
f_{\theta_{\mathcal{S}}}(X),\;
\mathrm{Fuse}\!\left(\left\{f_{\theta_{\mathcal{L}_j}}(X)\right\}_{j=1}^{M}\right)
\right),
\end{equation}
where $\theta_{\mathcal{L}_j}$ denotes the parameters of the $j$-th LM, $M$ is the number of source LMs, and $\mathrm{Fuse}(\cdot)$ denotes a fusion operator instantiated at different granularities. Existing methods can be organized into three representative groups: logit-level fusion for prediction-space alignment~\cite{wan2024knowledge, li2025infifusion}, parameter-level fusion for update-space composition and conflict mitigation~\cite{wan2024fusechat, jin2023dataless, zhang2023composing, sun2025catmerging}, and output-level fusion that aggregates generated candidates through quality-aware preference optimization~\cite{zou2024fusegen, wan2025fuserl}.

One important challenge is ensuring effective fusion of knowledge from potentially conflicting or redundant LLMs. Solutions may involve using the feedback of SMs, as discussed in Section~\ref{sec:knowledge_transfer_SM2LM}, to provide more accurate guidance and evaluations for LM domain adaptability, ultimately leading to a quality-aware fusion strategy that optimizes the use of available LLMs. FuseGen~\cite{zou2024fusegen} recently explored the transfer of knowledge from SMs to LMs through a generation-based approach.
Overall, multi-party collaboration goes beyond Eq.~\eqref{eq:problem} by introducing the number of knowledge carriers and the heterogeneous aggregation mechanisms into a joint design problem.

\section{Challenges and Opportunities}
\label{sec:challenges}

The taxonomy in Section~\ref{sec:taxonomy} characterizes LM--SM collaboration
through the lens of \emph{what is exchanged across the trust boundary}: logits
and representations, tokens, synthetic data, intermediate outputs,
compressed weights, adapters, tunable prompts, and natural-language prompts
and outputs. While this
carrier-centric view organizes the design space, it also
exposes a fundamental tension that constrains practical deployment. Each
carrier type simultaneously determines how much private-domain knowledge is
externalized, how much computation and communication is required to transmit it,
and how faithfully the receiving party can exploit it for the target task. In
the notation of Eq.~\eqref{eq:problem}, the same carrier is the variable through
which the utility objective $\mathcal{F}$, the leakage measures $M_p$ and
$M_L$, the integrity risk $M_r$, and the resource cost $M_e$ become coupled. Jointly optimizing utility, privacy and security assurance, and efficiency without trade-offs is generally infeasible, and current systems
typically sacrifice at least one~\cite{zhang2023trading, chen2020breaking}.
This tension helps explain why, despite rapid methodological progress,
wide deployment of LM--SM collaboration remains limited in domains where data
governance, model ownership, and resource budgets must be satisfied
simultaneously.

In this section, we analyze the resulting challenges along three complementary
axes. Section~\ref{sec:privacy_threats} examines \textbf{security and privacy
threats} arising from cross-boundary information flow, covering data-privacy, model-security, and integrity attacks organized by carrier
type, and surveys the corresponding protection techniques.
Section~\ref{sec:efficiency_considerations}
characterizes \textbf{efficiency constraints} in terms of communication,
computation, and LM query cost, and maps known optimization strategies onto
the collaboration paradigms where they apply.
Finally, Section~\ref{sec:tri_lemma} synthesizes these constraints into a \textbf{Collaborative Trilemma}, a three-way tension
between domain utility, privacy and security assurance, and resource efficiency, and
identifies the open design challenges that remain.

\subsection{Security and Privacy Threats}
\label{sec:privacy_threats}

In conventional deployments, privacy risks are typically analyzed within a single training or serving pipeline, where extracting private data from trained LMs is a continuous area of research, such as training data memorization~\cite{carlini2021extracting}, query-time extraction attacks~\cite{nasr2023scalable, orekondy2019knockoff, krishna2019thieves, birch2023model, carlini2024stealing}, and inference attacks on model outputs~\cite{nasr2019comprehensive, fu2024membership}. The LM--SM collaboration opens doors to new data privacy attacks that exploit   \textbf{cross-boundary information carriers}~\cite{yang2019federated, kairouz2021advances, zhu2019deep, he2019model, gao2023pcat}, thereby creating additional leakage channels~\cite{melis2019exploiting,lin2025against}.
These carriers act as observable proxies of private-domain knowledge~\cite{zhu2019deep, takahashi2023breaching, dimitrov2024spear, wang2025privacy}. In context-augmented collaboration, for example, prompts and outputs may contain retrieved evidence, execution summaries, feedback, tool results, or agent traces~\cite{greshake2023not, debenedetti2024agentdojo, flemings2025estimating, hui2024pleak}. 
The severity of their exposure depends on what is shared, how reversible it is, and where the trust boundary lies: SM-side outbound carriers $\mathcal{J}_{\pi}$ primarily expose private-domain information, so their risk is measured by $M_p$, whereas LM-side returned carriers $\mathcal{I}_{\pi}$ may expose proprietary model behavior, so their risk is measured by $M_L$. In addition, carriers in both directions are consumed by the receiving party, so a poisoned or manipulated carrier can corrupt downstream training or inference; this integrity risk is measured by $M_r$. In this section, we first review representative attacks on these collaborative carriers, and then move on to discuss corresponding protection techniques.

\subsubsection{Carrier-Specific Data Privacy Attacks}\label{subsubsec:carrier_data_privacy_attacks}
Even without direct access to raw private corpora, an adversary can still exploit the \emph{transmitted carriers} exchanged across the trust boundary, which act as observable proxies of private-domain knowledge and open attack channels that are absent in fully local execution~\cite{melis2019exploiting, zhu2019deep, carlini2021extracting, flemings2025estimating, hui2024pleak}. This subsection focuses on data-privacy leakage, which occurs when SM-side outbound carriers $\mathcal{J}_{\pi}$ derived from $\mathcal{D}_{target}$ or private context cross the trust boundary toward the LM, allowing private-domain information to be inferred or extracted from the transmitted signal; carriers flowing in the reverse direction, from LM to SM, instead expose proprietary LM behavior under $M_L$ (\S~\ref{subsubsec:model_security_attacks}). Table~\ref{tab:carrier_attack_alignment} summarizes representative data-privacy attacks by carrier type, attack target, and key references.

\begin{table*}[t]
\centering
\caption{Representative data-privacy attacks on LM--SM collaborative carriers, organized by carrier type, attack target, and key references.}
\label{tab:carrier_attack_alignment}
\renewcommand{\arraystretch}{0.90}
\resizebox{\textwidth}{!}{
\begin{tabular}{l|l|l|l}
\toprule
\textbf{Transferred Carrier} & \textbf{Representative Attacks} & \textbf{Attack Target} & \textbf{Key References} \\
\midrule

\multirow{5}{*}{Logits / Representations}
& Paired-Logits Inversion Attack & Private class patterns & \keyref{takahashi2023breaching}; \keyref{xiao2024privacy} \\
\cmidrule{2-4}
& Membership Inference Attacks & Membership of private records & \keyref{yang2022fd}; \keyref{shi2025unveiling} \\
\cmidrule{2-4}
& Property Inference Attacks & Private-domain properties & \keyref{melis2019exploiting}; \keyref{shi2025unveiling} \\
\cmidrule{2-4}
& Embedding Inversion Attacks & Semantic content or attributes & \keyref{li2023sentence}; \keyref{huang2024transferable} \\
\cmidrule{2-4}
& Prompt Inversion Attacks & Private prompt or context & \keyref{morris2024language}; \keyref{qu2025prompt} \\
\midrule

Tokens
& Side-Channel Attacks & Private-context influence & \keyref{wei2024speculation}; \keyref{debenedetti2024privacy} \\
\midrule

\multirow{4}{*}{Synthetic Data}
& Membership Inference Attacks & Membership of source records & \keyref{hayes2019logan}; \keyref{guepin2023synthetic} \\
\cmidrule{2-4}
& Attribute Inference Attacks & Sensitive attributes & \keyref{stadler2022synthetic}; \keyref{giomi2023unified} \\
\cmidrule{2-4}
& Training Data Extraction Attacks & Memorized private content & \keyref{carlini2021extracting}; \keyref{nasr2023scalable} \\
\cmidrule{2-4}
& Property Inference Attacks & Aggregate domain properties & \keyref{giomi2023unified}; \keyref{wang2024property} \\
\midrule

\multirow{4}{*}{Intermediate Outputs}
& Input Reconstruction Attacks & Private inputs & \keyref{he2019model}; \keyref{gao2023pcat} \\
\cmidrule{2-4}
& Attribute Inference Attacks & Sensitive input attributes & \keyref{pasquini2021unleashing}; \keyref{driouich2022attribute} \\
\cmidrule{2-4}
& Label Inference Attacks & Private labels & \keyref{erdougan2022unsplit}; \keyref{liu2024similarity} \\
\cmidrule{2-4}
& Gradient Inversion Attacks & Private examples and labels & \keyref{zhu2019deep}; \keyref{geiping2020inverting} \\
\midrule

\multirow{3}{*}{Adapters / Tunable Prompts}
& PEFT Gradient Inversion Attacks & Private fine-tuning examples & \keyref{sami2025gradient} \\
\cmidrule{2-4}
& Membership Inference Attacks & Training-set membership & \keyref{mireshghallah2022memorization}; \keyref{ran2025loraleak} \\
\cmidrule{2-4}
& User Inference Attacks & User-level participation & \keyref{kandpal2024user} \\
\midrule

\multirow{4}{*}{Prompts / Outputs}
& RAG Data Extraction Attacks & Private retrieved content & \keyref{rag-privacy}; \keyref{flemings2025estimating} \\
\cmidrule{2-4}
& RAG Membership Inference Attacks & Datastore membership & \keyref{anderson2025my}; \keyref{naseh2025riddle} \\
\cmidrule{2-4}
& Memory Extraction Attacks & Stored agent memory & \keyref{wang2025unveiling} \\
\cmidrule{2-4}
& Prompt Injection Attacks & Private task context or traces & \keyref{greshake2023not}; \keyref{zhang2024privacyasst} \\
\bottomrule
\end{tabular}
}
\end{table*}

\paragraph{\textbf{Logits / Representations.}}
Logits and hidden representations are transmitted carriers in Distillation-based Transfer, SM$\rightarrow$LM (\S~\ref{subsubsec:s2l_distillation}), and Collaborative Decoding (\S~\ref{subsubsec:collaborative_decoding}). In these settings, data holders keep $\mathcal{D}_{target}$ local, but transmitted carriers can still expose private-domain information because they are computed from models trained on $\mathcal{D}_{target}$ or from private context. In Distillation-based Transfer, data holders release logits or hidden representations on public anchors, proxy queries, or server-side public data~\cite{li2019fedmd,gong2022preserving,chen2024datashunt,cheng2021fedgems,yumultimodal,fan2025fedmkt}; these carriers can encode class boundaries, response statistics, and distributional properties of $\mathcal{D}_{target}$. In Collaborative Decoding, the SM sends per-step next-token scores to the LM for fusion; high-scoring tokens may expose information about the private context~\cite{morris2024language,qu2025prompt}. We organize the attacks below by the collaboration setting in which logits or representations cross the trust boundary.

\par \textbf{Distillation-based Transfer} (\S~\ref{subsubsec:s2l_distillation}):
\begin{itemize}
    \item \textbf{Paired-Logits Inversion Attack.} This attack aims to reconstruct representative patterns of private classes rather than individual private samples. Public anchors form the carrier leakage path through paired client--server logits, from which private-class patterns can be recovered. The attack exploits discrepancies between responses released on the same public anchors, as these discrepancies encode class-level response patterns learned from the private training data~\cite{takahashi2023breaching,xiao2024privacy}.
    \item \textbf{Membership Inference Attacks.} Membership inference asks whether a candidate record was included in the private training set. Here, the carrier leakage path passes through confidence, loss, or teacher--student response patterns released on public anchors. These aggregate response signals can expose membership information when they remain statistically tied to the private training data, even when no private input is queried directly~\cite{yang2022fd,liu2023mia,shi2025unveiling}.
    \item \textbf{Property Inference Attacks.} Property inference targets corpus-level properties of $\mathcal{D}_{target}$, such as class distribution, group composition, or data domain. Aggregate response statistics on public anchors provide the carrier leakage path for estimating private-domain properties. Such statistics may correlate with distributional regularities learned from the private domain, especially when released responses remain tied to private-domain training data~\cite{ganju2018property,melis2019exploiting,salem2020updates,shi2025unveiling}.
    \item \textbf{Embedding Inversion Attacks.} Embedding inversion seeks to recover semantic content or attributes encoded in released representations. Embeddings from privately trained encoders, computed on public anchors, carry the attribute or content signal that defines the carrier leakage path. The attack exploits this encoded information: inversion methods such as decoder-based reconstruction and nearest-neighbor search can recover meaningful content from representations released on public anchor inputs~\cite{li2023sentence,huang2024transferable,staabbeyond}.
\end{itemize}
\par \textbf{Collaborative Decoding} (\S~\ref{subsubsec:collaborative_decoding}):
\begin{itemize}
    \item \textbf{Prompt Inversion Attacks.} Prompt inversion seeks to reconstruct private prompt or context information from inference-time signals observed by the LM. The carrier leakage path runs through SM-side next-token scores computed from the private context and prefix during LM-side fusion. The attack exploits high-probability token scores across decoding steps, which can expose private-context cues needed to reconstruct the hidden context~\cite{morris2024language,qu2025prompt}.
\end{itemize}


\paragraph{\textbf{Tokens.}}
Collaborative Decoding (\S~\ref{subsubsec:collaborative_decoding}) can also transmit token-level carriers, including draft or synchronized tokens, together with verification signals such as accept--reject decisions~\cite{leviathan2023fast,xia2024unlocking}. The accept--reject patterns and correction traces produced by these signals form side channels that may reveal private-context influence during decoding. 
\begin{itemize}
    \item \textbf{Side-Channel Attacks.} These attacks infer private-context influence from observable token-level verification signals rather than from the final semantic content alone. Token-level verification signals provide the carrier leakage path from private context or input to inferred private information. Existing work shows that acceptance, rejection, or correction patterns in speculative decoding can leak private-context influence~\cite{wei2024speculation,debenedetti2024privacy}.
\end{itemize}

\paragraph{\textbf{Synthetic Data.}}
In generation-based transfer (\S~\ref{subsubsec:s2l_generation}), synthetic data is released as a generated carrier derived from $\mathcal{D}_{target}$. Although raw private data are not transmitted, synthetic samples can preserve record-level signatures, hidden-attribute correlations, memorized spans, or distributional properties of $\mathcal{D}_{target}$~\cite{carlini2021extracting,nasr2023scalable,lindifferentially}. We summarize four representative risks below: membership inference, attribute inference, training data extraction, and property inference. 
\begin{itemize}
    \item \textbf{Membership Inference Attacks.} Membership inference tests whether a candidate record was included in $\mathcal{D}_{target}$. The released synthetic corpus provides the carrier leakage path from $\mathcal{D}_{target}$ to membership signals. The attack exploits record-specific signatures preserved in the released synthetic corpus, such as close synthetic neighbors, elevated surrogate likelihood, or overrepresented rare feature combinations~\cite{hayes2019logan,zhang2022membership,guepin2023synthetic}.
    \item \textbf{Attribute Inference Attacks.} Attribute inference estimates hidden sensitive attributes associated with private records or subgroups. Hidden attributes can leak through $\mathcal{D}_{syn}$ when the synthetic carrier preserves correlations from $\mathcal{D}_{target}$. The attack exploits feature correlations, linkability patterns, or reconstruction signals preserved in $\mathcal{D}_{syn}$, so sensitive attributes can be inferred without exact record reconstruction~\cite{stadler2022synthetic,annamalai2024linear,giomi2023unified}.
    \item \textbf{Training Data Extraction Attacks.} Training data extraction seeks to recover private content from the released synthetic corpus. Private content reaches the adversary through the released synthetic corpus when $\mathcal{D}_{syn}$ preserves memorized spans from $\mathcal{D}_{target}$. The attack exploits memorization by the generator, causing sensitive spans to appear in generated outputs~\cite{carlini2019secret,carlini2021extracting,carlini2023extracting,nasr2023scalable}. 
    \item \textbf{Property Inference Attacks.} Property inference estimates corpus-level properties of $\mathcal{D}_{target}$, such as class proportions, subgroup prevalence, rare property existence, or domain composition. The synthetic corpus forms the carrier leakage path for estimating private-domain properties inherited from $\mathcal{D}_{target}$. The attack exploits aggregate statistics or feature-level consistencies in the released synthetic corpus rather than reconstructing individual records~\cite{stadler2022synthetic,giomi2023unified,wang2024property}.
\end{itemize}

\paragraph{\textbf{Intermediate Outputs.}}
Intermediate outputs are the transmitted carriers in Split Learning (\S~\ref{subsubsec:split_execution}). Returned outputs and gradients are phase-dependent return carriers: inference returns task outputs, whereas training or split fine-tuning returns gradients, with gradients forming the main training-time leakage surface. The relevant carriers are forward cut-layer states and training-time gradients computed from private inputs, labels, or losses. These carriers may expose private inputs, sensitive attributes, or labels even when raw data remain local~\cite{zhu2019deep,he2019model,gao2023pcat}. We summarize four representative risks below: input reconstruction, attribute inference, label inference, and gradient inversion. 
\begin{itemize}
    \item \textbf{Input Reconstruction Attacks.} Input reconstruction recovers the private input that produced a transmitted cut-layer activation. The cut-layer state $z=f_c(x)$ carries the private input signal used for reconstruction. The attack exploits activation matching or decoder-based reconstruction, where the adversary searches for an input whose client-side representation matches the observed intermediate output~\cite{he2019model,erdougan2022unsplit,pasquini2021unleashing,gao2023pcat}.
    \item \textbf{Attribute Inference Attacks.} Attribute inference estimates sensitive attributes preserved by transmitted intermediate states without reconstructing the full input. Sensitive attributes can be inferred when the cut-layer state $z=f_c(x)$ preserves attribute-dependent information from the private input. The attack exploits attribute-dependent information retained in cut-layer activations or related intermediate carriers; split-learning inference attacks show that such carriers can expose private features even without sharing raw training data~\cite{pasquini2021unleashing,driouich2022attribute}.
    \item \textbf{Label Inference Attacks.} Label inference targets private labels that are not explicitly transmitted. Loss or gradient signals provide the carrier leakage path from the private label $y$ to the inferred label. The attack exploits cut-layer gradients, activation patterns, or similarity rules that vary systematically with candidate labels during split training or inference~\cite{erdougan2022unsplit,kariyappa2023exploit,liu2023distance,liu2024similarity}.
    \item \textbf{Gradient Inversion Attacks.} Gradient inversion recovers private training examples from transmitted backward signals. The gradient carrier $\nabla\ell$ exposes information from the private example $(x,y)$ that can be used to reconstruct the input or label. The attack exploits gradient matching: candidate examples are optimized until their induced gradients align with the observed gradient carrier~\cite{zhu2019deep,geiping2020inverting,dimitrov2024spear,petrov2024dager}.
\end{itemize}

\paragraph{\textbf{Adapters / Tunable Prompts.}}
Adapters and tunable prompts are parameter carriers in SM$\rightarrow$LM parameter-based transfer (\S~\ref{subsubsec:s2l_parameter}). Here, tunable prompts refer to learned prompt parameters, such as soft or prefix prompts, not natural-language prompts used in context-augmented collaboration. The data-privacy risk comes from releasing SM-side adapters, tunable prompts, or federated PEFT updates learned from $\mathcal{D}_{target}$; these carriers can leak training samples, training-set membership, or user-level participation~\cite{sami2025gradient, mireshghallah2022memorization, kandpal2024user}. We summarize three representative risks below: PEFT gradient inversion, membership inference, and user inference. 
\begin{itemize}
    \item \textbf{PEFT Gradient Inversion Attacks.} PEFT gradient inversion reconstructs private training examples from gradients or updates of lightweight PEFT modules. Shared PEFT updates or gradients provide the carrier leakage path from $\mathcal{D}_{target}$ to reconstructed inputs. The attack exploits gradient matching on adapter or tunable-prompt parameters, showing that local fine-tuning samples can leak even when the backbone model remains frozen~\cite{sami2025gradient}.
    \item \textbf{Membership Inference Attacks.} Membership inference tests whether a record was included in the private data used to train the released adapter or tunable prompt. Released adapter or tunable-prompt behavior provides the carrier leakage path from private training data to membership signals. The attack exploits behavioral differences in adapter or tunable-prompt modules between training members and non-members, which remain detectable through query access to the released module~\cite{mireshghallah2022memorization,ran2025loraleak,nakai2024does}.
    \item \textbf{User Inference Attacks.} User inference targets user-level contribution rather than record-level membership. Fine-tuned or adapted model behavior can carry user-level signals from local user data to inferred identity or participation. Existing user-inference attacks show that user-level contribution can be inferred from fine-tuned LMs using only a few user samples and model access~\cite{kandpal2024user}.
\end{itemize}

\paragraph{\textbf{Prompts / Outputs.}}
Context-Augmented Collaboration (\S~\ref{subsubsec:context_augmented_collaboration}) transmits natural-language prompts and output-level contextual carriers rather than token-level decoding traces. These carriers include user queries, retrieval requests, subtask instructions, retrieved evidence, tool outputs, execution summaries, and agent-generated feedback. Here, prompts denote natural-language requests or composed context, not tunable prompt parameters used in parameter-based transfer. Data leakage occurs when these carriers are grounded in $\mathcal{D}_{target}$, private retrieval memory, private tool state, or prior user--agent interactions and then cross the trust boundary as text-level outputs~\cite{rag-privacy,flemings2025estimating,wang2025unveiling,zhang2024privacyasst}. We summarize four representative risks below: RAG data extraction, RAG membership inference, memory extraction, and prompt injection attacks.
\begin{itemize}
    \item \textbf{RAG Data Extraction Attacks.} RAG data extraction recovers private content from retrieval-augmented systems. A request query can trigger retrieval from private memory $\mathcal{M}_{\mathrm{priv}}$, making the generated output the carrier leakage path for private content. The attack exploits the generated response, which may contain verbatim or near-verbatim private content retrieved from the SM-side knowledge base~\cite{rag-privacy,flemings2025estimating,qifollow,jiang2024rag}.
    \item \textbf{RAG Membership Inference Attacks.} RAG membership inference tests whether a target document or passage is contained in the private retrieval datastore. Retrieved context or generated answers can expose datastore membership signals in response to a target query. The attack exploits response similarity, answerability, or retrieval-grounded evidence that differs when the target document is present in $\mathcal{M}_{\mathrm{priv}}$~\cite{anderson2025my,li2025generating,naseh2025riddle}.
    \item \textbf{Memory Extraction Attacks.} Memory extraction recovers stored agent memory, demonstrations, or prior user--agent interactions. The request prompt can induce private memory access, with the response carrying the disclosed memory content. The attack exploits recall or summarization behavior that exposes historical context preserved in the agent state~\cite{wang2025unveiling}.
    \item \textbf{Prompt Injection Attacks.} Prompt injection can manipulate an agentic workflow so that private task context, retrieved evidence, tool results, credentials, or intermediate plans are returned outside the intended trust boundary. A malicious instruction can route tool or retrieval access toward output-side exfiltration of private task context or intermediate results. The attack exploits conflicting instructions between malicious inputs and system prompts in LLM-integrated applications, with sensitive data carried by returned traces rather than by parameters or activations~\cite{greshake2023not,debenedetti2024agentdojo,zhang2024privacyasst,ferrag2025prompt}.
\end{itemize}


\subsubsection{Carrier-Specific Model Attacks}\label{subsubsec:model_security_attacks}

The LM-side carriers $\mathcal{I}_{\pi}$ returned across the trust boundary act as observable proxies of proprietary LM behavior, even when LM parameters and internal states are never directly exposed~\cite{oliynyk2023know, zhao2025survey, hui2024pleak, sha2024prompt}, opening attack channels that do not exist in fully local LM serving~\cite{tramer2016stealing, orekondy2019knockoff, carlini2024stealing, wang2025stolenlora}. This subsection focuses on model-security leakage: LM-side carriers derived from $\theta_{\mathcal{L}}$ or the LM service boundary cross toward the SM, allowing proprietary behavior, task capability, ownership signals, or service-side prompts captured by $M_L$ (\S~\ref{sec:problem_definition}) to be inferred, reconstructed, reused, or weakened. The attacks below are organized by attack family, and each family retains the carrier-specific routes through which LM assets are exposed. Table~\ref{tab:carrier_model_attack_alignment} summarizes representative model-security attacks by carrier type, attack target, and key references.

\begin{table*}[t]
\centering
\caption{Representative model-security attacks on LM--SM collaborative carriers, organized by carrier type, attack target, and key references.}
\label{tab:carrier_model_attack_alignment}
\renewcommand{\arraystretch}{1.10}
\resizebox{\textwidth}{!}{
\begin{tabular}{l|l|l|l}
\toprule
\textbf{Transferred Carrier} & \textbf{Representative Attacks} & \textbf{Attack Target} & \textbf{Key References} \\
\midrule

\multirow{2}{*}{Student Model / Logits / Representations}
& Model Extraction Attacks & Decision behavior / task capability & \keyref{tramer2016stealing}; \keyref{carlini2024stealing} \\
\cmidrule{2-4}
& Watermark / Fingerprint Attacks & Ownership evidence / verification signal & \keyref{xu2024instructionalfingerprintinglargelanguage}; \keyref{pan2025can} \\
\midrule

\multirow{2}{*}{Synthetic Data}
& Model Extraction Attacks & Decision behavior / task capability & \keyref{krishna2019thieves}; \keyref{birch2023model} \\
\cmidrule{2-4}
& Watermark / Fingerprint Attacks & Ownership evidence / verification signal & \keyref{chen2025demark}; \keyref{hitaj2018have} \\
\midrule

\multirow{2}{*}{Compressed Weights / Adapters}
& Model Extraction Attacks & Task-specific adapter capability & \keyref{wang2025stolenlora}; \keyref{carlini2024stealing} \\
\cmidrule{2-4}
& Watermark / Fingerprint Attacks & Ownership evidence / verification signal & \keyref{zong2024ipremover}; \keyref{zhang2024large} \\
\midrule

Logits / Tokens
& Model Extraction Attacks & Token-level decoding behavior & \keyref{carlini2024stealing}; \keyref{oliynyk2023know} \\
\midrule

\multirow{3}{*}{Prompts / Outputs}
& Prompt Leaking Attacks & System prompt / hidden instructions & \keyref{hui2024pleak} \\
\cmidrule{2-4}
& Prompt Stealing Attacks & Prompt template / application logic & \keyref{sha2024prompt} \\
\cmidrule{2-4}
& Watermark / Fingerprint Attacks & Ownership evidence / verification signal & \keyref{chen2025demark}; \keyref{zong2024ipremover} \\
\bottomrule
\end{tabular}
}
\end{table*}

\paragraph{\textbf{Model Extraction Attacks.}}
Model extraction aims to reproduce proprietary LM task behavior in an unauthorized surrogate model~\cite{tramer2016stealing, orekondy2019knockoff}. In LM--SM collaboration, the extraction route depends on which carrier exposes LM behavior or task capability, and the items below follow these carrier routes.
\begin{itemize}
    \item \textbf{Student Model / Logits / Representations.} In distillation-based transfer, teacher logits, hidden representations, and the released student model carry task behavior across the trust boundary (\S~\ref{subsubsec:distillation-based-transfer-LM2SM})~\cite{hinton2015distilling, hsieh2023distilling}. In the surrogate-training route, teacher logits or representations serve as the carrier attack path for training a model that copies the proprietary task behavior; building on classical API-based extraction~\cite{tramer2016stealing, orekondy2019knockoff, krishna2019thieves, carlini2024stealing}, the carrier-specific distillation variant exploits teacher logits or representations on anchor inputs to train an unauthorized surrogate that reproduces teacher predictions~\cite{birch2023model, oliynyk2023know, zhao2025survey}. In the direct-reuse route, the released student model itself becomes the carrier attack path for unauthorized reuse of the copied task behavior; the adversary may directly redeploy it, use it as a teacher for further unauthorized distillation~\cite{pan2025can}, or fine-tune it to evade ownership verification~\cite{xu2024instructionalfingerprintinglargelanguage}.
    \item \textbf{Synthetic Data.} In LM$\rightarrow$SM generation-based transfer (\S~\ref{subsubsec:generation-based-transfer-LM2SM}), synthetic data is the transmitted carrier through which the LM encodes its task knowledge into $\mathcal{D}_{syn}$~\cite{ye2022zerogen, meng2022generating, liu2023retrieval}. Since $\mathcal{D}_{syn}$ consists of input--output pairs generated or labeled by the LM, it can serve as supervision for training an unauthorized surrogate without access to LM parameters or logits, and repeated generation further expands the input--output coverage available for behavior approximation~\cite{krishna2019thieves, birch2023model, carlini2024stealing}.
    \item \textbf{Compressed Weights / Adapters.} In LM$\rightarrow$SM parameter-based transfer (\S~\ref{subsubsec:parameter-based-transfer-LM2SM}), compressed weights and adapters are reusable parameterized carriers~\cite{hu2021lora, fan2023fatellm, peng2024fedpft}. Unlike transient logits or outputs, these carriers can be copied and redeployed directly, making ownership and unauthorized reuse central model-security concerns~\cite{wang2025stolenlora}. LoRA extraction reproduces adapter task capability: the released module provides the carrier attack path for direct redeployment or downstream distillation of the stolen capability~\cite{wang2025stolenlora, carlini2024stealing}.
    \item \textbf{Tokens / Transmitted Logits.} In collaborative decoding (\S~\ref{subsubsec:collaborative_decoding}), transmitted LM logits carry token-prediction behavior across decoding steps, and token-wise carriers include verified or accepted tokens exchanged during decoding~\cite{zhang2024cogenesis,thareja2026dp,leviathan2023fast,xia2024unlocking}. LM logits collected across decoding steps provide a soft-label carrier path for surrogate training because they reveal token-ranking behavior beyond observed token choices~\cite{carlini2024stealing, oliynyk2023know, zhao2025survey}. Although a hard-label trace exposes less per-step information than a logit vector, repeated decoding interactions can accumulate accepted-token patterns that support coarse approximation of LM decoding behavior~\cite{carlini2024stealing}.
\end{itemize}

\paragraph{\textbf{Prompt Leaking and Prompt Stealing Attacks.}}
Prompt leaking and prompt stealing target the hidden system prompts, prompt templates, and application logic embedded in an LM-side service~\cite{hui2024pleak, sha2024prompt}. In context-augmented collaboration (\S~\ref{subsubsec:context_augmented_collaboration}), the exchanged prompts and outputs provide the carrier route for this exposure. The risk is model-side rather than data-side: what leaks is proprietary application behavior, not user private data.
\begin{itemize}
    \item \textbf{Prompt Leaking Attacks.} Prompt leaking discloses hidden system prompts, templates, or policy logic when the application directly emits them in generated outputs or interaction traces. Interaction prompts or generated outputs provide the carrier attack path when they disclose hidden prompts and expose application-side logic. The attack exploits the tendency of LMs to reproduce system prompt content in outputs, often triggered through adversarial or indirect extraction queries~\cite{hui2024pleak}.
    \item \textbf{Prompt Stealing Attacks.} Prompt stealing aims to reconstruct hidden prompt templates and application logic through repeated interaction, even when the prompt is never directly emitted. Repeated interactions provide the carrier attack path for inferring prompt templates and copying application-side behavior. The attack exploits consistent behavioral patterns across outputs (such as response format, instruction-following style, or templated phrasing) to infer the underlying prompt structure~\cite{sha2024prompt}.
\end{itemize}

\paragraph{\textbf{Watermark / Fingerprint Attacks.}}
Watermarking and fingerprinting attach ownership or verification signals to released carriers (\S~\ref{sec:privacy_defenses}); watermark and fingerprint attacks target these signals rather than carrier task content~\cite{chen2025demark, zong2024ipremover}. Because protected carriers range from student models and synthetic data to compressed weights, adapters, and outputs, these attacks cross-cut multiple rows of Table~\ref{tab:carrier_model_attack_alignment} and are grouped here by attack mechanism.
\begin{itemize}
    \item \textbf{Watermark Removal / Evasion Attacks.} These attacks remove or weaken ownership marks while preserving carrier task utility. Watermarked carriers or generated outputs provide the carrier attack path when fine-tuning, paraphrasing, or token substitution weakens the signal needed for ownership tracing. The attack exploits the fact that watermark signals can often be weakened while preserving task utility, using fine-tuning, smoothing, or substitution operations~\cite{chen2025demark, wu2024bypassing, diaa2025optimizing, chang2025watermark, hitaj2018have}.
    \item \textbf{Watermark Stealing / Fingerprint Inversion Attacks.} These attacks infer or reconstruct the hidden ownership signal used by a verification scheme, enabling forgery or ambiguous ownership claims. Protected carriers provide the carrier attack path for recovering watermark or fingerprint signals and compromising ownership verification~\cite{zong2024ipremover,zhang2024large}. The attack exploits query access, detector feedback, or available protected carriers to recover the verification trigger or watermark pattern, allowing the adversary to forge or dispute ownership~\cite{jovanovic2024watermark,sadasivan2025can}.
\end{itemize}


\subsubsection{Carrier-Specific Integrity and Robustness Attacks}
\label{subsubsec:carrier_integrity_attacks}

Transferred carriers pose integrity risks to the receiving party. LM$\rightarrow$SM carriers $\mathcal{I}_{\pi}$ and SM$\rightarrow$LM carriers $\mathcal{J}_{\pi}$ can be adversarially manipulated before they are used by the receiving party. A manipulated carrier can lead the collaborative workflow to rely on poisoned supervision, backdoored parameter updates, corrupted retrieval context, or manipulated intermediate states~\cite{greshake2023not,debenedetti2024agentdojo,zou2025poisonedrag,chen2024agentpoison,wan2023poisoning,liu2025loratk}. This subsection focuses on integrity and robustness failures captured by $M_r$ (Eq.~\eqref{eq:problem}), where the harm is not information leakage but a deviation in training, adaptation, retrieval, or inference behavior at the receiving party. Table~\ref{tab:carrier_integrity_attack_alignment} summarizes representative integrity and robustness attacks by carrier type, attack target, and key references.

\begin{table*}[t]
\centering
\caption{Representative integrity and robustness attacks on LM--SM collaborative carriers, organized by carrier type, attack target, and key references.}
\label{tab:carrier_integrity_attack_alignment}
\renewcommand{\arraystretch}{1.10}
\resizebox{\textwidth}{!}{
\begin{tabular}{l|l|l|l}
\toprule
\textbf{Transferred Carrier} & \textbf{Representative Attacks} & \textbf{Attack Target} & \textbf{Key References} \\
\midrule

\multirow{4}{*}{Logits / Representations / Intermediate Outputs}
& \multirow{2}{*}{Backdoor Attacks} & Triggered student behavior & \keyref{cheng2024transferring}; \keyref{de2025pay} \\
\cmidrule{3-4}
& & Triggered split-model behavior & \keyref{bai2023villain} \\
\cmidrule{2-4}
& Training-Hijacking Attacks & Client-side training objective & \keyref{pasquini2021unleashing} \\
\cmidrule{2-4}
& Trojan Activation Attacks & Activation-steered generation & \keyref{wang2024trojan}; \keyref{li2025backdoorllm} \\
\midrule

\multirow{2}{*}{Synthetic Data}
& Data Poisoning Attacks & Adapted model behavior & \keyref{wan2023poisoning}; \keyref{bowen2025scaling} \\
\cmidrule{2-4}
& Backdoor Attacks & Triggered instruction-following behavior & \keyref{xu2024instructions}; \keyref{yan2024backdooring} \\
\midrule

\multirow{2}{*}{Adapters / Tunable Prompts}
& PEFT Backdoor Attacks & Triggered adapter or prompt behavior & \keyref{liu2025loratk}; \keyref{sun2025peftguard} \\
\cmidrule{2-4}
& Model Poisoning Attacks & Corrupted aggregate model behavior & \keyref{bagdasaryan2020backdoor}; \keyref{fang2020local} \\
\midrule

\multirow{4}{*}{Prompts / Outputs}
& Prompt Injection Attacks & Workflow control and tool use & \keyref{greshake2023not}; \keyref{debenedetti2024agentdojo} \\
\cmidrule{2-4}
& Tool-Selection Hijacking & Tool selection or workflow routing & \keyref{shi2025prompt}; \keyref{sneh2025tooltweak} \\
\cmidrule{2-4}
& RAG and Memory Poisoning Attacks & Retrieved evidence and agent memory & \keyref{zou2025poisonedrag}; \keyref{chen2024agentpoison} \\
\cmidrule{2-4}
& Backdoor Attacks & Context-conditioned reasoning or prediction & \keyref{zhao2024universal}; \keyref{xiang2024badchain} \\
\bottomrule
\end{tabular}
}
\end{table*}

\paragraph{\textbf{Logits / Representations / Intermediate Outputs.}}
In distillation-based transfer (\S~\ref{subsubsec:distillation-based-transfer-LM2SM}, \S~\ref{subsubsec:s2l_distillation}), logits and representations carry teacher supervision across the trust boundary; in split learning (\S~\ref{subsubsec:split_execution}), intermediate outputs carry internal computation states between the participating models. These carriers can transmit corrupted supervision, manipulated states, or forged training signals. They can therefore transfer backdoor behavior, hijack client-side training, or steer downstream generation and prediction behavior~\cite{cheng2024transferring,de2025pay,bai2023villain,pasquini2021unleashing,wang2024trojan,li2025backdoorllm}.
\begin{itemize}
    \item \textbf{Backdoor Attacks.} Backdoor attacks aim to insert trigger-dependent behavior into the downstream model through transferred supervision or split-learning interaction. In distillation, teacher logits or representations provide the inbound carrier path from a poisoned teacher to the student. In split learning, malicious client-side carriers provide a path for client-side backdoors to enter the collaborative model. The attack exploits training on carriers that appear clean, allowing hidden trigger behavior to be learned from poisoned logits, representations, or client-controlled intermediate signals~\cite{cheng2024transferring,de2025pay,bai2023villain}.
    \item \textbf{Training-Hijacking Attacks.} Training-hijacking attacks aim to replace the client model training objective by forging the returned training signal. In split learning, the returned gradient carrier provides the inbound carrier path from a malicious server to the client-side model. The attack exploits limited client ability to validate received gradients: forged gradients steer the client model toward the attacker objective, creating an integrity deviation that later enables reconstruction of private inputs~\cite{pasquini2021unleashing}.
    \item \textbf{Trojan Activation Attacks.} Trojan activation attacks aim to induce targeted generation behavior by manipulating internal activation directions. Manipulated activation states provide the inbound carrier path in collaborations that exchange intermediate computation states. The attack exploits activation-level steering, where an attacker-controlled direction in the model internal state shifts generation toward attacker-specified behavior~\cite{wang2024trojan,li2025backdoorllm}.
\end{itemize}

\paragraph{\textbf{Synthetic Data.}}
In LM$\rightarrow$SM and SM$\rightarrow$LM generation-based transfer (\S~\ref{subsubsec:generation-based-transfer-LM2SM}, \S~\ref{subsubsec:s2l_generation}), synthetic data carries task supervision or domain guidance across the trust boundary. The receiving party may use this carrier for training, adaptation, or instruction tuning, so poisoned or backdoored examples can be treated as ordinary supervision and change adapted model behavior~\cite{wan2023poisoning,bowen2025scaling,xu2024instructions,yan2024backdooring,li2025backdoorllm}.
\begin{itemize}
    \item \textbf{Data Poisoning Attacks.} Data poisoning attacks aim to corrupt downstream adaptation by inserting malicious examples into synthetic or instruction-tuning data. The poisoned dataset provides the inbound carrier path from the generator or data provider to the downstream model. The attack exploits the appearance of poisoned examples as ordinary supervision and can shift the adapted model toward incorrect or attacker-chosen behavior after training~\cite{wan2023poisoning,bowen2025scaling,li2025backdoorllm}.
    \item \textbf{Backdoor Attacks.} Backdoor attacks aim to insert trigger-dependent behavior through synthetic instruction carriers. Poisoned instruction-tuning examples provide the inbound carrier path, with the instruction field or response field carrying the trigger and target behavior. The attack exploits instruction tuning to bind malicious behavior to a trigger; virtual prompt injection follows this route by making the adapted model behave as if an attacker-specified hidden prompt were appended under the trigger condition~\cite{xu2024instructions,yan2024backdooring}.
\end{itemize}

\paragraph{\textbf{Adapters / Tunable Prompts.}}
In parameter-based transfer (\S~\ref{subsubsec:parameter-based-transfer-LM2SM}, \S~\ref{subsubsec:s2l_parameter}), adapters and tunable prompts carry compact parameter updates or learned prompt parameters across the trust boundary. After loading or aggregation, these carriers can modify model behavior without changing the full backbone. Shared PEFT modules, LoRA weights, learned prompt parameters, and federated parameter updates therefore act as attack surfaces for backdoor injection or model poisoning~\cite{liu2025loratk,sun2025peftguard,yao2024poisonprompt,li2025backdoorllm,bagdasaryan2020backdoor,fang2020local}.
\begin{itemize}
    \item \textbf{PEFT Backdoor Attacks.} PEFT backdoor attacks aim to insert trigger-dependent behavior through parameterized carriers loaded by the downstream model. Adapter carriers such as LoRA modules and PEFT updates, together with learned prompt parameters, provide the inbound carrier path from a malicious provider to the downstream model. The attack exploits the small and modular form of these carriers, keeping benign behavior on clean inputs while activating attacker-specified behavior after the carrier is attached to the model~\cite{liu2025loratk,sun2025peftguard,yao2024poisonprompt,li2025backdoorllm}.
    \item \textbf{Model Poisoning Attacks.} Model poisoning attacks aim to corrupt the aggregated model by manipulating uploaded parameter carriers in multi-party collaboration (\S~\ref{sec:multi-party}). Poisoned client updates, including malicious adapters or PEFT updates, provide the inbound carrier path from compromised clients to the shared model. The attack exploits the limited ability of the aggregation step to distinguish malicious updates from heterogeneous benign updates~\cite{bagdasaryan2020backdoor,fang2020local}.
\end{itemize}

\paragraph{\textbf{Prompts / Outputs.}}
In context-augmented collaboration (\S~\ref{subsubsec:context_augmented_collaboration}), prompts and outputs carry textual requests, retrieved evidence, tool descriptions, tool metadata, tool outputs, agent memory, demonstrations, and reasoning traces across the trust boundary. These carriers are used directly at inference time, so malicious text or retrieved content can redirect tool use, poison retrieved evidence or memory, or trigger context-conditioned behavior~\cite{greshake2023not,debenedetti2024agentdojo,zou2025poisonedrag,xue2024badrag,cheng2024trojanrag,chen2024agentpoison,zhao2024universal,xiang2024badchain}.
\begin{itemize}
    \item \textbf{Prompt Injection Attacks.} Prompt injection attacks aim to override constraints or redirect an LM-integrated workflow through malicious instructions in an inbound text carrier. Web pages, retrieved documents, tool outputs, and user-provided prompts provide the inbound carrier path from untrusted text to the assembled context. The attack exploits instruction conflicts between malicious content and system or developer prompts, which can lead the LM to ignore constraints, call unintended tools, or leak information through later outputs~\cite{greshake2023not,debenedetti2024agentdojo}.
    \item \textbf{Tool-Selection Hijacking.} Tool-selection hijacking attacks aim to manipulate the SM-side tool interface by poisoning tool descriptions, tool metadata, or tool outputs. These carriers can cause the collaborative workflow to select attacker-controlled tools, invoke unintended tools, or redirect subsequent reasoning and actions. The attack exploits tool selection driven by text carriers that are inserted into the inference context or used by the SM before interaction with the LM~\cite{shi2025prompt,sneh2025tooltweak}.
    \item \textbf{RAG and Memory Poisoning Attacks.} RAG and memory poisoning attacks aim to corrupt external context that the LM later retrieves or reuses. Poisoned passages, memory entries, or knowledge-base records provide the inbound carrier path from a compromised corpus or agent state to the generated answer. The attack exploits retrieval or memory reuse, steering responses toward attacker-chosen claims, incorrect facts, harmful tool use, or persistent behavioral changes across turns~\cite{zou2025poisonedrag,xue2024badrag,chen2024agentpoison}.
    \item \textbf{Backdoor Attacks.} Backdoor attacks aim to insert trigger-dependent behavior into inference-time context carriers. Poisoned demonstrations, retrieved examples, or malicious reasoning traces provide the inbound carrier path from context construction to the final prediction or response. The attack exploits trigger-conditioned patterns in in-context learning, RAG, or chain-of-thought prompting, inducing attacker-specified predictions or conclusions after the trigger appears~\cite{zhao2024universal,cheng2024trojanrag,xiang2024badchain}.
\end{itemize}

\subsubsection{Protection Techniques}
\label{sec:privacy_defenses}

\begin{table*}[t]
\centering
\small
\setlength{\tabcolsep}{4pt}
\renewcommand{\arraystretch}{1.00}
\caption{A carrier-aligned summary of defense strategies for LM--SM collaboration, organized by protection target, defense strategy, protected carrier, and key references.}
\label{tab:mitigation_landscape}
\resizebox{\textwidth}{!}{%
\begin{tabular}{l|l|l|l}
\toprule
\textbf{Protection Target} & \textbf{Defense Strategy} & \textbf{Protected Carrier} & \textbf{Key References} \\
\midrule
\multirow{7}{*}{\textbf{Data Privacy ($M_p$)}}
& \multirow{2}{*}{\textbf{Differential Privacy (DP)}}
& Logits / Representations; Intermediate Outputs; Synthetic Data
& \multirow{2}{*}{\shortstack[l]{\keyref{dwork2006calibrating};\\ \keyref{shen2023split}}} \\
& & Adapters / Tunable Prompts; Prompts & \\
\cmidrule{2-4}
& \textbf{Adversarial Training}
& Logits / Representations; Intermediate Outputs
& \keyref{song2020information}; \keyref{xhonneux2024efficient} \\
\cmidrule{2-4}
& \textbf{Privacy-Preserving Representation Learning}
& Logits / Representations; Intermediate Outputs
& \keyref{duan2024reimagining}; \keyref{noorbakhsh2024inf2guard} \\
\cmidrule{2-4}
& \textbf{Secure Computation}
& Intermediate Outputs; Prompts
& \keyref{dong2023pumasecureinferencellama7b}; \keyref{de2025encryptedllm} \\
\cmidrule{2-4}
& \textbf{Machine Unlearning}
& Synthetic Data; Adapters / Tunable Prompts
& \keyref{jang-etal-2023-knowledge}; \keyref{yao2024machineunlearning} \\
\cmidrule{2-4}
& \textbf{Text Sanitization}
& Prompts / Outputs
& \keyref{edemacu2025privacy}; \keyref{chong2024casper} \\
\midrule
\multirow{8}{*}{\textbf{Model Security ($M_L$)}}
& \multirow{2}{*}{\textbf{Watermarking}}
& Student Model; Compressed Weights / Adapters
& \multirow{2}{*}{\shortstack[l]{\keyref{li2023watermarking};\\ \keyref{kirchenbauer2023watermark}}} \\
& & Synthetic Data; Outputs & \\
\cmidrule{2-4}
& \multirow{2}{*}{\textbf{Fingerprinting}}
& Student Model / Logits / Representations
& \multirow{2}{*}{\shortstack[l]{\keyref{xu2024instructionalfingerprintinglargelanguage};\\ \keyref{iourovitski2024hide}}} \\
& & Compressed Weights / Adapters; Outputs & \\
\cmidrule{2-4}
& \textbf{Weight Obfuscation}
& Compressed Weights / Adapters
& \keyref{wang2024taylor} \\
\cmidrule{2-4}
& \multirow{2}{*}{\textbf{Model Extraction Defenses}}
& Student Model / Logits / Representations; Synthetic Data
& \multirow{2}{*}{\shortstack[l]{\keyref{juuti2019prada};\\ \keyref{tang2024modelguard}}} \\
& & Tokens; Outputs & \\
\cmidrule{2-4}
& \textbf{Prompt Obfuscation / Guardrails}
& Prompts / Outputs
& \keyref{pape2025prompt}; \keyref{dong2024guardrails} \\
\midrule
\multirow{5}{*}{\textbf{Integrity / Robustness ($M_r$)}}
& \textbf{Robust Aggregation}
& Compressed Weights / Adapters; Logits / Representations
& \keyref{wan2022shielding}; \keyref{khedekar2025sybil} \\
\cmidrule{2-4}
& \textbf{Representation-level Defenses}
& Logits / Representations; Intermediate Outputs
& \keyref{yang2022not}; \keyref{khan2024make} \\
\cmidrule{2-4}
& \textbf{LLM Guardrails}
& Prompts / Outputs
& \keyref{banerjee2025safeinfer}; \keyref{xu2024safedecoding} \\
\cmidrule{2-4}
& \multirow{2}{*}{\textbf{Backdoor Detection}}
& Synthetic Data; Compressed Weights / Adapters
& \multirow{2}{*}{\shortstack[l]{\keyref{sun2025peftguard};\\ \keyref{rieger2025safesplit}}} \\
& & Intermediate Outputs; Prompts / Outputs & \\
\bottomrule
\end{tabular}%
}
\end{table*}

\par

SM-side carriers can expose private-domain information, the data-privacy risk measured by $M_p$ (\S~\ref{subsubsec:carrier_data_privacy_attacks}), and LM-side carriers can expose proprietary model internals, the model-security risk measured by $M_L$ (\S~\ref{subsubsec:model_security_attacks}). Carriers received in either direction can corrupt downstream collaborative behavior, the integrity risk measured by $M_r$ (\S~\ref{subsubsec:carrier_integrity_attacks}).
We therefore organize protection techniques into three families, each addressing one risk class: \textbf{Data Privacy} transforms or constrains outbound SM-side carriers to bound $M_p$, \textbf{Model Security} protects the assets carried by outbound LM-side carriers under $M_L$, and \textbf{Integrity and Robustness} defenses validate or constrain inbound carriers to bound $M_r$ before they affect downstream collaborative behavior. Table~\ref{tab:mitigation_landscape} summarizes these mechanisms by protection goal, protected carrier, and key references.

\textbf{Data Privacy.} Following the data-privacy attack routes in \S~\ref{subsubsec:carrier_data_privacy_attacks}, data-privacy defenses aim to reduce the private-domain information exposed by outbound SM-side carriers, by transforming, constraining, or hiding each carrier before it crosses the trust boundary~\cite{zhu2019deep, song2020information, duan2023privacy, rag-privacy}. Existing techniques include:

\begin{itemize}
    \item \textbf{Differential Privacy (DP)}. DP \cite{dwork2006calibrating} works by adding carefully calibrated noise in a way that the outcome is not significantly affected by the presence or absence of any single individual. DP has been the most common technique for protecting private training data \cite{abadi2016deep}, intermediate results \cite{plant-etal-2021-cape-noise,lyu-etal-2020-differentially,lyu2020towards,DP-bert2021,shen2023split,chen-etal-2023-customizedText}, tunable prompts \cite{li2023privacypreserving, promptpate_2023,hong2024dpopt} and retrieved information \cite{cheng2025remoterag,koga2024privacypreservingretrievalaugmentedgeneration}. In LM-SM collaboration, DP can be applied to various stages, from raw private data to the transmitted carriers between LMs and SMs. For example, in split learning, where intermediate activations are exchanged between LMs and SMs, DP noise can be added to these activations to prevent reconstruction attacks.~\cite{plant-etal-2021-cape-noise, lyu-etal-2020-differentially, shen2023split}. However, its utility degradation and privacy accounting under repeated or multi-carrier collaboration remain limiting factors~\cite{dwork2014algorithmic, mironov2017renyi}.

    \item \textbf{Adversarial Training}. Adversarial training \cite{li-etal-2018-towards,coavoux2018privacy,xhonneux2024efficient,song2020information} enhances model robustness by training the model to withstand adversarial examples. In SM-LM collaboration, adversarial training can be used to train SMs that are less susceptible to privacy attacks. Adversarial training can also make the transferred information (e.g., prompts, gradients) robust to adversarial perturbations, therefore less informative to potential attackers.

    \item \textbf{Information Reduction.} Privacy-preserving representation learning reduces recoverable private information in shared logits, representations, and intermediate outputs while preserving task utility~\cite{wang2021improving,duan2024reimagining,noorbakhsh2024inf2guard}. Typical mechanisms regularize the transmitted signal, constrain its mutual information with private inputs, or train it against inference and reconstruction attacks. Recent model-inversion defenses also use trapdoor signals to mislead reconstruction attacks rather than only reducing information leakage~\cite{liu2024trap}. In LM--SM collaboration, this route is most relevant when the transmitted carrier is a latent or prediction-side signal rather than raw text.

    \item \textbf{Machine unlearning.} Unlearning \cite{right_to_be_forget_Shintre2019MakingML,jang-etal-2023-knowledge,huang2024offset,yao2024machineunlearning} focus on removing the influence of specific data points or concepts from a trained model. In SM-LM collaboration, unlearning techniques could be applied locally on the SMs to remove traces of highly private data.
    However, it cannot retract carriers that have already crossed the trust boundary or been absorbed by another party.

    \item \textbf{Text Sanitization and Obfuscation.} Text sanitization \cite{edemacu2025privacy,pilan2022text,chen2023hide,kan2023protecting,chong2024casper} and obfuscation techniques \cite{zhou-etal-2022-textfusion,zhou2023textobfuscator,yao2024privacy,zhang2023latticegen} both operate on input text for preventing data attacks on large language models. They can also be used for removing or obfuscating sensitive information transmitted from SM to LM.
    They remove identifiers, sensitive spans, or revealing paraphrases from user queries and retrieved evidence, but the same rewriting can reduce semantic fidelity and downstream utility.

    \item \textbf{Secure Computation.} Homomorphic Encryption \cite{jin2024fedmlheefficienthomomorphicencryptionbasedprivacypreserving,chen-etal-2022-x,de2025encryptedllm}, Multiparty computation (MPC) \cite{limpcformer,li2024nimbus,dong2023pumasecureinferencellama7b}, and Trusted Execution Environment (TEE) ~\cite{frikha2024obfuscatune} have also been explored for protecting intermediate information and prompts, but their communication and computation costs still grow sharply with model size, sequence length, and interaction frequency.
    
\end{itemize}

\textbf{Model Security.} Following the model-security attack routes in \S~\ref{subsubsec:model_security_attacks}, model-security defenses aim to preserve model ownership, limit unauthorized behavior cloning, and audit whether released carriers have been copied or reused outside the intended trust boundary. In LM--SM collaboration, this is most relevant for prediction-side carriers, distilled student or proxy models, compressed weights, adapters, tunable prompts, and output-side behavior. Existing techniques include:
\begin{itemize}
    \item \textbf{Watermarking and Model Fingerprinting.} Watermarking embeds verifiable ownership signals into released artifacts or generated content, while fingerprinting tests whether a suspicious model reproduces protected behavior~\cite{li2023watermarking, kirchenbauer2023watermark, xu2024instructionalfingerprintinglargelanguage, iourovitski2024hide}. These techniques support $M_L$ auditing after a parameterized, generated, or output-side carrier has been copied or reused.
    \item \textbf{Weight-space Transformation / Obfuscation.} This route transforms released parameterized artifacts before sharing, rather than relying on a single standardized primitive~\cite{wang2024taylor}. It targets transmitted compressed weights or adapters, but the transformation may affect deployability or downstream adaptation.
    \item \textbf{Model Extraction Defenses.} This term refers to prediction perturbation, query-pattern detection, and query-cost shaping methods rather than a single canonical primitive~\cite{ma2021nasty, juuti2019prada, dziedzic2022increasing, tang2024modelguard}. These defenses limit repeated prediction-side or output-side carrier access, but they must preserve enough signal for legitimate collaboration.
    \item \textbf{Prompt Obfuscation / Guardrails.} Prompt obfuscation and guardrails aim to prevent hidden prompts, templates, and application logic from being disclosed or reconstructed through outputs or repeated interactions~\cite{pape2025prompt,dong2024guardrails,hui2024pleak,sha2024prompt}. They are most relevant to Prompts / Outputs in context-augmented collaboration, but they remain heuristic and may be bypassed by adaptive prompt leaking or stealing attacks.
\end{itemize}

\textbf{Integrity and Robustness.}
Following the integrity attack routes in \S~\ref{subsubsec:carrier_integrity_attacks}, integrity and robustness defenses aim to preserve downstream collaborative behavior by validating or constraining received carriers before they affect model state, retrieval context, or generated output at the receiving collaborator. Existing techniques include:
\begin{itemize}
    \item \textbf{Robust Aggregation.} Robust aggregation is most relevant in multi-party or repeated aggregation settings, where the system must filter poisoned client carriers before updating a shared model or decision rule~\cite{wan2022shielding, sheikhi2025hybrid, chenrobust, khedekar2025sybil}.
    \item \textbf{Backdoor Detection.} Backdoor detection identifies hidden triggers or malicious behaviors embedded in received carriers before they are consumed by the collaborator~\cite{hou2024ibd, goldblum2022dataset, dong2021black, sun2025peftguard, rieger2025safesplit}.
    \item \textbf{Representation-level Defenses.} This family regularizes, denoises, filters, or hardens shared latent and prediction carriers, especially logits, representations, and intermediate outputs~\cite{yang2022not, chen2023tutorial, wang2022improved, goldblum2022dataset, khan2024make}.
    \item \textbf{LLM Guardrails.} LLM guardrails apply filtering, policy checking, and safety-aligned decoding to Prompts and Outputs in context-augmented collaboration~\cite{dong2024guardrails, banerjee2025safeinfer, robeysmoothllm, xu2024safedecoding}. Their role is to constrain harmful or manipulative textual carriers without turning the discussion into a general LLM safety problem.
\end{itemize}

Table~\ref{tab:mitigation_landscape} summarizes the mitigation landscape and aligns each defense family with the carrier types it most directly protects.

In summary, no single defense uniformly protects all collaboration paradigms. Distillation-based transfer benefits from output regularization and model extraction defenses; Split Learning requires intermediate-output protection; context-augmented collaboration demands prompt sanitization or obfuscation, retrieval privacy, output guardrails, and trajectory-level monitoring; and parameter-based transfer requires ownership-preserving defenses for compressed weights, adapters, and tunable prompts. The open challenge is therefore not merely to adopt a stronger primitive, but to match the defense stack to the exact carrier exposed across the trust boundary.

\paragraph{\textbf{Open Challenges.}}
The preceding analysis exposes fundamental gaps in the current protection landscape that reflect misalignments between how current defenses are designed and how LM--SM collaboration actually operates.
\begin{itemize}
    \item \textbf{Carrier abstraction does not lead to measurable privacy reduction.} A tacit assumption shared across all carrier types is that replacing raw private data with an abstract carrier constitutes an implicit privacy gain. This conflates the \emph{form} of transmitted information with its \emph{information content}: logits preserve the decision-boundary geometry of the private distribution and enable membership inference even without raw data access~\cite{melis2019exploiting, nasr2019comprehensive, shokri2017membership, fu2024membership}; synthetic data inherit distributional statistics and memorized traces from the source domain~\cite{carlini2021extracting, lindifferentially, takahashi2023breaching}; and representations encode attribute structure that can be recovered by inversion attacks~\cite{li2023sentence, huang2024transferable, fredrikson2015model, xiao2024privacy}. The open question is not which carrier type is chosen, but how much private information the chosen carrier makes geometrically recoverable, a quantity that current defenses neither measure nor bound at the carrier level.

    \item \textbf{Defenses cannot retroactively cross the trust boundary.} All defense techniques surveyed in Section~\ref{sec:privacy_defenses} are unilateral operations applied on the originating side. Once a carrier crosses the trust boundary and is absorbed into LM weights or context~\cite{shen2023split, song2020information}, no SM-side operation can retract the propagated knowledge. Machine unlearning is paradigmatic: the SM can remove the influence of a sensitive instance from its own parameters, but the effect of that instance has already entered the LM via synthetic data, distilled logits or representations, compressed weights, or adapters~\cite{jang-etal-2023-knowledge, yao2024machineunlearning, huang2024offset}. The LM never directly labeled what it received as private, so it cannot identify specific targets to forget even when trained data can be extracted from it post-hoc~\cite{carlini2021extracting, nasr2023scalable, kandpal2024user}. This asymmetry makes right-to-be-forgotten guarantees difficult to enforce after carrier transfer in LM--SM collaboration~\cite{right_to_be_forget_Shintre2019MakingML}.

    \item \textbf{Multi-carrier deployments invalidate single-paradigm privacy accounting.} Each paradigm in this survey is analyzed with a single active carrier, yet agentic deployments routinely activate multiple carrier channels within a single session: intermediate outputs, prompts, outputs, and logits or tokens may all draw on the same private corpus simultaneously~\cite{shen2023split, rag-privacy, flemings2025estimating}. Differential privacy composition theorems require that composed mechanisms operate on statistically independent inputs~\cite{dwork2006calibrating, abadi2016deep, dwork2014algorithmic, bun2016concentrated}. This independence assumption is challenged when multiple carriers share a private source, as in generation-based and context-augmented paradigms that query the same local knowledge base~\cite{lindifferentially, duan2023privacy}. Compounding this, the LM pretraining distribution may already overlap with the private domain, causing the effective privacy guarantee of DP fine-tuning to be systematically overestimated~\cite{kandpal2023large, carlini2021extracting}. Privacy accounting for multi-carrier LM--SM collaboration therefore remains underdeveloped.

    \item \textbf{Single-step defenses provide limited guarantees against multi-step adversaries.} Many defenses reviewed above are evaluated under single-step or fixed-interaction assumptions, such as one query, one response, or one enforcement point. In multi-step agentic workflows, an adversary can instead accumulate sub-threshold signals across successive turns, each individually evading detection, and reconstruct private content from the aggregate~\cite{zhang2024privacyasst, hui2024pleak, flemings2025estimating, greshake2023not}. Attack strategies such as indirect prompt injection, adversarial RAG poisoning, and cross-agent manipulation further demonstrate that the relevant threat in agentic deployment is a persistent, session-level adversary rather than a one-shot query~\cite{chen2024agentpoison, zou2025poisonedrag, amayuelas2024multiagent}. The surveyed defenses therefore provide limited guarantees under adaptive multi-round interaction.
\end{itemize}

\subsection{Efficiency Considerations}
\label{sec:efficiency_considerations}

\begin{table*}[t]
\centering
\caption{Carrier-specific efficiency profiles in LM--SM collaboration, mapping each transferred carrier to its characteristic burden on communication ($\epsilon_e^{(comm)}$), local computation ($\epsilon_e^{(comp)}$), LM query cost ($\epsilon_e^{(query)}$), and key references. $d$ denotes the logit or representation dimension, $V$ the vocabulary size, and $k$ the number of draft tokens verified per batch; ``--'' marks a sub-budget with negligible burden.}
\label{tab:carrier_efficiency}
\renewcommand{\arraystretch}{1.10}
\resizebox{\textwidth}{!}{
\begin{tabular}{l|l|l|l|l}
\toprule
\textbf{Transferred Carrier}
& \textbf{Comm Burden ($\epsilon_e^{(comm)}$)}
& \textbf{Comp Burden ($\epsilon_e^{(comp)}$)}
& \textbf{Query Burden ($\epsilon_e^{(query)}$)}
& \textbf{Key References} \\
\midrule

\multirow{2}{*}{\shortstack[l]{Logits / Representations}}
& $O(|\mathcal{D}_{public}|{\cdot}d)$ signals / round
& SM forward pass / sample
& offline LM batch calls
& \keyref{cheng2021fedgems}; \keyref{yumultimodal} \\
\cmidrule{2-5}
& $O(V)$ logits / step
& SM forward pass / token
& per-step LM call
& \keyref{zhang2024cogenesis}; \keyref{thareja2026dp} \\
\midrule

Tokens
& $O(k)$ tokens / verification
& $k$-token draft generation
& one LM verification / $k$ tokens
& \keyref{leviathan2023fast}; \keyref{xia2024unlocking} \\
\midrule

Synthetic Data
& $O(|\mathcal{D}_{syn}|)$ corpus
& SM generation / downstream training
& offline LM calls when LM generates
& \keyref{ye2022zerogen}; \keyref{lindifferentially} \\
\midrule

Intermediate Outputs
& activations + gradients / step
& split forward / backward on SM side
& --
& \keyref{shen2023split}; \keyref{mudvari2024splitllm} \\
\midrule

Student Model
& one-time GB-scale transfer
& offline distillation / local deployment
& --
& \keyref{xiao2023offsite}; \keyref{peng2024fedpft} \\
\midrule

Compressed Weights / Adapters
& compressed block or adapter / transfer
& local deployment / adaptation
& --
& \keyref{fan2023fatellm}; \keyref{peng2024fedpft} \\
\cmidrule{1-5}
Adapters / Tunable Prompts
& adapter or prompt update / round
& PEFT or prompt tuning
& --
& \keyref{hu2021lora}; \keyref{zhang2023fedpetuning} \\
\midrule

Prompts
& context tokens / turn
& retrieve / rerank / compress
& per-turn LM call
& \keyref{pan2024llmlingua}; \keyref{guo2025accrag} \\
\midrule

Outputs
& response / plan / tool-call tokens
& parse / execute / monitor
& output-token LM cost
& \keyref{chen2023frugalgpt}; \keyref{yi2025ecoagent} \\

\bottomrule
\end{tabular}
}
\end{table*}

Efficiency in LM--SM collaboration is formalized in our framework as the resource constraint $M_e \preceq \boldsymbol{\epsilon}_e$ in Eq.~\eqref{eq:problem}, introduced in Section~\ref{sec:problem_definition}, where $M_e$ captures the total resource consumption induced by the collaborative policy $\pi$. We further distinguish three complementary sub-budgets $\boldsymbol{\epsilon}_e = (\epsilon_e^{(comm)}, \epsilon_e^{(comp)}, \epsilon_e^{(query)})$: $\epsilon_e^{(comm)}$ bounds the volume and frequency of carrier exchanges $(\mathcal{I}_{\pi}, \mathcal{J}_{\pi})$ across the trust boundary, $\epsilon_e^{(comp)}$ constrains the local computation on the SM side required to generate or process these carriers, and $\epsilon_e^{(query)}$ limits the monetary and latency cost of invoking the remote LM. All three sub-budgets are jointly shaped by the carrier type instantiated by $\pi$ across different collaboration paradigms (Table~\ref{tab:carrier_efficiency}). In Section~\ref{subsubsec:carrier_efficiency}, we characterize these carrier-specific efficiency profiles. In Section~\ref{subsubsec:efficiency_optimization} we discuss the corresponding optimization techniques for each aspect.
\subsubsection{Carrier-Specific Efficiency Profiles}
\label{subsubsec:carrier_efficiency}

In cross-silo LM--SM collaboration, the resource burden imposed by the collaborative policy $\pi$ is fundamentally determined by the type of carrier $(\mathcal{I}_\pi, \mathcal{J}_\pi)$ crossing the trust boundary, as summarized in Table~\ref{tab:carrier_efficiency}.  Below we briefly discuss the dominant bottleneck and the corresponding cost for each carrier type.

\paragraph{\textbf{Logits / Representations.}}
Logits and representations are high-dimensional carriers, so their communication burden scales with both the signal dimensionality and the collaboration frequency. The dominant cost regime, however, depends on when the carrier is exchanged. In Distillation-based Transfer, logits or representations are released over public samples, so the cost scales with the number of public samples and the per-sample signal dimensionality~\cite{cheng2021fedgems, yumultimodal}. Because these signals can be precomputed and transferred in batches, the binding cost is the total volume that must be transmitted or stored, not per-step online latency.
In inference-time Collaborative Decoding, by contrast, each decoding step transmits a vocabulary-sized logit vector and triggers one LM call, so the bottleneck shifts from cumulative volume to synchronous communication and per-step query cost~\cite{zhang2024cogenesis,thareja2026dp}.

\paragraph{\textbf{Tokens.}}
Token-wise collaborative decoding reduces carrier dimensionality by replacing per-step logit transmission with $k$ draft tokens that the LM verifies in a batch~\cite{leviathan2023fast, xia2024unlocking, kim2024speculative}. Speculative decoding therefore lowers the carrier volume over a $k$-token verification window from $O(kV)$ to $O(k)$, and one LM verification call can amortize multiple token decisions. This efficiency gain depends on SM draft cost and the acceptance rate of the LM verifier, creating a trade-off between SM-side draft computation and LM query cost.

\paragraph{\textbf{Synthetic Data.}}
In generation-based transfer, the carrier is an explicit proxy corpus $\mathcal{D}_{syn}$. Its communication cost scales with corpus volume, but the transfer can often be scheduled as a one-time or offline operation. The heavier burden lies in generation and subsequent training: one party must synthesize the corpus, and the receiving party must consume it through downstream optimization~\cite{ye2022zerogen, wang2018dataset, zhao2021dataset}. Query cost, when the LM is the generator, is large in total token volume but temporally flexible~\cite{lindifferentially}; this separates synthetic data from latency-bound interactive carriers.

\paragraph{\textbf{Intermediate Outputs.}}
Split Learning is dominated by synchronization frequency. Hidden states and backward gradients cross the boundary at every step, so even moderate per-step payloads accumulate over the training or inference trajectory~\cite{shen2023split}. The SM-side computation depends on the cut layer, since an earlier cut reduces local computation but increases reliance on remote execution, while a later cut has the opposite effect~\cite{kang2017neurosurgeon, mudvari2024splitllm}. Because the LM acts as a co-executor rather than an invoked API, this carrier largely removes $\epsilon_e^{(query)}$ and concentrates optimization on the communication and local-computation balance.

\paragraph{\textbf{Student Models / Compressed Weights / Adapters / Tunable Prompts.}}
Distillation-based and parameter-based transfer both transmit reusable artifacts across the trust boundary, but the transmitted carrier differs. In LM$\rightarrow$SM distillation-based transfer, the carrier is a released student model, including compact proxy or emulator variants. Parameter-based transfer instead transmits a compressed weight block, adapter, or tunable prompt. For these reusable carriers, the dominant efficiency factor is not interaction frequency but artifact size, update granularity, and local adaptation cost. Student-model transfer is typically a one-time transfer that can reach GB scale for full student models, while its computation cost is paid during offline distillation or compression~\cite{xiao2023offsite, peng2024fedpft}. Parametric carriers are usually smaller, but compressed weight blocks may still be large, whereas PEFT adapters and tunable prompts can be synchronized as MB-scale or smaller updates across rounds~\cite{fan2023fatellm, peng2024fedpft}. They also reduce local training cost because only a small parameter subset is updated locally~\cite{hu2021lora, zhang2023fedpetuning}. After transfer, both carrier groups impose little query pressure because the receiving party can reuse the transferred artifact without per-call LM access.

\paragraph{\textbf{Prompts / Outputs.}}
Context-augmented collaboration stresses all three sub-budgets per interaction, but the direction matters. \emph{Prompts (SM$\to$LM)} carry user queries, retrieved evidence, or tool results, so long contexts increase communication and paid token usage while retrieval, reranking, and compression add SM-side computation~\cite{pan2024llmlingua, xu2024recomp, guo2025accrag}. \emph{Outputs (LM$\to$SM)} carry responses, plans, or tool-call specifications, so their cost grows with output length and interaction depth in agentic workflows~\cite{chen2023frugalgpt, ding2024hybrid, yi2025ecoagent}. This carrier family is therefore sensitive not only to payload size, but also to turn count and workflow control.

In summary, no carrier is uniformly efficient across all three aspects. Logit-based carriers emphasize communication volume and online latency; synthetic data concentrates cost in offline generation and training; intermediate outputs impose high-frequency synchronization; student models require one-time model transfer and offline distillation; compressed weights, adapters, and tunable prompts trade artifact size for local adaptability; and Prompts / Outputs scale with both token budget and interaction depth. The optimization techniques in Section~\ref{subsubsec:efficiency_optimization} should therefore be selected according to the dominant bottleneck of the chosen carrier.

\subsubsection{Efficiency Optimization Techniques}
\label{subsubsec:efficiency_optimization}

Table~\ref{tab:efficiency_techniques} summarizes the efficiency optimization landscape, mapping each technique to the carrier type and efficiency factor that it most directly targets ($\epsilon_e^{(comm)}$, $\epsilon_e^{(comp)}$, or $\epsilon_e^{(query)}$).

\begin{table*}[t]
\centering
\small
\setlength{\tabcolsep}{4pt}
\renewcommand{\arraystretch}{0.95}
\caption{Efficiency optimization techniques for LM--SM collaboration, organized by technique, primary carrier, target efficiency sub-budget ($\epsilon_e^{(comm)}$/$\epsilon_e^{(comp)}$/$\epsilon_e^{(query)}$), and key references.}
\label{tab:efficiency_techniques}
\resizebox{\textwidth}{!}{%
\begin{tabular}{l|l|l|c|l}
\toprule
\textbf{Category} & \textbf{Technique} & \textbf{Primary Carrier} & \textbf{Sub-budget} & \textbf{Key References} \\
\midrule

\multirow{6}{*}{\shortstack[l]{\textbf{Communication}\\\textbf{Efficiency}}}
& One-shot Communication
    & Student Model; Compressed Weights / Adapters
    & $\epsilon_e^{(comm)}$
    & \keyref{xiao2023offsite}; \keyref{peng2024fedpft} \\
\cmidrule{2-5}
& Prototype-based Federated Aggregation
    & Logits / Representations
    & $\epsilon_e^{(comm)}$
    & \keyref{tan2022fedproto}; \keyref{zhang2024fedtgp} \\
\cmidrule{2-5}
& Asynchronous / Local Updates
    & Logits / Representations; Compressed Weights / Adapters
    & $\epsilon_e^{(comm)}$
    & \keyref{liu2024vertical}; \keyref{cao2024sfprompt} \\
\cmidrule{2-5}
& Quantization \& Activation Compression
    & Intermediate Outputs
    & $\epsilon_e^{(comm)}$
    & \keyref{yao2022zeroquant}; \keyref{shen2023split} \\
\cmidrule{2-5}
& Observation Filtering
    & Prompts / Outputs
    & $\epsilon_e^{(comm)}$
    & \keyref{yi2025ecoagent}; \keyref{fancore} \\
\cmidrule{2-5}
& Prompt / Context Compression
    & Prompts
    & $\epsilon_e^{(comm)}$, $\epsilon_e^{(query)}$
    & \keyref{pan2024llmlingua}; \keyref{xu2024recomp} \\
\midrule

\multirow{4}{*}{\shortstack[l]{\textbf{Computation}\\\textbf{Efficiency}}}
& Parameter-efficient Fine-tuning (PEFT)
    & Compressed Weights / Adapters; Adapters / Tunable Prompts
    & $\epsilon_e^{(comp)}$
    & \keyref{hu2021lora}; \keyref{fan2023fatellm} \\
\cmidrule{2-5}
& Split-point Selection
    & Intermediate Outputs
    & $\epsilon_e^{(comp)}$, $\epsilon_e^{(comm)}$
    & \keyref{kang2017neurosurgeon}; \keyref{chen2024adaptive} \\
\cmidrule{2-5}
& Dataset Distillation \& Condensation
    & Synthetic Data
    & $\epsilon_e^{(comp)}$, $\epsilon_e^{(comm)}$
    & \keyref{wang2018dataset}; \keyref{zhao2021dataset} \\
\cmidrule{2-5}
& Speculative Decoding
    & Tokens
    & $\epsilon_e^{(comm)}$, $\epsilon_e^{(comp)}$, $\epsilon_e^{(query)}$
    & \keyref{leviathan2023fast}; \keyref{xia2024unlocking} \\
\midrule

\multirow{5}{*}{\shortstack[l]{\textbf{LM Query}\\\textbf{Cost}}}
& Query Routing
    & Prompts; Tokens; Synthetic Data
    & $\epsilon_e^{(query)}$
    & \keyref{chen2023frugalgpt}; \keyref{ding2024hybrid} \\
\cmidrule{2-5}
& Prompt / Context Compression
    & Prompts
    & $\epsilon_e^{(query)}$, $\epsilon_e^{(comm)}$
    & \keyref{pan2024llmlingua}; \keyref{guo2025accrag} \\
\cmidrule{2-5}
& Speculative Batching
    & Tokens
    & $\epsilon_e^{(query)}$
    & \keyref{leviathan2023fast}; \keyref{kim2024speculative} \\
\cmidrule{2-5}
& Offline / Planned Batch Synthesis
    & Synthetic Data
    & $\epsilon_e^{(query)}$
    & \keyref{ye2022zerogen}; \keyref{lindifferentially} \\
\cmidrule{2-5}
& Cascading Inference
    & Prompts / Outputs
    & $\epsilon_e^{(query)}$
    & \keyref{chen2023frugalgpt}; \keyref{yi2025ecoagent} \\
\bottomrule
\end{tabular}%
}
\end{table*}

\paragraph{\textbf{Communication Efficiency ($\epsilon_e^{(comm)}$).}}
\label{subsubsec:comm_efficiency}
Communication cost is typically the bottleneck of cross-silo approaches. Besides model copyright protection, another significant reason for cross-silo knowledge transfer is the benefit on communication cost, since transferring entire LM may require the transmission of billions of model parameters, which is prohibitive. Benefiting from the compressed nature of transferred knowledge, the communication overhead of cross-silo knowledge transfer is typically much lower than transferring entire model parameters, as evidenced by previous research reports~\cite{cheng2021fedgems}. This effect further manifests when the model parameters transferred become large~\cite{cheng2021fedgems}.
Existing communication optimization techniques reduce either the \emph{payload size} of each carrier exchange or the \emph{synchronization frequency} across the trust boundary. We align representative techniques along these two axes, classifying them within each dimension according to the specific carrier type they use.

\noindent\textbf{Reducing payload size:}
\begin{itemize}
    \item \textbf{Prototype-based Federated Aggregation} replaces sample-wise logits or representations with class-level prototypes, reducing carrier volume while requiring prototypes to remain representative under data heterogeneity~\cite{tan2022fedproto, zhang2024fedtgp, zhang2024upload}.
    \item \textbf{Quantization \& Activation Compression} lowers the bit-width or dimensionality of activations and gradients exchanged in Split Learning, reducing per-round payload while risking precision loss~\cite{yao2022zeroquant, shen2023split}.
    \item \textbf{Prompt / Context Compression} compresses retrieved chunks or long context windows into shorter prompts before LM invocation~\cite{pan2024llmlingua, xu2024recomp, guo2025accrag}. This reduces both communication and token usage, but it shifts work to the SM side and may drop task-relevant evidence.
    \item \textbf{Observation Filtering} filters multi-step execution traces before transmission in agentic settings, reducing the amount of textual carriers that cross the boundary~\cite{yi2025ecoagent, fancore}. The cost is additional local selection logic and possible loss of useful task state.
\end{itemize}

\textbf{Reducing communication frequency:}
\begin{itemize}
    \item \textbf{One-shot Communication} compresses knowledge into a reusable student model, compressed weight block, adapter, or tunable prompt so that the carrier crosses the boundary once rather than across repeated rounds, at the cost of a larger offline preparation step~\cite{xiao2023offsite, peng2024fedpft}.
    \item \textbf{Asynchronous / Local Updates} allows each party to perform multiple local updates between synchronization rounds, amortizing per-round latency in distillation-based or parameter-based collaboration~\cite{liu2024vertical, cao2024sfprompt}.
    \item \textbf{Speculative Decoding} replaces per-token LM interaction with batched verification over $k$ draft tokens, converting many synchronization steps into a single verification call~\cite{leviathan2023fast, xia2024unlocking, kim2024speculative}.
\end{itemize}

\paragraph{\textbf{Computation Efficiency ($\epsilon_e^{(comp)}$).}}
\label{subsubsec:comp_efficiency}
A major advantage of enabling cross-silo knowledge transfer between LMs and SMs is that the SMs are no longer required to have abundant computational GPU resources to host the LMs locally, which would significantly hinder its applications in real-world scenarios where the majority of data parties are resource-constrained devices or organizations. Cross-silo knowledge transfer approaches allow SMs to have lightweight model architectures which are computationally efficient to deploy.
Through cross-silo knowledge transfer, the computation burden on the client side is shifted to the resource-abundant server, allowing resource-constrained parties to share the benefit of the high performance of LMs. For example, \cite{deng2023mutual} accomplished server-side fine-tuning of Llama-7B with cross-silo knowledge transfer from SMs who used BERT and DistilBERT models as local models, which represents roughly a 100-times reduction in model size. The remaining SM-side computation is carrier-dependent: adapters require local lightweight training, Synthetic Data requires generation or training over a proxy corpus, Intermediate Outputs require repeated split-head execution, and speculative decoding requires local draft generation. The following techniques either shrink local trainable state or choose where computation is placed:
\begin{itemize}
    \item \textbf{Parameter-efficient Fine-tuning (PEFT)} restricts local updates to low-rank or adapter parameters, making parameter-based collaboration feasible on edge hardware~\cite{hu2021lora, zhang2023fedpetuning, fan2023fatellm}. The benefit is lower local compute and memory use; the trade-off is reduced adaptation capacity relative to full fine-tuning.
    \item \textbf{Dataset Distillation \& Condensation} replaces the full Synthetic Data carrier with a compact distillate, reducing downstream training cost and communication volume~\cite{wang2018dataset, zhao2021dataset, zhang2024one}. The risk is that the compact corpus may underrepresent private-domain diversity.
    \item \textbf{Split-point Selection} chooses the partition layer that balances local head computation against Intermediate Output traffic in Split Learning~\cite{kang2017neurosurgeon, chen2024adaptive, mudvari2024splitllm}. It optimizes a boundary placement problem rather than simply reducing computation.
    \item \textbf{Speculative Decoding} deliberately increases SM-side draft computation to reduce LM verification calls and communication frequency~\cite{leviathan2023fast, xia2024unlocking, kim2024speculative}. Its efficiency gain is therefore conditional on a high draft acceptance rate.
\end{itemize}
However, existing approaches mostly assume a predefined set of local model structures, and studying how local models can adapt to local and server-side resources remains an important topic for future work. Furthermore, future work should take an evolving perspective and consider the growing computation capabilities of both datacenter hardware and edge devices, including mobile devices with accelerated capabilities for fine-tuning pre-trained models~\cite{woisetschlager2023federated}.

\paragraph{\textbf{LM Query Cost ($\epsilon_e^{(query)}$).}}
\label{subsubsec:query_efficiency}
Due to the enormous size and costs of LM deployment, access to LM through API services can be highly expensive to small companies and individuals, putting query cost analysis and optimization into perspective. In addition, existing use of large LMs raises substantial environmental and sustainability concerns~\cite{wu2022sustainableaienvironmentalimplications}. Enabling cross-silo collaboration between large and small models can reduce avoidable LM invocations. 
The following techniques reduce $\epsilon_e^{(query)}$ by avoiding, batching, shortening, or routing LM calls:
\begin{itemize}
    \item \textbf{Query Routing.} Router models decide whether a request should be handled locally or escalated to the LM, balancing response quality against invocation cost~\cite{ding2024hybrid, chen2023frugalgpt}. The key risk is misrouting difficult inputs to an underpowered SM.
    \item \textbf{Offline / Planned Batch Synthesis.} In LM$\to$SM generation-based transfer, the LM can be queried in scheduled offline batches, making cost predictable and amortizable~\cite{ye2022zerogen, lindifferentially}. This reduces latency pressure but does not eliminate total token cost.
    \item \textbf{Prompt Compression.} Prompt compression shortens each LM call by distilling long contexts into compact Prompts before forwarding~\cite{pan2024llmlingua, xu2024recomp, guo2025accrag}. It reduces both query and communication cost, while shifting preprocessing to the SM side.
    \item \textbf{Speculative Batching.} Batch speculative decoding converts per-token LM calls into per-batch verification calls~\cite{leviathan2023fast, xia2024unlocking, kim2024speculative}. The resulting query reduction depends on how often the LM accepts the SM draft.
    \item \textbf{Cascading Inference.} Cascading inference invokes the LM only when the SM response fails to meet a confidence or quality threshold~\cite{chen2023frugalgpt, ding2024hybrid, yi2025ecoagent}. It reduces average query cost, but requires calibrated confidence estimation to avoid silent utility loss.
\end{itemize}
However, systematic query-cost optimization for LM--SM collaboration remains underexplored.

\paragraph{\textbf{Open Challenges.}}
The preceding analysis also exposes structural gaps that reflect misalignments between how efficiency is currently formulated and how carriers actually behave across the trust boundary.
\begin{itemize}
    \item \textbf{Knowledge carriers cross an economically asymmetric trust boundary.} For the budget $\boldsymbol{\epsilon}_e = (\epsilon_e^{(comm)}, \epsilon_e^{(comp)}, \epsilon_e^{(query)})$ in Eq.~\eqref{eq:problem}, the two sides of the trust boundary operate under qualitatively different cost functions: in API-mediated settings, the LM side is often monetized by per-token pricing and subject to rate limits~\cite{chen2023frugalgpt, ding2024hybrid}, while the SM side is bounded by device-level energy, memory, and thermal budgets~\cite{zhang2023fedpetuning, woisetschlager2023federated}. Most carrier-level optimizations in Section~\ref{subsubsec:carrier_efficiency} do not reduce total cost but \emph{relocate} it across the boundary: speculative decoding shifts cost from LM query count to SM draft compute~\cite{leviathan2023fast, xia2024unlocking}; prompt compression shifts cost from LM token usage to SM pre-processing~\cite{pan2024llmlingua, xu2024recomp}; cascading inference shifts cost from LM invocation to SM confidence estimation~\cite{chen2023frugalgpt, ding2024hybrid}. A key open challenge is how to aggregate these redirected costs into $M_e$ without preventing principals on either side of the trust boundary from evaluating a collaborative policy under their own budgets 
    ~\cite{deb2011multi}.

    \item \textbf{Carrier type is committed at design time rather than selected as a state-dependent policy.} Each paradigm in Section~\ref{sec:taxonomy} instantiates a small set of carrier types across the interaction trajectory: cross-silo distillation transmits Logits / Representations, collaborative decoding may transmit Logits or Tokens depending on the decoding interface, Split Learning transmits Intermediate Outputs, and context-augmented collaboration transmits Prompts / Outputs. The carrier-specific profiles in Table~\ref{tab:carrier_efficiency} show that the cheapest admissible carrier varies across task phases, with Logits / Representations supporting rich supervision, Tokens supporting low-volume verification, and Compressed Weights / Adapters / Tunable Prompts supporting long-horizon reuse~\cite{deng2023mutual, leviathan2023fast, hu2021lora}. A state-dependent routing policy $\pi : s_t \mapsto (\mathcal{I}_{\pi,t}, \mathcal{J}_{\pi,t})$ that selects the cheapest carrier per step would exploit this heterogeneity, but standard privacy accounting is not yet tailored to heterogeneous cross-carrier switching across trust boundaries~\cite{dwork2006calibrating, abadi2016deep}.

    \item \textbf{Throughput optimization increases the extractable capability per unit query budget.} Throughput-oriented carrier optimizations reduce $\epsilon_e^{(query)}$ by amortizing per-call overhead across more carrier signals: speculative decoding returns $k$ verified tokens per LM invocation~\cite{leviathan2023fast, xia2024unlocking, kim2024speculative}, and cascading inference defers LM access until SM confidence is exceeded~\cite{chen2023frugalgpt, ding2024hybrid}. The same amortization raises the oracle signal density returned per query, which is the denominator governing model extraction and logit-based inference attacks~\cite{orekondy2019knockoff, carlini2024stealing, birch2023model}. Existing efficiency benchmarks report throughput and monetary cost but not extractable capability per query budget, so an efficiency gain of $N\times$ may correspond to an unchanged, reduced, or amplified attack surface without any measured distinction. A notion of \emph{leakage-preserving} efficiency transformation, under which a speedup is evaluated by whether it raises the adversary's extraction rate at fixed query budget, remains underdeveloped in LM--SM collaboration benchmarks.
\end{itemize}

\subsection{Strategic Trade-offs: The Collaborative Trilemma}
\label{sec:tri_lemma}

Synthesizing the formal framework in Eq.~\eqref{eq:problem} and the carrier-specific analyses above, we formulate the \textbf{Collaborative Trilemma}. The design of an optimal collaborative policy $\pi^*$ is a multi-objective problem: maximize the domain utility functional $\mathcal{F}(\pi)$ while satisfying the privacy and security budgets $M_p \le \epsilon_p$, $M_L \le \epsilon_L$, and $M_r \le \epsilon_r$, together with the efficiency constraints $M_e \preceq \boldsymbol{\epsilon}_e$~\cite{deb2011multi, zhang2023trading, chen2020breaking}. The three objectives do not decouple into independent pairwise trade-offs. They share a single coupled variable, the carrier $(\mathcal{I}_\pi, \mathcal{J}_\pi)$, through which utility-relevant content, leakage risk, integrity risk, and resource cost must all cross the trust boundary. Tightening any one constraint acts on the same carrier instantiation that delivers the other two, so the three-way tension is structural rather than a coincidental aggregation of pairwise conflicts.

Optimizing all three objectives without trade-offs is generally infeasible in typical LM--SM deployments. Each pairwise tension below specifies the carrier transformation that the corresponding constraint forces.

\paragraph{\textbf{The Fidelity Penalty (Privacy and Security vs. Utility).}}
Enforcing tight privacy and security bounds on the carrier requires operations that transform it into a statistically detached form. Differential privacy injects noise into Logits / Representations or Intermediate Outputs~\cite{dwork2006calibrating, shen2023split}, and sanitization removes semantic fields from Prompts, including retrieved evidence~\cite{pilan2022text, chong2024casper}. The utility functional $\mathcal{F}$ is evaluated on this same transformed carrier, so the information removed to satisfy $M_p \le \epsilon_p$ or $M_L \le \epsilon_L$ or to reject carriers under $M_r \le \epsilon_r$ and the information consumed by downstream training or inference are drawn from the same signal. $\mathcal{F}$ therefore typically degrades as these bounds remove or perturb more task-relevant carrier content, and the penalty cannot be isolated from the carrier itself.

\paragraph{\textbf{The Protection Overhead (Privacy and Security vs. Efficiency).}}
Preserving carrier fidelity while tightening $M_p$ and $M_L$, or validating inbound carriers under $M_r$, introduces additional protection overhead through secure computation, filtering, detection, or guarded execution. Homomorphic encryption~\cite{chen-etal-2022-x} and secure multi-party computation~\cite{dong2023pumasecureinferencellama7b} retain the underlying signal for carriers such as Intermediate Outputs, Prompts, or prediction-side carriers, but they expand the payload and per-step processing cost by several orders of magnitude. The result can exceed $\epsilon_e^{(comm)}$ or $\epsilon_e^{(comp)}$, especially on the resource-constrained SM side. The carrier keeps its semantic content, but its protected representation no longer fits the resource profile that motivated lightweight collaboration.

\paragraph{\textbf{The Compression Loss (Efficiency vs. Utility).}}
Satisfying $\boldsymbol{\epsilon}_e$ under stringent bandwidth, compute, or query limits requires reducing carrier information content at the source. Prompt compression discards tokens from Prompts~\cite{pan2024llmlingua, xu2024recomp}, quantization truncates the precision of Intermediate Outputs~\cite{yao2022zeroquant}, and dataset distillation substitutes compact Synthetic Data for the full training set~\cite{zhao2021dataset}. What crosses the boundary after these operations is a lower-capacity substitute for the original signal, and the utility functional $\mathcal{F}$ inherits the residual bottleneck as representation collapse, context dropping, or distillation drift.

The intersection of the three tensions is not a single Pareto point that a stronger primitive can locate. It consists of the carrier-level open problems articulated in the preceding analysis: carrier abstraction that does not reduce privacy, defenses that cannot retroactively cross the boundary, multi-carrier executions without coherent privacy accounting, efficiency accounting that ignores the economic asymmetry of the trust boundary, and throughput optimizations that raise the extractable capability per query. Progress toward $\pi^*$ therefore requires carrier-level formalisms that close these gaps jointly, rather than stronger primitives within any single dimension. These gaps motivate the benchmarking, protocol, and deployment agenda developed in Section~\ref{sec:future_directions}.

\section{Future Directions}
\label{sec:future_directions}

The Collaborative Trilemma formalized in Section~\ref{sec:tri_lemma} reveals
that no existing paradigm simultaneously achieves high domain utility,
rigorous privacy and security assurance, and resource efficiency, and that progress on one
axis routinely incurs regressions on the others. Rather than treating this
tension as a fundamental barrier, we view it as a map for future research:
each dimension of the trilemma identifies an area where the field currently
lacks the tools, benchmarks, or protocols needed to make principled trade-offs.

Building on this diagnosis, this section outlines four clusters of open
directions. Section~\ref{subsec:benchmarking} surveys the need for \textbf{standardized
benchmarking and evaluation} frameworks that natively measure all three
trilemma dimensions rather than collapsing to single-axis accuracy
leaderboards.  Section~\ref{subsec:system_standardization} identifies gaps in \textbf{system
standardization and adaptive protocols}, including API interfaces that can
safely transmit non-text carriers and dynamic policies that navigate the
Pareto frontier in real time.  Section~\ref{subsec:applications} grounds these directions in
\textbf{application-specific demonstrations}, including urban intelligence, business
intelligence, and personalized intelligence, where the interplay of
utility, privacy and security assurance, and efficiency is most acute and the societal payoff of
solving the trilemma is highest. Section~\ref{subsec:continuous_adaptation}
discusses \textbf{continuous learning and task adaptation} under evolving tasks,
data distributions, model capabilities, and resource conditions.

\subsection{Benchmarking and Evaluation}
\label{subsec:benchmarking}

\textbf{Metrics Standardization.}  Previous works employing diverse methodologies often adopt disparate definitions of evaluation metrics, resulting in incomparable results. This issue is particularly acute for metrics with nuanced definitions that vary across application scenarios. Privacy serves as a prime example. While Differential Privacy (DP) offers a gold standard for quantifying privacy, DP-based approaches frequently necessitate utility sacrifices, hindering the evaluation of research directions that cannot be readily analyzed through this lens. For instance, distillation-based approaches are often perceived as reducing privacy risk because raw data do not cross the boundary \cite{dong2022privacy}, yet quantifying the privacy implications of emerging privacy-preserving techniques using DP remains an open challenge. This presents an opportunity to establish a unified and measurable framework that encompasses a broader range of cross-silo collaboration approaches, including distillation-based, generation-based, and parameter-based methods.

\textbf{Application-driven Evaluations.} Many existing approaches are method-driven: they aim to improve accuracy on predefined task types, but they often do not test whether the resulting collaboration remains usable under deployment constraints. In practical settings, communication cost, computation allocation, privacy protection, interpretability, and robustness constrain the system in addition to task utility. Current evaluations rarely measure these constraints jointly, making it difficult to compare the practical applicability of different approaches.

\textbf{Real-world Datasets and Benchmarks.} Existing approaches primarily evaluate domain-specific tasks using open-source, well-established benchmarks like IMDB \cite{maas2011learning_imdb}. However, these benchmarks often provide limited evidence for LM--SM collaboration. Their data are close to the public pre-training distribution, whereas the target setting assumes a private domain with a larger domain gap; this can inflate performance estimates. As a result, evaluations on standard downstream benchmarks may not indicate how a method behaves on practical private-domain tasks, especially when efficiency and privacy risks are considered~\cite{pmlr-v235-tramer24a}. To reduce this bias, the field needs private-domain benchmarks built from realistic sensitive datasets, together with reporting protocols that describe the domain gap. Current benchmarks also emphasize final task accuracy rather than vulnerability at the collaborative boundary. Future datasets should evaluate the transferred \emph{carriers} themselves by measuring how much private or proprietary information can be reconstructed from a prompt, intermediate output, or logit distribution under realistic constraints~\cite{carlini2021extracting}.

\textbf{Multi-objective Trade-offs.} Practical LM--SM collaboration requires evaluation beyond task utility. Depending on the application, the objective set may include efficiency, privacy and security, interpretability, and fairness. Previous work has studied trade-offs among utility, privacy, efficiency, and fairness in FL settings \cite{zhang2023trading, GU2022102907}, or optimized accuracy under communication and privacy constraints \cite{chen2020breaking}. Collaborative systems with LMs add further objectives, including LM query cost and LM ownership protection. To incorporate these measures and compare different approaches, we propose a flexible multi-objective benchmark over collaborative policies defined as
\begin{equation}
    \max_{\pi} f(\pi) = f(o_1(\pi), \ldots, o_n(\pi)),
\end{equation}
where $\pi$ denotes the collaborative policy or method to be evaluated; $o_1,\ldots,o_n$ denote key objectives such as domain utility, privacy and security assurance, and resource efficiency. The scoring function $f$ specifies how these objectives are aggregated. A policy $\pi_1$ is a Pareto improvement over another policy $\pi_2$ if it improves at least one objective without degrading any other considered objective.
In most realistic scenarios, methods exhibit trade-offs, so $f$ should be designed to compare and rank approaches under a unified criterion. For example, $f$ can be a weighted linear combination of multiple objectives. As established in the Collaborative Trilemma (Sec.~\ref{sec:tri_lemma}), future benchmarks must report three-dimensional metrics encompassing Domain Utility ($\mathcal{F}$), Privacy and Security Assurance ($M_p, M_L, M_r$ under $\epsilon_p, \epsilon_L, \epsilon_r$), and Resource Efficiency ($M_e$ under $\boldsymbol{\epsilon}_e$). Moving beyond one-dimensional accuracy leaderboards toward Pareto evaluations is essential for identifying practical collaborative policies.

\subsection{System Standardization and Adaptive Protocols}
\label{subsec:system_standardization}

\textbf{Standardized API Protocols.} Currently, commercial LM APIs predominantly support text-based prompts, which naturally aligns with Context-Augmented Collaboration. At the context level, early standardization is emerging: tool-calling interfaces and agent interoperability protocols such as the Model Context Protocol (MCP) and the Agent-to-Agent (A2A) protocol define typed exchanges of requests, tool results, and task state across services~\cite{ehtesham2025survey}. However, for architectures like Split Learning or Collaborative Decoding to achieve widespread adoption, the industry requires standardized API endpoints capable of receiving and returning intermediate representations, draft tokens, or logits safely and efficiently across the trust boundary~\cite{shen2023split}. Research frameworks have prototyped such interfaces for split execution~\cite{gu2025vflair, fan2023fatellm}, and serving-side precedents exist in truncated log-probability outputs and draft-verification interfaces for speculative decoding~\cite{leviathan2023fast}, yet no widely adopted cross-party contract specifies the tensor formats, synchronization semantics, failure handling, or protection hooks required by these carriers. Standardization is also a security problem: logit or log-probability interfaces have been shown to leak proprietary information about API-protected LMs~\cite{carlini2024stealing, finlayson2024logits}. A practical protocol should couple each endpoint with budget enforcement, such as top-$k$ truncation, rate limiting, and noise injection for transmitted logits under the relevant privacy or model-security budget~\cite{zhang2024cogenesis, thareja2026dp}.

\noindent\textbf{Dynamic Policy Switching.} The collaborative paradigms defined in this survey are often treated as static choices. A critical future direction is the development of adaptive systems that dynamically switch collaboration modes based on real-time constraints. 
Designing adaptive agents capable of crawling the Pareto frontier in real-time remains largely unexplored~\cite{ding2024hybrid, chen2024adaptive}.
Concretely, this can be framed as adaptive collaboration-policy selection, where a policy may choose whether to answer locally, invoke the LM, or switch among LM--SM collaboration modes~\cite{chen2023frugalgpt, ong2024routellm, leviathan2023fast, xia2024unlocking, kim2024speculative}. Such policies should be conditioned on the budgets and risks formalized earlier. 
Beyond finding a low-cost route, another key open problem is to select Pareto-efficient policies without violating privacy, ownership, or integrity constraints across successive carrier choices. Multi-carrier switching raises additional questions: how to account for cumulative leakage when a session alternates between Prompts / Outputs, Logits / Representations, Tokens, and Intermediate Outputs; how to prevent unstable policies from oscillating across modes; and how standardized APIs can support these carrier choices without increasing the attack surface. These issues remain largely open in LM--SM collaboration.

\subsection{Application-specific Demonstration}
\label{subsec:applications}

Building successful real-world demonstrations will not only facilitate the investigation and prioritization of techniques and architectures for achieving collaborative learning, but also help gather user data and feedbacks to further refine and adapt the technology.

\textbf{Urban Intelligence.} Image recognition is a well-established application in urban management. Due to highly sensitive nature of images and videos, centralized processing is often not possible. In the past, small image recognition models, such as ResNet~\cite{he2016deep} and the YOLO series~\cite{redmon2016you, redmon2017yolo9000, redmon2018yolov3, bochkovskiy2020yolov4, li2022yolov6, wang2023yolov7}, have been deployed in edge camera devices for performing time-critical tasks such as hazard detection. How to leverage the power of pre-trained generative AI models such as DALL·E~\cite{ramesh2021zero, ramesh2022hierarchical, betker2023improving}, GLIDE~\cite{nichol2022glide}, or Stable Diffusion~\cite{rombach2022high} to continuously improve the performance of small models is a new frontier. One possible solution is through synthetic data generation, with small models deployed on site for real-time monitoring and feedbacks. However, substantial barriers related to cost and efficiency have yet to be overcome. From a taxonomy perspective, this scenario heavily relies on \textit{Generation-based Transfer}~\cite{ye2022zerogen} or \textit{Split Learning}~\cite{shen2023split}, where the heavy computation ($\epsilon_e^{(comp)}$) on edge cameras must be natively integrated with cloud-side LM features.

\textbf{Business Intelligence.} Given the sensitive and proprietary nature of business data, corporations are generally reluctant to expose it to external Machine Learning as a Service (MLaaS) vendors. On the other hand, smaller models such as logistic regression and boosting trees, are efficient to analyze and mine various structured data, and are the incumbent in industries for years. A promising opportunity here is to build a collaborative AI system which combines the excellent understanding and generation capabilities of LMs and efficient data mining and retrieval capabilities of SMs. For example, AsiaInfo recently deployed a large-small collaboration technology for handling wireless network complaints and building multi-turn dialogue assistant. Test results show that it not only significantly enhanced user experience but also greatly improved the efficiency of conducting data analysis, formulating solutions, and making decisions\footnote{https://www.53ai.com/news/zhinengkefu/2024121360978.html}. A primary concern for corporations considering implementing such systems remains data security, underscoring the need for continued research to mitigate this risk. This application naturally aligns with \textit{Context-Augmented Collaboration} (Agentic Workflows)~\cite{pilan2022text, yi2025ecoagent}, where the SM acts as an internal data retriever traversing private domains, sending sanitized or compressed \emph{prompts} to the remote LM to reduce leakage risk while outsourcing reasoning.

\textbf{Personalized Intelligence.} Despite ongoing efforts to optimize large language models for edge devices \cite{xu2024ondevicelanguagemodelscomprehensive}, significant compromises in efficiency and performance remain, necessitating collaborative solutions. While local small models can capture user portrait and preferences in a timely manner, a global foundation model contains enriched knowledge trained on extensive data sources. Harnessing the power of both global knowledge and personalized knowledge can help enable personalized intelligence in domains such as recommendation systems\cite{zhang2024federated}, consumption management ~\cite{li2024introducing}, and education. In order to deal with large-scale deployment for personal devices, future directions should focus on making the collaborative AI system scalable and robust, possibly through the integration into on-device FL system \cite{bonawitz2019towards}. One natural deployment pattern maps to \textit{Parameter-based Transfer}, where a global foundation model distributes personalized adapters (\eg, LoRA weights) to local edge devices, minimizing synchronization overhead ($\epsilon_e^{(comm)}$) while adhering to strict on-device privacy boundaries~\cite{fan2023fatellm, hu2021lora}.

\subsection{Continuous Learning and Task Adaptation}
\label{subsec:continuous_adaptation}
A further open question is how to make collaborative systems learn and adjust based on the evolution of contexts, model capabilities, tasks, and computational development. Existing studies mostly focus on static tasks, datasets, model architectures, and computational constraints. Fully realizing the potential of LM--SM collaboration requires effectively incorporating continuous learning, task adaptation, and dynamic resource allocation~\cite{chen2024adaptive}.

In LM--SM collaboration, this issue can be studied as carrier adaptation over time: the useful carrier evolves as private data distributions, user needs, retrieval memories, and resource conditions evolve, which turns carrier selection into a time-dependent problem~\cite{liu2024online, chen2024cascade, zhang2023fedpetuning, fan2023fatellm}. Systems using parameter-based transfer must decide when to update adapters or tunable prompts without accumulating excessive communication or privacy cost. Systems using Context-Augmented Collaboration must decide how agent memory, retrieval indexes, and tool summaries are updated while limiting leakage from repeated interactions~\cite{wang2025unveiling, chen2024agentpoison}. Systems using Collaborative Decoding or proxy-based adaptation must decide when the SM-side drafter, proxy, or emulator should be refreshed so that it remains aligned with the LM and the current domain~\cite{liu2024tuning, ormazabal2023comblm, mitchell2024emulator, liu2024online}.

The main challenge is to support adaptation without making each update an additional privacy, security, integrity, or efficiency burden. Repeated carrier updates can leak information about newly observed data, amplify membership or reconstruction risks, propagate poisoned memory or retrieval updates, and increase long-term LM query and communication costs~\cite{salem2020updates, wang2025unveiling, chen2024agentpoison}. Future work therefore needs temporal benchmarks that measure not only final task accuracy, but also adaptation speed, forgetting, cumulative leakage, integrity failures, and resource use over a sequence of changing tasks.


\section{Conclusion}
\label{sec:conclusion}

This survey reviewed LM--SM collaboration for private-domain tasks from a cross-silo perspective. We organized prior work by the carriers exchanged across trust boundaries, including LM-to-SM transfer, SM-to-LM transfer, inference-time collaboration, and their multi-party extensions, and summarized the privacy, security, integrity, and efficiency constraints governing the exchange. We identified promising future directions including multi-objective benchmarks, budget-aware protocols, application-grounded demonstrations, and methods for continuous adaptation. With the rapid advance of LM applications, we hope this survey provide a structured reference for developing LM--SM collaboration systems that are practical in private domains while remaining attentive to privacy, security, integrity, and resource limits.

\printbibliography

@inproceedings{liu2022automix,
  title     = {Automix: Unveiling The Power Of Mixup For Stronger Classifiers},
  author    = {Liu, Zicheng and Li, Siyuan and Wu, Di and Liu, Zihan and Chen, Zhiyuan and Wu, Lirong and Li, Stan Z},
  booktitle = {ECCV},
  year      = {2022}
}

@inproceedings{hoffmanntraining,
  title     = {Training Compute-Optimal Large Language Models},
  author    = {Hoffmann, Jordan and Borgeaud, Sebastian and Mensch, Arthur and Buchatskaya, Elena and Cai, Trevor and Rutherford, Eliza and de Las Casas, Diego and Hendricks, Lisa Anne and Welbl, Johannes and Clark, Aidan and others},
  booktitle = {NeurIPS},
  year      = {2022}
}

@article{weiemergent,
  title   = {Emergent Abilities Of Large Language Models},
  author  = {Wei, Jason and Tay, Yi and Bommasani, Rishi and Raffel, Colin and Zoph, Barret and Borgeaud, Sebastian and Yogatama, Dani and Bosma, Maarten and Zhou, Denny and Metzler, Donald and others},
  journal = {TMLR},
  year    = {2022}
}

@inproceedings{zhang2025poison,
  title     = {Poison as cure: Visual noise for mitigating object hallucinations in lvms},
  author    = {Zhang, Kejia and Tao, Keda and Tang, Jiasheng and Wang, Huan},
  booktitle = {NeurIPS},
  year      = {2025}
}

@inproceedings{song2025injecting,
  title     = {Injecting Domain-Specific Knowledge Into Large Language Models: A Comprehensive Survey},
  author    = {Song, Zirui and Yan, Bin and Liu, Yuhan and Fang, Miao and Li, Mingzhe and Yan, Rui and Chen, Xiuying},
  booktitle = {EMNLP Findings},
  year      = {2025}
}

@inproceedings{he2019model,
  title     = {Model Inversion Attacks Against Collaborative Inference},
  author    = {He, Zecheng and Zhang, Tianwei and Lee, Ruby B},
  booktitle = {ACSAC},
  year      = {2019}
}

@article{deepseekai2024deepseekv3technicalreport,
  title   = {Deepseek-V3 Technical Report},
  author  = {Liu, Aixin and Feng, Bei and Xue, Bing and Wang, Bingxuan and Wu, Bochao and Lu, Chengda and Zhao, Chenggang and Deng, Chengqi and Zhang, Chenyu and Ruan, Chong and others},
  journal = {arXiv preprint arXiv:2412.19437},
  year    = {2024}
}

@inproceedings{wu2024fedbiot,
  title     = {FedBioT: LLM Local Fine-Tuning In Federated Learning Without Full Model},
  author    = {Wu, Feijie and Li, Zitao and Li, Yaliang and Ding, Bolin and Gao, Jing},
  booktitle = {KDD},
  year      = {2024}
}

@inproceedings{han2024amd,
  title     = {Amd: Automatic Multi-Step Distillation Of Large-Scale Vision Models},
  author    = {Han, Cheng and Wang, Qifan and Dianat, Sohail A and Rabbani, Majid and Rao, Raghuveer M and Fang, Yi and Guan, Qiang and Huang, Lifu and Liu, Dongfang},
  booktitle = {ECCV},
  year      = {2024}
}

@inproceedings{melis2019exploiting,
  title     = {Exploiting Unintended Feature Leakage In Collaborative Learning},
  author    = {Melis, Luca and Song, Congzheng and De Cristofaro, Emiliano and Shmatikov, Vitaly},
  booktitle = {SP},
  year      = {2019}
}

@inproceedings{zhang2023crash,
  title     = {Crash: Clustering, Removing, And Sharing Enhance Fine-Tuning Without Full Large Language Model},
  author    = {Zhang, Kai-yan and Ding, Ning and Qi, Biqing and Zhu, Xuekai and Long, Xinwei and Zhou, Bowen},
  booktitle = {EMNLP},
  year      = {2023}
}

@inproceedings{takahashi2023breaching,
  title     = {Breaching FedMD: Image Recovery Via Paired-Logits Inversion Attack},
  author    = {Takahashi, Hideaki and Liu, Jingjing and Liu, Yang},
  booktitle = {CVPR},
  year      = {2023}
}

@inproceedings{duan2023privacy,
  title     = {On The Privacy Risk Of In-Context Learning},
  author    = {Duan, Haonan and Dziedzic, Adam and Yaghini, Mohammad and Papernot, Nicolas and Boenisch, Franziska},
  booktitle = {ACL},
  year      = {2023}
}

@inproceedings{birch2023model,
  title     = {Model Leeching: An Extraction Attack Targeting LLMs},
  author    = {Birch, Lewis and Hackett, William and Trawicki, Stefan and Suri, Neeraj and Garraghan, Peter},
  booktitle = {CAMLIS},
  year      = {2023}
}

@article{nasr2023scalable,
  title   = {Scalable Extraction Of Training Data From (Production) Language Models},
  author  = {Nasr, Milad and Carlini, Nicholas and Hayase, Jonathan and Jagielski, Matthew and Cooper, A Feder and Ippolito, Daphne and Choquette-Choo, Christopher A and Wallace, Eric and Tram{\`e}r, Florian and Lee, Katherine},
  journal = {arXiv preprint arXiv:2311.17035},
  year    = {2023}
}

@inproceedings{liu2024evolving,
  title     = {Evolving Knowledge Distillation With Large Language Models And Active Learning},
  author    = {Liu, Chengyuan and Zhao, Fubang and Kuang, Kun and Kang, Yangyang and Jiang, Zhuoren and Sun, Changlong and Wu, Fei},
  booktitle = {COLING},
  year      = {2024}
}

@article{dong2024guardrails,
  title   = {Building Guardrails For Large Language Models},
  author  = {Dong, Yi and Mu, Ronghui and Jin, Gaojie and Qi, Yi and Hu, Jinwei and Zhao, Xingyu and Meng, Jie and Ruan, Wenjie and Huang, Xiaowei},
  journal = {arXiv preprint arXiv:2402.01822},
  year    = {2024}
}

@inproceedings{zou2024fusegen,
  title     = {Fusegen: PLM Fusion For Data-Generation Based Zero-Shot Learning},
  author    = {Zou, Tianyuan and Liu, Yang and Li, Peng and Zhang, Jianqing and Liu, Jingjing and Zhang, Ya-Qin},
  booktitle = {EMNLP},
  year      = {2024}
}

@inproceedings{ye2022progen,
  title     = {Progen: Progressive Zero-Shot Dataset Generation Via In-Context Feedback},
  author    = {Ye, Jiacheng and Gao, Jiahui and Wu, Zhiyong and Feng, Jiangtao and Yu, Tao and Kong, Lingpeng},
  booktitle = {EMNLP Findings},
  year      = {2022}
}

@inproceedings{gholami2024gold,
  title     = {GOLD: Generalized Knowledge Distillation Via Out-Of-Distribution-Guided Language Data Generation},
  author    = {Gholami, Mohsen and Akbari, Mohammad and Hu, Tianxi and Masrani, Vaden and Wang, Z and Zhang, Yong},
  booktitle = {NAACL Findings},
  year      = {2024}
}

@inproceedings{latouche2024zero,
  title     = {Zero-Shot Cross-Lingual Transfer For Synthetic Data Generation In Grammatical Error Detection},
  author    = {Latouche, Gaetan and Carbonneau, Marc-Andr{\'e} and Swanson, Ben},
  booktitle = {EMNLP},
  year      = {2024}
}

@article{woo2025synthetic,
  title   = {Synthetic Data Distillation Enables The Extraction Of Clinical Information At Scale},
  author  = {Woo, Elizabeth Geena and Burkhart, Michael C and Alsentzer, Emily and Beaulieu-Jones, Brett K},
  journal = {npj Digital Medicine},
  year    = {2025},
  volume  = {8},
  number  = {1},
  pages   = {267}
}

@inproceedings{li2025learning,
  title     = {Learning With Less: Knowledge Distillation From Large Language Models Via Unlabeled Data},
  author    = {Li, Juanhui and Nag, Sreyashi and Liu, Hui and Tang, Xianfeng and Sarwar, Sheikh Muhammad and Cui, Limeng and Gu, Hansu and Wang, Suhang and He, Qi and Tang, Jiliang},
  booktitle = {NAACL Findings},
  year      = {2025}
}

@inproceedings{liu2023retrieval,
  title     = {Retrieval-Based Knowledge Transfer: An Effective Approach For Extreme Large Language Model Compression},
  author    = {Liu, Jiduan and Liu, Jiahao and Wang, Qifan and Wang, Jingang and Cai, Xunliang and Zhao, Dongyan and Wang, Ran and Yan, Rui},
  booktitle = {EMNLP Findings},
  year      = {2023}
}

@inproceedings{gao2023self,
  title     = {Self-Guided Noise-Free Data Generation For Efficient Zero-Shot Learning},
  author    = {Gao, Jiahui and Pi, Renjie and Yong, Lin and Xu, Hang and Ye, Jiacheng and Wu, Zhiyong and Zhang, Weizhong and Liang, Xiaodan and Li, Zhenguo and Kong, Lingpeng},
  booktitle = {ICLR},
  year      = {2023}
}

@inproceedings{song2020information,
  title     = {Information Leakage In Embedding Models},
  author    = {Song, Congzheng and Raghunathan, Ananth},
  booktitle = {CCS},
  year      = {2020}
}

@inproceedings{zhang2023fedpetuning,
  title     = {FedPETuning: When Federated Learning Meets The Parameter-Efficient Tuning Methods Of Pre-Trained Language Models},
  author    = {Zhang, Zhuo and Yang, Yuanhang and Dai, Yong and Wang, Qifan and Yu, Yue and Qu, Lizhen and Xu, Zenglin},
  booktitle = {ACL},
  year      = {2023}
}

@inproceedings{wu2025model,
  title     = {Model-Based Large Language Model Customization As Service},
  author    = {Wu, Zhaomin and Guo, Jizhou and Hou, Junyi and He, Bingsheng and Fan, Lixin and Yang, Qiang},
  booktitle = {EMNLP},
  year      = {2025}
}

@inproceedings{gaodata,
  title     = {Data-Adaptive Differentially Private Prompt Synthesis For In-Context Learning},
  author    = {Gao, Fengyu and Zhou, Ruida and Wang, Tianhao and Shen, Cong and Yang, Jing},
  booktitle = {ICLR},
  year      = {2025}
}

@inproceedings{yao2025gradot,
  title     = {GradOT: Training-Free Gradient-Preserving Offsite-Tuning For Large Language Models},
  author    = {Yao, Kai and Tan, Zhaorui and Gao, Penglei and Li, Lichun and Wu, Kaixin and Wang, Yinggui and Zhao, Yuan and Ji, Yixin and Zhu, Jianke and Wang, Wei},
  booktitle = {ACL},
  year      = {2025}
}

@inproceedings{yao2025scaleot,
  title     = {ScaleOT: Privacy-Utility-Scalable Offsite-Tuning With Dynamic LayerReplace And Selective Rank Compression},
  author    = {Yao, Kai and Tan, zhaorui and Ye, Tiandi and Li, Lichun and Zhao, Yuan and Liu, Wenyan and Wang, Wei and Zhu, Jianke},
  booktitle = {AAAI},
  year      = {2025}
}

@inproceedings{zhu2019deep,
  title     = {Deep Leakage From Gradients},
  author    = {Zhu, Ligeng and Liu, Zhijian and Han, Song},
  booktitle = {NeurIPS},
  year      = {2019}
}

@inproceedings{zhongseeking,
  title     = {Seeking Neural Nuggets: Knowledge Transfer In Large Language Models From A Parametric Perspective},
  author    = {Zhong, Ming and An, Chenxin and Chen, Weizhu and Han, Jiawei and He, Pengcheng},
  booktitle = {ICLR},
  year      = {2024}
}

@inproceedings{huostquant,
  title     = {Ostquant: Refining Large Language Model Quantization With Orthogonal And Scaling Transformations For Better Distribution Fitting},
  author    = {Hu, Xing and Cheng, Yuan and Yang, Dawei and Chen, Zhixuan and Xu, Zukang and Yuan, Zhihang and Zhou, Sifan and others},
  booktitle = {ICLR},
  year      = {2025}
}

@inproceedings{liao2024instance,
  title     = {From Instance Training To Instruction Learning: Task Adapters Generation From Instructions},
  author    = {Liao, Huanxuan and He, Shizhu and Xu, Yao and Zhang, Yuanzhe and Hao, Yanchao and Liu, Shengping and Liu, Kang and Zhao, Jun},
  booktitle = {NeurIPS},
  year      = {2024}
}

@inproceedings{tang2024direct,
  title     = {Direct Distillation Between Different Domains},
  author    = {Tang, Jialiang and Chen, Shuo and Niu, Gang and Zhu, Hongyuan and Zhou, Joey Tianyi and Gong, Chen and Sugiyama, Masashi},
  booktitle = {ECCV},
  year      = {2024}
}

@inproceedings{hu2025trimllm,
  title     = {TrimLLM: Progressive Layer Dropping For Domain-Specific LLMs},
  author    = {Hu, Lanxiang and Rosing, Tajana and Zhang, Hao},
  booktitle = {ACL},
  year      = {2025}
}

@inproceedings{shao2025context,
  title     = {In-Context Meta LoRA Generation},
  author    = {Shao, Yihua and Yan, Minxi and Liu, Yang and Chen, Siyu and Chen, Wenjie and Long, Xinwei and Yan, Ziyang and Li, Lei and Zhang, Chenyu and Sebe, Nicu and others},
  booktitle = {IJCAI},
  year      = {2025}
}

@inproceedings{dong2025mlas,
  title     = {Mlas-LoRA: Language-Aware Parameters Detection And LoRA-Based Knowledge Transfer For Multilingual Machine Translation},
  author    = {Dong, Tianyu and Li, Bo and Liu, Jinsong and Zhu, Shaolin and Xiong, Deyi},
  booktitle = {ACL},
  year      = {2025}
}

@inproceedings{ling2024slimgpt,
  title     = {Slimgpt: Layer-Wise Structured Pruning For Large Language Models},
  author    = {Ling, Gui and Wang, Ziyang and Liu, Qingwen},
  booktitle = {NeurIPS},
  year      = {2024}
}

@inproceedings{xuinitializing,
  title     = {Initializing Models With Larger Ones},
  author    = {Xu, Zhiqiu and Chen, Yanjie and Vishniakov, Kirill and Yin, Yida and Shen, Zhiqiang and Darrell, Trevor and Liu, Lingjie and Liu, Zhuang},
  booktitle = {ICLR},
  year      = {2024}
}

@inproceedings{sunsimple,
  title     = {A Simple And Effective Pruning Approach For Large Language Models},
  author    = {Sun, Mingjie and Liu, Zhuang and Bair, Anna and Kolter, J Zico},
  booktitle = {ICLR},
  year      = {2024}
}

@inproceedings{yao2022zeroquant,
  title     = {ZeroQuant: Efficient And Affordable Post-Training Quantization For Large-Scale Transformers},
  author    = {Yao, Zhewei and Yazdani Aminabadi, Reza and Zhang, Minjia and Wu, Xiaoxia and Li, Conglong and He, Yuxiong},
  booktitle = {NeurIPS},
  year      = {2022}
}

@article{desalvo2024no,
  title   = {No More Hard Prompts: SoftSRV Prompting For Synthetic Data Generation},
  author  = {DeSalvo, Giulia and Kagy, Jean-Fracois and Karydas, Lazaros and Rostamizadeh, Afshin and Kumar, Sanjiv},
  journal = {arXiv preprint arXiv:2410.16534},
  year    = {2024}
}

@inproceedings{mironov2017renyi,
  title     = {Renyi Differential Privacy},
  author    = {Mironov, Ilya},
  booktitle = {CSF},
  year      = {2017}
}

@inproceedings{morris2024language,
  title     = {Language Model Inversion},
  author    = {Morris, John Xavier and Zhao, Wenting and Chiu, Justin T. and Shmatikov, Vitaly and Rush, Alexander M.},
  booktitle = {ICLR},
  year      = {2024}
}

@inproceedings{wang2025unveiling,
  title     = {Unveiling Privacy Risks In LLM Agent Memory},
  author    = {Wang, Bo and He, Weiyi and Zeng, Shenglai and Xiang, Zhen and Xing, Yue and Tang, Jiliang and He, Pengfei},
  booktitle = {ACL},
  year      = {2025}
}

@inproceedings{zhang2021parameterized,
  title     = {Parameterized Knowledge Transfer For Personalized Federated Learning},
  author    = {Zhang, Jie and Guo, Song and Ma, Xiaosong and Wang, Haozhao and Xu, Wenchao and Wu, Feijie},
  booktitle = {NeurIPS},
  year      = {2021}
}

@inproceedings{zhang2024fedtgp,
  title     = {FedTGP: Trainable Global Prototypes With Adaptive-Margin-Enhanced Contrastive Learning For Data And Model Heterogeneity In Federated Learning},
  author    = {Zhang, Jianqing and Liu, Yang and Hua, Yang and Cao, Jian},
  booktitle = {AAAI},
  year      = {2024}
}

@inproceedings{fang2022robust,
  title     = {Robust Federated Learning With Noisy And Heterogeneous Clients},
  author    = {Fang, Xiuwen and Ye, Mang},
  booktitle = {CVPR},
  year      = {2022}
}

@inproceedings{ye2022zerogen,
  title     = {ZeroGen: Efficient Zero-Shot Learning Via Dataset Generation},
  author    = {Ye, Jiacheng and Gao, Jiahui and Li, Qintong and Xu, Hang and Feng, Jiangtao and Wu, Zhiyong and Yu, Tao and Kong, Lingpeng},
  booktitle = {EMNLP},
  year      = {2022}
}

@inproceedings{meng2022generating,
  title     = {Generating Training Data With Language Models: Towards Zero-Shot Language Understanding},
  author    = {Meng, Yu and Huang, Jiaxin and Zhang, Yu and Han, Jiawei},
  booktitle = {NeurIPS},
  year      = {2022}
}

@inproceedings{yang2024depth,
  title     = {Depth Anything: Unleashing The Power Of Large-Scale Unlabeled Data},
  author    = {Yang, Lihe and Kang, Bingyi and Huang, Zilong and Xu, Xiaogang and Feng, Jiashi and Zhao, Hengshuang},
  booktitle = {CVPR},
  year      = {2024}
}

@inproceedings{zhang2024fedgmkd,
  title     = {FedGMKD: An Efficient Prototype Federated Learning Framework Through Knowledge Distillation And Discrepancy-Aware Aggregation},
  author    = {Zhang, Jian-qiao and Shan, Cai-feng and Han, Jungong},
  booktitle = {NeurIPS},
  year      = {2024}
}

@inproceedings{qi2025cross,
  title     = {Cross-Silo Feature Space Alignment For Federated Learning On Clients With Imbalanced Data},
  author    = {Qi, Zhuang and Meng, Lei and Li, Zhaochuan and Hu, Han and Meng, Xiangxu},
  booktitle = {AAAI},
  year      = {2025}
}

@article{wu2024global,
  title   = {Global Prototype Distillation For Heterogeneous Federated Learning},
  author  = {Wu, Shu and others},
  journal = {Scientific Reports},
  year    = {2024},
  volume  = {14},
  number  = {1},
  pages   = {12057}
}

@article{guo2024lud,
  title   = {Leveraging Logit Uncertainty For Better Knowledge Distillation},
  author  = {Guo, Zhen and Wang, Dong and He, Qiang and Zhang, Pengzhou},
  journal = {Scientific Reports},
  year    = {2024},
  volume  = {14},
  number  = {1},
  pages   = {31249}
}

@inproceedings{zhang2024one,
  title     = {One-Shot Federated Learning Via Synthetic Distiller-Distillate Communication},
  author    = {Zhang, Junyuan and Liu, Songhua and Wang, Xinchao},
  booktitle = {NeurIPS},
  year      = {2024}
}

@inproceedings{jin2025fedcpd,
  title     = {FedCPD: Personalized Federated Learning With Prototype-Enhanced Representation And Memory Distillation},
  author    = {Jin, Kaili and Xu, Li and Wang, Xiaoding and Hsieh, Sun-Yuan and Wu, Jie and Lin, Limei},
  booktitle = {IJCAI},
  year      = {2025}
}

@inproceedings{zhang2022ideal,
  title     = {Ideal: Query-Efficient Data-Free Learning From Black-Box Models},
  author    = {Zhang, Jie and others},
  booktitle = {ICLR},
  year      = {2023}
}

@inproceedings{xu2025slmrec,
  title     = {Slmrec: Distilling Large Language Models Into Small For Sequential Recommendation},
  author    = {Xu, Wujiang and Wu, Qitian and Liang, Zujie and Han, Jiaojiao and Ning, Xuying and Shi, Yunxiao and Lin, Wenfang and Zhang, Yongfeng},
  booktitle = {ICLR},
  year      = {2025}
}

@inproceedings{xiang2025evidential,
  title     = {Evidential Knowledge Distillation},
  author    = {Xiang, Liangyu and Gao, Junyu and Xu, Changsheng},
  booktitle = {ICCV},
  year      = {2025}
}

@inproceedings{sanyal2022towards,
  title     = {Towards Data-Free Model Stealing In A Hard Label Setting},
  author    = {Sanyal, Sunandini and Addepalli, Sravanti and Babu, R Venkatesh},
  booktitle = {CVPR},
  year      = {2022}
}

@inproceedings{wang2021zero,
  title     = {Zero-Shot Knowledge Distillation From A Decision-Based Black-Box Model},
  author    = {Wang, Zi},
  booktitle = {ICML},
  year      = {2021}
}

@inproceedings{zhou2023bridging,
  title     = {Bridging The Gap Between Decision And Logits In Decision-Based Knowledge Distillation For Pre-Trained Language Models},
  author    = {Zhou, Qinhong and Yang, Zonghan and Li, Peng and Liu, Yang},
  booktitle = {ACL},
  year      = {2023}
}

@inproceedings{gong2022preserving,
  title     = {Preserving Privacy In Federated Learning With Ensemble Cross-Domain Knowledge Distillation},
  author    = {Gong, Xuan and Sharma, Abhishek and Karanam, Srikrishna and Wu, Ziyan and Chen, Terrence and Doermann, David and Innanje, Arun},
  booktitle = {AAAI},
  year      = {2022}
}

@inproceedings{li2019fedmd,
  title     = {FedMD: Heterogenous Federated Learning Via Model Distillation},
  author    = {Li, Daliang and Wang, Junpu},
  booktitle = {NeurIPS Workshop},
  year      = {2019}
}

@inproceedings{peng2024fedpft,
  title     = {FedPFT: Federated Proxy Fine-Tuning Of Foundation Models},
  author    = {Peng, Zhaopeng and Fan, Xiaoliang and Chen, Yufan and Wang, Zheng and Pan, Shirui and Wen, Chenglu and Zhang, Ruisheng and Wang, Cheng},
  booktitle = {IJCAI},
  year      = {2024}
}

@article{xiao2023offsite,
  title   = {Offsite-Tuning: Transfer Learning Without Full Model},
  author  = {Xiao, Guangxuan and Lin, Ji and Han, Song},
  journal = {arXiv preprint arXiv:2302.04870},
  year    = {2023}
}

@inproceedings{zhou2022bert,
  title     = {BERT Learns To Teach: Knowledge Distillation With Meta Learning},
  author    = {Zhou, Wangchunshu and Xu, Canwen and McAuley, Julian},
  booktitle = {ACL},
  year      = {2022}
}

@inproceedings{guminillm,
  title     = {MiniLLM: Knowledge Distillation Of Large Language Models},
  author    = {Gu, Yuxian and Dong, Li and Wei, Furu and Huang, Minlie},
  booktitle = {ICLR},
  year      = {2024}
}

@article{xu2024llavadi,
  title   = {LLaVADI: What Matters For Multimodal Large Language Models Distillation},
  author  = {Xu, Shilin and Li, Xiangtai and Yuan, Haobo and Qi, Lu and Tong, Yunhai and Yang, Ming-Hsuan},
  journal = {arXiv preprint arXiv:2407.19409},
  year    = {2024}
}

@inproceedings{agarwal2024policy,
  title     = {On-Policy Distillation Of Language Models: Learning From Self-Generated Mistakes},
  author    = {Agarwal, Rishabh and Vieillard, Nino and Zhou, Yongchao and Stanczyk, Piotr and Garea, Sabela Ramos and Geist, Matthieu and Bachem, Olivier},
  booktitle = {ICLR},
  year      = {2024}
}

@article{wang2024comprehensive,
  title   = {A Comprehensive Survey Of Small Language Models In The Era Of Large Language Models: Techniques, Enhancements, Applications, Collaboration With LLMs, And Trustworthiness},
  author  = {Wang, Fali and Zhang, Zhiwei and Zhang, Xianren and Wu, Zongyu and Mo, Tzuhao and Lu, Qiuhao and Wang, Wanjing and Li, Rui and Xu, Junjie and Tang, Xianfeng and others},
  journal = {ACM TIST},
  year    = {2024},
  volume  = {16},
  number  = {6}
}

@article{chen2025survey,
  title   = {A Survey On Collaborative Mechanisms Between Large And Small Language Models},
  author  = {Chen, Yi and Zhao, JiaHao and Han, HaoHao},
  journal = {arXiv preprint arXiv:2505.07460},
  year    = {2025}
}

@article{wang2025survey,
  title   = {A Survey On Collaborating Small And Large Language Models For Performance, Cost-Effectiveness, Cloud-Edge Privacy, And Trustworthiness},
  author  = {Wang, Fali and Chen, Jihai and Yang, Shuhua and Al-Lawati, Ali and Tang, Linli and Liu, Hui and Wang, Suhang},
  journal = {arXiv preprint arXiv:2510.13890},
  year    = {2025}
}

@article{xu2024survey,
  title   = {A Survey Of Resource-Efficient LLM And Multimodal Foundation Models},
  author  = {Xu, Mengwei and Yin, Wangsong and Cai, Dongqi and Yi, Rongjie and Xu, Daliang and Wang, Qipeng and Wu, Bingyang and Zhao, Yihao and Yang, Chen and Wang, Shihe and others},
  journal = {arXiv preprint arXiv:2401.08092},
  year    = {2024}
}

@article{menghani2023efficient,
  title   = {Efficient Deep Learning: A Survey On Making Deep Learning Models Smaller, Faster, And Better},
  author  = {Menghani, Gaurav},
  journal = {ACM Computing Surveys},
  year    = {2023},
  volume  = {55},
  number  = {12},
  pages   = {1--37}
}

@article{vepakomma2018split,
  title   = {Split Learning For Health: Distributed Deep Learning Without Sharing Raw Patient Data},
  author  = {Vepakomma, Praneeth and Gupta, Otkrist and Swedish, Tristan and Raskar, Ramesh},
  journal = {arXiv preprint arXiv:1812.00564},
  year    = {2018}
}

@inproceedings{dong2023tunable,
  title     = {Tunable Soft Prompts Are Messengers In Federated Learning},
  author    = {Dong, Chenhe and Xie, Yuexiang and Ding, Bolin and Shen, Ying and Li, Yaliang},
  booktitle = {EMNLP Findings},
  year      = {2023}
}

@article{gou2021knowledge,
  title   = {Knowledge Distillation: A Survey},
  author  = {Gou, Jianping and Yu, Baosheng and Maybank, Stephen J and Tao, Dacheng},
  journal = {IJCV},
  year    = {2021},
  volume  = {129},
  pages   = {1789--1819}
}

@inproceedings{chen2024datashunt,
  title     = {Data Shunt: Collaboration Of Small And Large Models For Lower Costs And Better Performance},
  author    = {Chen, Dong and Zhuang, Yueting and Zhang, Shuo and Liu, Jinfeng and Dong, Su and Tang, Siliang},
  booktitle = {AAAI},
  year      = {2024}
}

@inproceedings{hsieh2023distilling,
  title     = {Distilling Step-By-Step! Outperforming Larger Language Models With Less Training Data And Smaller Model Sizes},
  author    = {Hsieh, Cheng-yu and Li, Chun-liang and Yeh, Chih-kuan and Nakhost, Hootan and Fujii, Yasuhisa and Ratner, Alex and Krishna, Ranjay and Lee, Chen-yu and Pfister, Tomas},
  booktitle = {ACL Findings},
  year      = {2023}
}

@article{hinton2015distilling,
  title   = {Distilling The Knowledge In A Neural Network},
  author  = {Hinton, Geoffrey and Vinyals, Oriol and Dean, Jeff},
  journal = {arXiv preprint arXiv:1503.02531},
  year    = {2015}
}

@article{yang2024learning,
  title   = {Learning From Human Educational Wisdom: A Student-Centered Knowledge Distillation Method},
  author  = {Yang, Shunzhi and Yang, Jinfeng and Zhou, MengChu and Huang, Zhenhua and Zheng, Wei-Shi and Yang, Xiong and Ren, Jin},
  journal = {TPAMI},
  year    = {2024},
  volume  = {46},
  number  = {6},
  pages   = {4188--4205}
}

@inproceedings{tramer2016stealing,
  title     = {Stealing Machine Learning Models Via Prediction APIs},
  author    = {Tram{\`e}r, Florian and Zhang, Fan and Juels, Ari and Reiter, Michael K and Ristenpart, Thomas},
  booktitle = {USENIX Security},
  year      = {2016}
}

@article{dwork2014algorithmic,
  title   = {The Algorithmic Foundations Of Differential Privacy},
  author  = {Dwork, Cynthia and Roth, Aaron and others},
  journal = {Foundations and Trends{\textregistered} in Theoretical Computer Science},
  year    = {2014},
  volume  = {9},
  number  = {3--4},
  pages   = {211--407}
}

@inproceedings{bun2016concentrated,
  title     = {Concentrated Differential Privacy: Simplifications, Extensions, And Lower Bounds},
  author    = {Bun, Mark and Steinke, Thomas},
  booktitle = {TCC},
  year      = {2016}
}

@inproceedings{abadi2016deep,
  title     = {Deep Learning With Differential Privacy},
  author    = {Abadi, Martin and Chu, Andy and Goodfellow, Ian and McMahan, H Brendan and Mironov, Ilya and Talwar, Kunal and Zhang, Li},
  booktitle = {CCS},
  year      = {2016}
}

@inproceedings{chen2023frugalgpt,
  title     = {FrugalGPT: How To Use Large Language Models While Reducing Cost And Improving Performance},
  author    = {Chen, Lingjiao and Zaharia, Matei and Zou, James},
  booktitle = {NeurIPS},
  year      = {2023}
}

@inproceedings{leviathan2023fast,
  title     = {Fast Inference From Transformers Via Speculative Decoding},
  author    = {Leviathan, Yaniv and Kalman, Matan and Matias, Yossi},
  booktitle = {ICML},
  year      = {2023}
}

@article{ieee2025split,
  title   = {Split Fine-Tuning For Large Language Models In Wireless Networks},
  author  = {Zhang, Songge and Cheng, Guoliang and Wu, Wen and Huang, Xinyu and Song, Lingyang and Shen, Xuemin},
  journal = {IEEE JSTSP},
  year    = {2025}
}

@inproceedings{liu2025ecolora,
  title     = {EcoLoRA: Communication-Efficient Federated Fine-Tuning Of Large Language Models},
  author    = {Liu, Han and Wen, Ruoyao and Nair, Srijith and Liu, Jia and Lou, Wenjing and Zhang, Chongjie and Yeoh, William and Vorobeychik, Yevgeniy and Zhang, Ning},
  booktitle = {EMNLP},
  year      = {2025}
}

@inproceedings{lin2020mcunet,
  title     = {Mcunet: Tiny Deep Learning On Iot Devices},
  author    = {Lin, Ji and Chen, Wei-Ming and Lin, Yujun and Gan, Chuang and Han, Song and others},
  booktitle = {NeurIPS},
  year      = {2020}
}

@inproceedings{alizadeh2024llm,
  title     = {LLM In A Flash: Efficient Large Language Model Inference With Limited Memory},
  author    = {Alizadeh, Keivan and Mirzadeh, Seyed Iman and Belenko, Dmitry and Khatamifard, S and Cho, Minsik and Del Mundo, Carlo C and Rastegari, Mohammad and Farajtabar, Mehrdad},
  booktitle = {ACL},
  year      = {2024}
}

@inproceedings{mcmahan2017communication,
  title     = {Communication-Efficient Learning Of Deep Networks From Decentralized Data},
  author    = {McMahan, Brendan and Moore, Eider and Ramage, Daniel and Hampson, Seth and y Arcas, Blaise Aguera},
  booktitle = {AISTATS},
  year      = {2017}
}

@article{sha2024prompt,
  title   = {Prompt Stealing Attacks Against Large Language Models},
  author  = {Sha, Zeyang and Zhang, Yang},
  journal = {arXiv preprint arXiv:2402.12959},
  year    = {2024}
}

@article{kairouz2021advances,
  title   = {Advances And Open Problems In Federated Learning},
  author  = {Kairouz, Peter and McMahan, H Brendan},
  journal = {Foundations and trends in machine learning},
  year    = {2021},
  volume  = {14},
  number  = {1-2},
  pages   = {1--210}
}

@inproceedings{orekondy2019knockoff,
  title     = {Knockoff Nets: Stealing Functionality Of Black-Box Models},
  author    = {Orekondy, Tribhuvanesh and Schiele, Bernt and Fritz, Mario},
  booktitle = {CVPR},
  year      = {2019}
}

@inproceedings{driouich2022attribute,
  title     = {A Novel Model-Based Attribute Inference Attack In Federated Learning},
  author    = {Driouich, Ilias and Xu, Chuan and Neglia, Giovanni and Giroire, Frederic and Thomas, Eoin},
  booktitle = {NeurIPS Workshop},
  year      = {2022}
}

@article{liu2024vertical,
  title   = {Vertical Federated Learning: Concepts, Advances, And Challenges},
  author  = {Liu, Yang and Kang, Yan and Zou, Tianyuan and Pu, Yanhong and He, Yuanqin and Ye, Xiaozhou and Ouyang, Ye and Zhang, Ya-Qin and Yang, Qiang},
  journal = {TKDE},
  year    = {2024},
  volume  = {36},
  number  = {7},
  pages   = {3615--3634}
}

@inproceedings{liang2023less,
  title     = {Less Is More: Task-Aware Layer-Wise Distillation For Language Model Compression},
  author    = {Liang, Chen and Zuo, Simiao and Zhang, Qingru and He, Pengcheng and Chen, Weizhu and Zhao, Tuo},
  booktitle = {ICML},
  year      = {2023}
}

@inproceedings{carlini2021extracting,
  author    = {Nicholas Carlini and Florian Tram{\`e}r and Eric Wallace and Matthew Jagielski and Ariel Herbert-Voss and Katherine Lee and Adam Roberts and Tom Brown and Dawn Song and {\'U}lfar Erlingsson and Alina Oprea and Colin Raffel},
  title     = {Extracting Training Data from Large Language Models},
  booktitle = {USENIX Security},
  year      = {2021}
}

@inproceedings{kandpal2023large,
  title     = {Large Language Models Struggle To Learn Long-Tail Knowledge},
  author    = {Kandpal, Nikhil and Deng, Haikang and Roberts, Adam and Wallace, Eric and Raffel, Colin},
  booktitle = {ICML},
  year      = {2023}
}

@inproceedings{gururangan2020don,
  title     = {Don't Stop Pretraining: Adapt Language Models To Domains And Tasks},
  author    = {Gururangan, Suchin and Marasovi{\'c}, Ana and Swayamdipta, Swabha and Lo, Kyle and Beltagy, Iz and Downey, Doug and Smith, Noah A},
  booktitle = {ACL},
  year      = {2020}
}

@inproceedings{ong2024routellm,
  title     = {Routellm: Learning To Route LLMs With Preference Data},
  author    = {Ong, Isaac and Almahairi, Amjad and Wu, Vincent and Chiang, Wei-Lin and Wu, Tianhao and Gonzalez, Joseph E and Kadous, M Waleed and Stoica, Ion},
  booktitle = {ICLR},
  year      = {2024}
}

@inproceedings{xu2024recomp,
  title     = {Recomp: Improving Retrieval-Augmented LMs With Compression And Selective Augmentation},
  author    = {Xu, Fangyuan and Shi, Weijia and Choi, Eunsol},
  booktitle = {ICLR},
  year      = {2024}
}

@inproceedings{burns2024weak,
  title     = {Weak-To-Strong Generalization: Eliciting Strong Capabilities With Weak Supervision},
  author    = {Burns, Collin and others},
  booktitle = {ICML},
  year      = {2024}
}

@inproceedings{hendrycks2020measuring,
  title     = {Measuring Massive Multitask Language Understanding},
  author    = {Hendrycks, Dan and Burns, Collin and Basart, Steven and Zou, Andy and Mazeika, Mantas and Song, Dawn and Steinhardt, Jacob},
  booktitle = {ICLR},
  year      = {2020}
}

@article{Bakare2024DATAPL,
  title   = {Data Privacy Laws And Compliance: A Comparative Review Of The EU GDPR And USA Regulations},
  author  = {Seun Solomon Bakare and Adekunle Oyeyemi Adeniyi and Chidiogo Uzoamaka Akpuokwe and Nkechi Emmanuella Eneh},
  journal = {Computer Science \& IT Research Journal},
  year    = {2024}
}

@inproceedings{shokri2017membership,
  title     = {Membership Inference Attacks Against Machine Learning Models},
  author    = {Shokri, Reza and Stronati, Marco and Song, Congzheng and Shmatikov, Vitaly},
  booktitle = {SP},
  year      = {2017}
}

@inproceedings{dwork2006calibrating,
  title     = {Calibrating Noise To Sensitivity In Private Data Analysis},
  author    = {Dwork, Cynthia and McSherry, Frank and Nissim, Kobbi and Smith, Adam},
  booktitle = {TCC},
  year      = {2006}
}

@article{ren2024advancesopenchallengesfederated,
  title   = {Advances And Open Challenges In Federated Foundation Models},
  author  = {Ren, Chao and Yu, Han and Peng, Hongyi and Tang, Xiaoli and Zhao, Bo and Yi, Liping and Tan, Alysa Ziying and Gao, Yulan and Li, Anran and Li, Xiaoxiao and others},
  journal = {IEEE Communications Surveys \& Tutorials},
  year    = {2025}
}

@inproceedings{nasr2019comprehensive,
  title     = {Comprehensive Privacy Analysis Of Deep Learning: Passive And Active White-Box Inference Attacks Against Centralized And Federated Learning},
  author    = {Nasr, Milad and Shokri, Reza and Houmansadr, Amir},
  booktitle = {SP},
  year      = {2019}
}

@inproceedings{fredrikson2015model,
  title     = {Model Inversion Attacks That Exploit Confidence Information And Basic Countermeasures},
  author    = {Fredrikson, Matt and Jha, Somesh and Ristenpart, Thomas},
  booktitle = {CCS},
  year      = {2015}
}

@inproceedings{ganju2018property,
  title     = {Property Inference Attacks On Fully Connected Neural Networks Using Permutation Invariant Representations},
  author    = {Ganju, Karan and Wang, Qi and Yang, Wei and Gunter, Carl A and Borisov, Nikita},
  booktitle = {CCS},
  year      = {2018}
}

@inproceedings{liu2024mobilellm,
  title     = {Mobilellm: Optimizing Sub-Billion Parameter Language Models For On-Device Use Cases},
  author    = {Liu, Zechun and Zhao, Changsheng and Iandola, Forrest and Lai, Chen and Tian, Yuandong and Fedorov, Igor and Xiong, Yunyang and Chang, Ernie and Shi, Yangyang and Krishnamoorthi, Raghuraman and others},
  booktitle = {ICML},
  year      = {2024}
}

@inproceedings{zhang2024cogenesis,
  title     = {Cogenesis: A Framework Collaborating Large And Small Language Models For Secure Context-Aware Instruction Following},
  author    = {Zhang, Kaiyan and Wang, Jianyu and Hua, Ermo and Qi, Biqing and Ding, Ning and Zhou, Bowen},
  booktitle = {ACL},
  year      = {2024}
}

@article{li2024synergizing,
  title   = {Synergizing Foundation Models And Federated Learning: A Survey},
  author  = {Li, Shenghui and Ye, Fanghua and Fang, Meng and Zhao, Jiaxu and Chan, Yun-Hin and Ngai, Edith C-H and Voigt, Thiemo},
  journal = {arXiv preprint arXiv:2406.12844},
  year    = {2024}
}

@inproceedings{narasimhanfaster,
  title     = {Faster Cascades Via Speculative Decoding},
  author    = {Narasimhan, Harikrishna and Jitkrittum, Wittawat and Rawat, Ankit Singh and Kim, Seungyeon and Gupta, Neha and Menon, Aditya Krishna and Kumar, Sanjiv},
  booktitle = {ICLR},
  year      = {2025}
}

@inproceedings{bachmannjudge,
  title     = {Judge Decoding: Faster Speculative Sampling Requires Going Beyond Model Alignment},
  author    = {Bachmann, Gregor and Anagnostidis, Sotiris and Pumarola, Albert and Georgopoulos, Markos and Sanakoyeu, Artsiom and Du, Yuming and Sch{\"o}nfeld, Edgar and Thabet, Ali and Kohler, Jonas K},
  booktitle = {ICLR},
  year      = {2025}
}

@inproceedings{jiao2020tinybert,
  title     = {TinyBERT: Distilling BERT For Natural Language Understanding},
  author    = {Jiao, Xiaoqi and Yin, Yichun and Shang, Lifeng and Jiang, Xin and Chen, Xiao and Li, Linlin and Wang, Fang and Liu, Qun},
  booktitle = {EMNLP Findings},
  year      = {2020}
}

@article{touvron2023llama,
  title   = {Llama: Open And Efficient Foundation Language Models},
  author  = {Touvron, Hugo and Lavril, Thibaut and Izacard, Gautier and Martinet, Xavier and Lachaux, Marie-Anne and Lacroix, Timoth{\'e}e and Rozi{\`e}re, Baptiste and Goyal, Naman and Hambro, Eric and Azhar, Faisal and others},
  journal = {arXiv preprint arXiv:2302.13971},
  year    = {2023}
}

@article{meta2024llama3-1,
  title   = {Introducing Llama 3.1: Our Most Capable Models To Date},
  author  = {MetaAI},
  journal = {MetaAI},
  year    = {2024}
}

@inproceedings{li2023llavamed,
  title     = {LLaVA-Med: Training A Large Language-And-Vision Assistant For Biomedicine In One Day},
  author    = {Li, Chunyuan and Wong, Cliff and Zhang, Sheng and Usuyama, Naoto and Liu, Haotian and Yang, Jianwei and Naumann, Tristan and Poon, Hoifung and Gao, Jianfeng},
  booktitle = {NeurIPS},
  year      = {2023}
}

@article{abdin2024phi,
  title   = {Phi-3 Technical Report: A Highly Capable Language Model Locally On Your Phone},
  author  = {Abdin, Marah and Jacobs, Sam Ade and Awan, Ammar Ahmad and Aneja, Jyoti and Awadallah, Ahmed and Awadalla, Hany and Bach, Nguyen and Bahree, Amit and Bakhtiari, Arash and Behl, Harkirat and others},
  journal = {arXiv preprint arXiv:2404.14219},
  year    = {2024}
}

@article{javaheripi2023phi,
  title   = {Phi-2: The Surprising Power Of Small Language Models},
  author  = {Javaheripi, Mojan and Bubeck, S{\'e}bastien and Abdin, Marah and Aneja, Jyoti and Bubeck, Sebastien and Mendes, Caio C{\'e}sar Teodoro and Chen, Weizhu and Del Giorno, Allie and Eldan, Ronen and Gopi, Sivakanth and others},
  journal = {Microsoft Research Blog},
  year    = {2023}
}

@article{kaplan2020scaling,
  title   = {Scaling Laws For Neural Language Models},
  author  = {Kaplan, Jared and McCandlish, Sam and Henighan, Tom and Brown, Tom B and Chess, Benjamin and Child, Rewon and Gray, Scott and Radford, Alec and Wu, Jeffrey and Amodei, Dario},
  journal = {arXiv preprint arXiv:2001.08361},
  year    = {2020}
}

@inproceedings{xia2024unlocking,
  title     = {Unlocking Efficiency In Large Language Model Inference: A Comprehensive Survey Of Speculative Decoding},
  author    = {Heming Xia and Zhe Yang and Qingxiu Dong and Peiyi Wang and Yongqi Li and Tao Ge and Tianyu Liu and Wenjie Li and Zhifang Sui},
  booktitle = {ACL},
  year      = {2024}
}

@inproceedings{chen2024cascade,
  title     = {Cascade Speculative Drafting For Even Faster LLM Inference},
  author    = {Chen, Zi-yi and Yang, Xiao-cong and Lin, Jia-cheng and Sun, Chen-kai and Chang, Kevin C and Huang, Jie},
  booktitle = {NeurIPS},
  year      = {2024}
}

@inproceedings{kim2024speculative,
  title     = {Speculative Decoding With Big Little Decoder},
  author    = {Kim, Sehoon and Mangalam, Karttikeya and Moon, Suhong and Malik, Jitendra and Mahoney, Michael W and Gholami, Amir and Keutzer, Kurt},
  booktitle = {NeurIPS},
  year      = {2024}
}

@inproceedings{sun2024spectr,
  title     = {Spectr: Fast Speculative Decoding Via Optimal Transport},
  author    = {Sun, Zi-teng and Suresh, Ananda Theertha and Ro, Jae Hun and Beirami, Ahmad and Jain, Himanshu and Yu, Felix},
  booktitle = {NeurIPS},
  year      = {2024}
}

@inproceedings{muennighoff2023scaling,
  title     = {Scaling Data-Constrained Language Models},
  author    = {Muennighoff, Niklas and Rush, Alexander and Barak, Boaz and Le Scao, Teven and Tazi, Nouamane and Piktus, Aleksandra and Pyysalo, Sampo and Wolf, Thomas and Raffel, Colin A},
  booktitle = {NeurIPS},
  year      = {2023}
}

@inproceedings{liu2024online,
  title     = {Online Speculative Decoding},
  author    = {Liu, Xiao-xuan and Hu, Lan-xiang and Bailis, Peter and Stoica, Ion and Deng, Zhi-jie and Cheung, Alvin and Zhang, Hao},
  booktitle = {ICML},
  year      = {2024}
}

@inproceedings{pan2024llmlingua,
  title     = {Llmlingua-2: Data Distillation For Efficient And Faithful Task-Agnostic Prompt Compression},
  author    = {Pan, Zhuoshi and Wu, Qianhui and Jiang, Huiqiang and Xia, Menglin and Luo, Xufang and Zhang, Jue and Lin, Qingwei and R{\"u}hle, Victor and Yang, Yuqing and Lin, Chin-Yew and others},
  booktitle = {ACL Findings},
  year      = {2024}
}

@inproceedings{zhou2023textobfuscator,
  title     = {Textobfuscator: Making Pre-Trained Language Model A Privacy Protector Via Obfuscating Word Representations},
  author    = {Zhou, Xin and Lu, Yi and Ma, Ruotian and Gui, Tao and Wang, Yuran and Ding, Yong and Zhang, Yibo and Zhang, Qi and Huang, Xuan-Jing},
  booktitle = {ACL Findings},
  year      = {2023}
}

@inproceedings{li2023protecting,
  title     = {Protecting Intellectual Property Of Large Language Model-Based Code Generation APIs Via Watermarks},
  author    = {Li, Zongjie and Wang, Chaozheng and Wang, Shuai and Gao, Cuiyun},
  booktitle = {CCS},
  year      = {2023}
}

@article{chu2024history,
  title   = {History, Development, And Principles Of Large Language Models-An Introductory Survey},
  author  = {Chu, Zhibo and Ni, Shiwen and Wang, Zichong and Feng, Xi and Li, Chengming and Hu, Xiping and Xu, Ruifeng and Yang, Min and Zhang, Wenbin},
  journal = {AI and Ethics},
  year    = {2025},
  volume  = {5},
  number  = {3},
  pages   = {1955--1971}
}

@article{zhou2023comprehensive,
  title   = {A Comprehensive Survey On Pretrained Foundation Models: A History From BERT To ChatGPT},
  author  = {Zhou, Ce and Li, Qian and Li, Chen and Yu, Jun and Liu, Yixin and Wang, Guangjing and Zhang, Kai and Ji, Cheng and Yan, Qiben and He, Lifang and others},
  journal = {International Journal of Machine Learning and Cybernetics},
  year    = {2025},
  volume  = {16},
  number  = {12},
  pages   = {9851--9915}
}

@inproceedings{wan2024knowledge,
  title     = {Knowledge Fusion Of Large Language Models},
  author    = {Wan, Fanqi and Huang, Xinting and Cai, Deng and Quan, Xiaojun and Bi, Wei and Shi, Shuming},
  booktitle = {ICLR},
  year      = {2024}
}

@inproceedings{wan2024fusechat,
  title     = {Fusechat: Knowledge Fusion Of Chat Models},
  author    = {Wan, Fanqi and Yang, Ziyi and Zhong, Longguang and Quan, Xiaojun and Huang, Xinting and Bi, Wei},
  booktitle = {EMNLP},
  year      = {2025}
}

@inproceedings{jin2023dataless,
  title     = {Dataless Knowledge Fusion By Merging Weights Of Language Models},
  author    = {Jin, Xisen and Ren, Xiang and Preotiuc-Pietro, Daniel and Cheng, Pengxiang},
  booktitle = {ICLR},
  year      = {2023}
}

@inproceedings{zhang2023composing,
  title     = {Composing Parameter-Efficient Modules With Arithmetic Operation},
  author    = {Zhang, Jinghan and Liu, Junteng and He, Junxian and others},
  booktitle = {NeurIPS},
  year      = {2023}
}

@inproceedings{tan2022fedproto,
  title     = {FedProto: Federated Prototype Learning Across Heterogeneous Clients},
  author    = {Tan, Yue and others},
  booktitle = {AAAI},
  year      = {2022}
}

@inproceedings{tan2022federated,
  title     = {Federated Learning From Pre-Trained Models: A Contrastive Learning Approach},
  author    = {Tan, Yue and Long, Guodong and Ma, Jie and Liu, Lu and Zhou, Tianyi and Jiang, Jing},
  booktitle = {NeurIPS},
  year      = {2022}
}

@inproceedings{erdougan2022unsplit,
  title     = {Unsplit: Data-Oblivious Model Inversion, Model Stealing, And Label Inference Attacks Against Split Learning},
  author    = {Erdo{\u{g}}an, Ege and K{\"u}p{\c{c}}{\"u}, Alptekin and {\c{C}}i{\c{c}}ek, A Erc{\"u}ment},
  booktitle = {ACM WPES},
  year      = {2022}
}

@inproceedings{geiping2020inverting,
  title     = {Inverting Gradients-How Easy Is It To Break Privacy In Federated Learning?},
  author    = {Geiping, Jonas and Bauermeister, Hartmut and Dr{\"o}ge, Hannah and Moeller, Michael},
  booktitle = {NeurIPS},
  year      = {2020}
}

@inproceedings{lyu-etal-2020-differentially,
  title     = {Differentially Private Representation For NLP: Formal Guarantee And An Empirical Study On Privacy And Fairness},
  author    = {Lyu, Lingjuan and He, Xuanli and Li, Yitong},
  booktitle = {EMNLP Findings},
  year      = {2020}
}

@inproceedings{li-etal-2018-towards,
  title     = {Towards Robust And Privacy-Preserving Text Representations},
  author    = {Li, Yitong and Baldwin, Timothy and Cohn, Trevor},
  booktitle = {ACL},
  year      = {2018}
}

@inproceedings{coavoux2018privacy,
  title     = {Privacy-Preserving Neural Representations Of Text},
  author    = {Coavoux, Maximin and Narayan, Shashi and Cohen, Shay B},
  booktitle = {EMNLP},
  year      = {2018}
}

@inproceedings{hu2021lora,
  title     = {LoRA: Low-Rank Adaptation Of Large Language Models},
  author    = {Hu, Edward J and Shen, Yelong and Wallis, Phillip and Allen-Zhu, Zeyuan and Li, Yuanzhi and Wang, Shean and Wang, Lu and Chen, Weizhu},
  booktitle = {ICLR},
  year      = {2022}
}

@inproceedings{woisetschlager2023federated,
  title     = {Federated Fine-Tuning Of LLMs On The Very Edge: The Good, The Bad, The Ugly},
  author    = {Woisetschl{\"a}ger, Herbert and Isenko, Alexander and Wang, Shiqiang and Mayer, Ruben and Jacobsen, Hans-Arno},
  booktitle = {Workshop on Data Management for End-to-End Machine Learning},
  year      = {2024}
}

@article{xu2024ondevicelanguagemodelscomprehensive,
  title   = {On-Device Language Models: A Comprehensive Review},
  author  = {Xu, Jiajun and Li, Zhiyuan and Chen, Wei and Wang, Qun and Gao, Xin and Cai, Qi and Ling, Ziyuan},
  journal = {arXiv preprint arXiv:2409.00088},
  year    = {2024}
}

@inproceedings{zhou-etal-2022-textfusion,
  title     = {TextFusion: Privacy-Preserving Pre-Trained Model Inference Via Token Fusion},
  author    = {Zhou, Xin and Lu, Jinzhu and Gui, Tao and Ma, Ruotian and Fei, Zichu and Wang, Yuran and Ding, Yong and Cheung, Yibo and Zhang, Qi and Huang, Xuanjing},
  booktitle = {EMNLP},
  year      = {2022}
}

@inproceedings{plant-etal-2021-cape-noise,
  title     = {Cape: Context-Aware Private Embeddings For Private Language Learning},
  author    = {Plant, Richard and Gkatzia, Dimitra and Giuffrida, Valerio},
  booktitle = {EMNLP},
  year      = {2021}
}

@inproceedings{BlackVIP_2023_CVPR,
  title     = {BlackVIP: Black-Box Visual Prompting For Robust Transfer Learning},
  author    = {Oh, Changdae and Hwang, Hyeji and Lee, Hee-young and Lim, YongTaek and Jung, Geunyoung and Jung, Jiyoung and Choi, Hosik and Song, Kyungwoo},
  booktitle = {CVPR},
  year      = {2023}
}

@article{frikha2024obfuscatune,
  title   = {Obfuscatune: Obfuscated Offsite Fine-Tuning And Inference Of Proprietary LLMs On Private Datasets},
  author  = {Frikha, Ahmed and Walha, Nassim and Mendes, Ricardo and Nakka, Krishna Kanth and Jiang, Xue and Zhou, Xuebing},
  journal = {arXiv preprint arXiv:2407.02960},
  year    = {2024}
}

@inproceedings{lu2023zoopflexploringblackboxfoundation,
  title     = {ZooPFL: Exploring Black-Box Foundation Models For Personalized Federated Learning},
  author    = {Lu, Wang and Yu, Hao and Wang, Jindong and Teney, Damien and Wang, Haohan and Zhu, Yao and Chen, Yiqiang and Yang, Qiang and Xie, Xing and Ji, Xiangyang},
  booktitle = {International Workshop on Trustworthy Federated Learning},
  year      = {2024}
}

@inproceedings{cheng2024black,
  title     = {Black-Box Prompt Optimization: Aligning Large Language Models Without Model Training},
  author    = {Cheng, Jiale and Liu, Xiao and Zheng, Kehan and Ke, Pei and Wang, Hongning and Dong, Yuxiao and Tang, Jie and Huang, Minlie},
  booktitle = {ACL},
  year      = {2024}
}

@inproceedings{chen2024instructzero,
  title     = {INSTRUCTZERO: Efficient Instruction Optimization For Black-Box Large Language Models},
  author    = {Chen, Lichang and Chen, Jiuhai and Goldstein, Tom and Huang, Heng and Zhou, Tianyi},
  booktitle = {ICML},
  year      = {2024}
}

@inproceedings{li-liang-2021-prefix,
  title     = {Prefix-Tuning: Optimizing Continuous Prompts For Generation},
  author    = {Li , Xiang-Lisa and others},
  booktitle = {ACL},
  year      = {2021}
}

@inproceedings{yumultimodal,
  title     = {Multimodal Federated Learning Via Contrastive Representation Ensemble},
  author    = {Yu, Qiying and Liu, Yang and Wang, Yimu and Xu, Ke and Liu, Jingjing},
  booktitle = {ICLR},
  year      = {2023}
}

@inproceedings{fu2024membership,
  title     = {Membership Inference Attacks Against Fine-Tuned Large Language Models Via Self-Prompt Calibration},
  author    = {Fu, Wenjie and Wang, Huandong and Gao, Chen and Liu, Guanghua and Li, Yong and Jiang, Tao},
  booktitle = {NeurIPS},
  year      = {2024}
}

@article{kurakin2023harnessing,
  title   = {Harnessing Large-Language Models To Generate Private Synthetic Text},
  author  = {Kurakin, Alexey and Ponomareva, Natalia and Syed, Umar and MacDermed, Liam and Terzis, Andreas},
  journal = {arXiv preprint arXiv:2306.01684},
  year    = {2023}
}

@inproceedings{yu2024privacy,
  title     = {Privacy-Preserving Instructions For Aligning Large Language Models},
  author    = {Yu, Da and Kairouz, Peter and Oh, Sewoong and Xu, Zheng},
  booktitle = {ICML},
  year      = {2024}
}

@inproceedings{lindifferentially,
  title     = {Differentially Private Synthetic Data Via Foundation Model APIs 1: Images},
  author    = {Lin, Zinan and others},
  booktitle = {ICLR},
  year      = {2024}
}

@inproceedings{dong2022privacy,
  title     = {Privacy For Free: How Does Dataset Condensation Help Privacy?},
  author    = {Dong, Tian and Zhao, Bo and Lyu, Lingjuan},
  booktitle = {ICML},
  year      = {2022}
}

@article{zhang2023trading,
  title   = {Trading Off Privacy, Utility, And Efficiency In Federated Learning},
  author  = {Zhang, Xiaojin and Kang, Yan and Chen, Kai and Fan, Lixin and Yang, Qiang},
  journal = {ACM TIST},
  year    = {2023},
  volume  = {14},
  number  = {6},
  pages   = {1--32}
}

@article{GU2022102907,
  title   = {Privacy, Accuracy, And Model Fairness Trade-Offs In Federated Learning},
  author  = {Xiuting Gu and Zhu Tianqing and Jie Li and Tao Zhang and Wei Ren and Kim-Kwang Raymond Choo},
  journal = {Computers \& Security},
  year    = {2022},
  volume  = {122},
  pages   = {102907}
}

@inproceedings{chen2020breaking,
  title     = {Breaking The Communication-Privacy-Accuracy Trilemma},
  author    = {Chen, Wei-Ning and Kairouz, Peter and Ozgur, Ayfer},
  booktitle = {NeurIPS},
  year      = {2020}
}

@inproceedings{pmlr-v235-tramer24a,
  title     = {Position: Considerations For Differentially Private Learning With Large-Scale Public Pretraining},
  author    = {Tramer, Florian and Kamath, Gautam and Carlini, Nicholas},
  booktitle = {ICML},
  year      = {2024}
}

@inproceedings{faiz2024llmcarbonmodelingendtoendcarbon,
  title     = {LLMCarbon: Modeling The End-To-End Carbon Footprint Of Large Language Models},
  author    = {Faiz, Ahmad and Kaneda, Sotaro and Wang, Ruhan and Osi, Rita and Sharma, Prateek and Chen, Fan and Jiang, Lei},
  booktitle = {ICLR},
  year      = {2024}
}

@article{deb2011multi,
  title   = {A Review Of Multi-Objective Optimization: Methods And Its Applications},
  author  = {Gunantara, Nyoman},
  journal = {Cogent Engineering},
  year    = {2018},
  volume  = {5},
  number  = {1},
  pages   = {1502242}
}

@inproceedings{ding2024hybrid,
  title     = {Hybrid LLM: Cost-Efficient And Quality-Aware Query Routing},
  author    = {Ding, Dujian and Mallick, Ankur and Wang, Chi and Sim, Robert and Mukherjee, Subhabrata and Ruhle, Victor and Lakshmanan, Laks VS and Awadallah, Ahmed Hassan},
  booktitle = {ICLR},
  year      = {2024}
}

@article{wu2022sustainableaienvironmentalimplications,
  title   = {Sustainable AI: Environmental Implications, Challenges And Opportunities},
  author  = {Wu, Carole-Jean and Raghavendra, Ramya and Gupta, Udit and Acun, Bilge and Ardalani, Newsha and Maeng, Kiwan and Chang, Gloria and Aga, Fiona and Huang, Jinshi and Bai, Charles and others},
  journal = {PMLR},
  year    = {2022},
  volume  = {4},
  pages   = {795--813}
}

@article{fedbert,
  title   = {FedBERT: When Federated Learning Meets Pre-Training},
  author  = {Tian, Yuanyishu and Wan, Yao and Lyu, Lingjuan and Yao, Dezhong and Jin, Hai and Sun, Lichao},
  journal = {ACM TIST},
  year    = {2022},
  volume  = {13},
  number  = {4}
}

@inproceedings{zhao2023fedprompt,
  title     = {FedPrompt: Communication-Efficient And Privacy-Preserving Prompt Tuning In Federated Learning},
  author    = {Zhao, Haodong and Du, Wei and Li, Fangqi and Li, Peixuan and Liu, Gongshen},
  booktitle = {ICASSP},
  year      = {2023}
}

@article{jin2024fedmlheefficienthomomorphicencryptionbasedprivacypreserving,
  title   = {FedML-HE: An Efficient Homomorphic-Encryption-Based Privacy-Preserving Federated Learning System},
  author  = {Jin, Weizhao and Yao, Yuhang and Han, Shanshan and Gu, Jiajun and Joe-Wong, Carlee and Ravi, Srivatsan and Avestimehr, Salman and He, Chaoyang},
  journal = {arXiv preprint arXiv:2303.10837},
  year    = {2023}
}

@article{chen2024adaptive,
  title   = {Adaptive Layer Splitting For Wireless LLM Inference In Edge Computing: A Model-Based Reinforcement Learning Approach},
  author  = {Chen, Yuxuan and Li, Rongpeng and Yu, Xiaoxue and Zhao, Zhifeng and Zhang, Honggang},
  journal = {arXiv preprint arXiv:2406.02616},
  year    = {2024}
}

@article{mudvari2024splitllm,
  title   = {Splitllm: Collaborative Inference Of LLMs For Model Placement And Throughput Optimization},
  author  = {Mudvari, Akrit and Jiang, Yuang and Tassiulas, Leandros},
  journal = {arXiv preprint arXiv:2410.10759},
  year    = {2024}
}

@article{cao2024sfprompt,
  title   = {SFPrompt: Communication-Efficient Split Federated Fine-Tuning For Large Pre-Trained Models Over Resource-Limited Devices},
  author  = {Cao, Linxiao and Zhu, Yifei and Gong, Wei},
  journal = {arXiv preprint arXiv:2407.17533},
  year    = {2024}
}

@article{wang2023privatelora,
  title   = {Privatelora For Efficient Privacy Preserving LLM},
  author  = {Wang, Yiming and Lin, Yu and Zeng, Xiaodong and Zhang, Guannan},
  journal = {arXiv preprint arXiv:2311.14030},
  year    = {2023}
}

@inproceedings{chen-etal-2022-x,
  title     = {THE-X: Privacy-Preserving Transformer Inference With Homomorphic Encryption},
  author    = {Chen, Tianyu and Bao, Hangbo and Huang, Shaohan and Dong, Li and Jiao, Binxing and Jiang, Daxin and Zhou, Haoyi and Li, Jianxin and Wei, Furu},
  booktitle = {ACL Findings},
  year      = {2022}
}

@article{dong2023pumasecureinferencellama7b,
  title   = {Puma: Secure Inference Of Llama-7b In Five Minutes},
  author  = {Dong, Ye and Lu, Wen-jie and Zheng, Yancheng and Wu, Haoqi and Zhao, Derun and Tan, Jin and Huang, Zhicong and Hong, Cheng and Wei, Tao and Chen, Wenguang},
  journal = {Security and Safety},
  year    = {2025},
  volume  = {4},
  pages   = {2025014}
}

@article{oliynyk2023know,
  title   = {I Know What You Trained Last Summer: A Survey On Stealing Machine Learning Models And Defences},
  author  = {Oliynyk, Daryna and Mayer, Rudolf and Rauber, Andreas},
  journal = {ACM Computing Surveys},
  year    = {2023},
  volume  = {55},
  number  = {14s},
  pages   = {1--41}
}

@inproceedings{villalobos2024,
  title     = {Position: Will We Run Out Of Data? Limits Of LLM Scaling Based On Human-Generated Data},
  author    = {Villalobos, Pablo and Ho, Anson and Sevilla, Jaime and Besiroglu, Tamay and Heim, Lennart and Hobbhahn, Marius},
  booktitle = {ICML},
  year      = {2024}
}

@article{yang2024survey,
  title   = {Survey On Knowledge Distillation For Large Language Models: Methods, Evaluation, And Application},
  author  = {Yang, Chuanpeng and Zhu, Yao and Lu, Wang and Wang, Yidong and Chen, Qian and Gao, Chenlong and Yan, Bingjie and Chen, Yiqiang},
  journal = {ACM TIST},
  year    = {2024}
}

@article{yang2019federated,
  title   = {Federated Machine Learning: Concept And Applications},
  author  = {Yang, Qiang and Liu, Yang and Chen, Tianjian and Tong, Yongxin},
  journal = {ACM TIST},
  year    = {2019},
  volume  = {10},
  number  = {2},
  pages   = {1--19}
}

@article{li2025edge,
  title   = {Collaborative Inference And Learning Between Edge SLMs And Cloud LLMs: A Survey Of Algorithms, Execution, And Open Challenges},
  author  = {Li, Senyao and Wang, Haozhao and Xu, Wenchao and Zhang, Rui and Guo, Song and Yuan, Jingling and Zhong, Xian and Zhang, Tianwei and Li, Ruixuan},
  journal = {arXiv preprint arXiv:2507.16731},
  year    = {2025}
}

@article{edemacu2025privacy,
  title   = {Privacy Preserving Prompt Engineering: A Survey},
  author  = {Edemacu, Kennedy and Wu, Xintao},
  journal = {ACM Computing Surveys},
  year    = {2025},
  volume  = {57},
  number  = {10},
  pages   = {1--36}
}

@article{pilan2022text,
  title   = {The Text Anonymization Benchmark (Tab): A Dedicated Corpus And Evaluation Framework For Text Anonymization},
  author  = {Pil{\'a}n, Ildik{\'o} and Lison, Pierre and {\O}vrelid, Lilja and Papadopoulou, Anthi and S{\'a}nchez, David and Batet, Montserrat},
  journal = {Computational Linguistics},
  year    = {2022},
  volume  = {48},
  number  = {4},
  pages   = {1053--1101}
}

@article{iourovitski2024hide,
  title   = {Hide And Seek: Fingerprinting Large Language Models With Evolutionary Learning},
  author  = {Iourovitski, Dmitri and Sharma, Sanat and Talwar, Rakshak},
  journal = {arXiv preprint arXiv:2408.02871},
  year    = {2024}
}

@inproceedings{xu2024instructionalfingerprintinglargelanguage,
  title     = {Instructional Fingerprinting Of Large Language Models},
  author    = {Xu, Jiashu and Wang, Fei and Ma, Mingyu and Koh, Pang Wei and Xiao, Chaowei and Chen, Muhao},
  booktitle = {NAACL},
  year      = {2024}
}

@inproceedings{xhonneux2024efficient,
  title     = {Efficient Adversarial Training In LLMs With Continuous Attacks},
  author    = {Xhonneux, Sophie and Sordoni, Alessandro and G{\"u}nnemann, Stephan and Gidel, Gauthier and Schwinn, Leo},
  booktitle = {NeurIPS},
  year      = {2024}
}

@inproceedings{right_to_be_forget_Shintre2019MakingML,
  title     = {Making Machine Learning Forget},
  author    = {Saurabh Shintre and Kevin A. Roundy and Jasjeet Dhaliwal},
  booktitle = {Annual Privacy Forum},
  year      = {2019}
}

@inproceedings{jang-etal-2023-knowledge,
  title     = {Knowledge Unlearning For Mitigating Privacy Risks In Language Models},
  author    = {Jang, Joel and Yoon, Dongkeun and Yang, Sohee and Cha, Sungmin and Lee, Moontae and Logeswaran, Lajanugen and Seo, Minjoon},
  booktitle = {ACL},
  year      = {2023}
}

@inproceedings{yao2024machineunlearning,
  title     = {Machine Unlearning Of Pre-Trained Large Language Models},
  author    = {Yao, Jin and Chien, Eli and Du, Minxin and Niu, Xinyao and Wang, Tianhao and Cheng, Zezhou and Yue, Xiang},
  booktitle = {ACL},
  year      = {2024}
}

@inproceedings{liu2024ddk,
  title     = {DDK: Distilling Domain Knowledge For Efficient Large Language Models},
  author    = {Liu, Jiaheng and Zhang, Chenchen and Guo, Jinyang and Zhang, Yuanxing and Que, Haoran and Deng, Ken and Bai, Zhiqi and Liu, Jie and Zhang, Ge and Wang, Jiakai and others},
  booktitle = {NeurIPS},
  year      = {2024}
}

@inproceedings{wei2022chain,
  title     = {Chain Of Thought Prompting Elicits Reasoning In Large Language Models},
  author    = {Jason Wei and Xuezhi Wang and Dale Schuurmans and Maarten Bosma and brian ichter and Fei Xia and Ed H. Chi and Quoc V Le and Denny Zhou},
  booktitle = {NeurIPS},
  year      = {2022}
}

@inproceedings{brown2020language,
  title     = {Language Models Are Few-Shot Learners},
  author    = {Brown, Tom and Mann, Benjamin and Ryder, Nick and Subbiah, Melanie and Kaplan, Jared D and Dhariwal, Prafulla and Neelakantan, Arvind and Shyam, Pranav and Sastry, Girish and Askell, Amanda and Agarwal, Sandhini and Herbert-Voss, Ariel and Krueger, Gretchen and Henighan, Tom and Child, Rewon and Ramesh, Aditya and Ziegler, Daniel and Wu, Jeffrey and Winter, Clemens and Hesse, Chris and Chen, Mark and Sigler, Eric and Litwin, Mateusz and Gray, Scott and Chess, Benjamin and Clark, Jack and Berner, Christopher and McCandlish, Sam and Radford, Alec and Sutskever, Ilya and Amodei, Dario},
  booktitle = {NeurIPS},
  year      = {2020}
}

@inproceedings{ruida2023let,
  title     = {Let's Synthesize Step By Step: Iterative Dataset Synthesis With Large Language Models By Extrapolating Errors From Small Models},
  author    = {Ruida, WANG and Zhou, Wangchunshu and Sachan, Mrinmaya},
  booktitle = {EMNLP},
  year      = {2023}
}

@article{chen2023tutorial,
  title   = {Tutorial: Toward Robust Deep Learning Against Poisoning Attacks},
  author  = {Chen, Huili and Koushanfar, Farinaz},
  journal = {ACM TESC},
  year    = {2023},
  volume  = {22},
  number  = {3},
  pages   = {1--15}
}

@inproceedings{hou2024ibd,
  title     = {IBD-PSC: Input-Level Backdoor Detection Via Parameter-Oriented Scaling Consistency},
  author    = {Hou, Linshan and Feng, Ruili and Hua, Zhongyun and Luo, Wei and Zhang, Leo Yu and Li, Yiming},
  booktitle = {ICML},
  year      = {2024}
}

@inproceedings{dong2021black,
  title     = {Black-Box Detection Of Backdoor Attacks With Limited Information And Data},
  author    = {Dong, Yinpeng and Yang, Xiao and Deng, Zhijie and Pang, Tianyu and Xiao, Zihao and Su, Hang and Zhu, Jun},
  booktitle = {ICCV},
  year      = {2021}
}

@inproceedings{wang2022improved,
  title     = {Improved Certified Defenses Against Data Poisoning With (Deterministic) Finite Aggregation},
  author    = {Wang, Wenxiao and Levine, Alexander J and Feizi, Soheil},
  booktitle = {ICML},
  year      = {2022}
}

@inproceedings{banerjee2025safeinfer,
  title     = {Safeinfer: Context Adaptive Decoding Time Safety Alignment For Large Language Models},
  author    = {Banerjee, Somnath and Layek, Sayan and Tripathy, Soham and Kumar, Shanu and Mukherjee, Animesh and Hazra, Rima},
  booktitle = {AAAI},
  year      = {2025}
}

@article{goldblum2022dataset,
  title   = {Dataset Security For Machine Learning: Data Poisoning, Backdoor Attacks, And Defenses},
  author  = {Goldblum, Micah and Tsipras, Dimitris and Xie, Chulin and Chen, Xinyun and Schwarzschild, Avi and Song, Dawn and M{\k{a}}dry, Aleksander and Li, Bo and Goldstein, Tom},
  journal = {TPAMI},
  year    = {2022}
}

@inproceedings{yang2022not,
  title     = {Not All Poisons Are Created Equal: Robust Training Against Data Poisoning},
  author    = {Yang, Yu and Liu, Tian Yu and Mirzasoleiman, Baharan},
  booktitle = {ICML},
  year      = {2022}
}

@inproceedings{xie2024differentially,
  title     = {Differentially Private Synthetic Data Via Foundation Model APIs 2: Text},
  author    = {Xie, Chulin and Lin, Zinan and Backurs, Arturs and Gopi, Sivakanth and Yu, Da and Inan, Huseyin A and Nori, Harsha and Jiang, Haotian and Zhang, Huishuai and Lee, Yin Tat and others},
  booktitle = {ICML},
  year      = {2024}
}

@article{sheikhi2025hybrid,
  title   = {Hybrid Reputation Aggregation: A Robust Defense Mechanism For Adversarial Federated Learning In 5G And Edge Network Environments},
  author  = {Sheikhi, Saeid and Kostakos, Panos and Loven, Lauri},
  journal = {IEEE Open Journal of the Communications Society},
  year    = {2025},
  volume  = {7},
  pages   = {370--385}
}

@article{khedekar2025sybil,
  title   = {Sybil-Aware Adaptive Defence Framework For Robust Federated Learning},
  author  = {Khedekar, Dnyanesh and Mahapatra, Tanmaya and Rajput, Amitesh Singh},
  journal = {Pervasive and Mobile Computing},
  year    = {2025},
  pages   = {102157}
}

@inproceedings{wan2022shielding,
  title     = {Shielding Federated Learning: Robust Aggregation With Adaptive Client Selection},
  author    = {Wan, Wei and Hu, Shengshan and Lu, Jianrong and Zhang, Leo Yu and Jin, Hai and He, Yuanyuan},
  booktitle = {IJCAI},
  year      = {2022}
}

@inproceedings{wang2024taylor,
  title     = {Taylor Unswift: Secured Weight Release For Large Language Models Via Taylor Expansion},
  author    = {Wang, Guanchu and Chuang, Yu-Neng and Tang, Ruixiang and Zhong, Shaochen and Yuan, Jiayi and Jin, Hongye and Liu, Zirui and Chaudhary, Vipin and Xu, Shuai and Caverlee, James and others},
  booktitle = {EMNLP},
  year      = {2024}
}

@article{liu2021learning,
  title   = {Learning To Teach With Student Feedback},
  author  = {Liu, Yi-tao and Sun, Tian-xiang and Qiu, Xi-peng and Huang, Xuan-jing},
  journal = {arXiv preprint arXiv:2109.04641},
  year    = {2021}
}

@inproceedings{pham2021meta,
  title     = {Meta Pseudo Labels},
  author    = {Pham, Hieu and Dai, Zihang and Xie, Qizhe and Le, Quoc V},
  booktitle = {CVPR},
  year      = {2021}
}

@inproceedings{li2022shadow,
  title     = {Shadow Knowledge Distillation: Bridging Offline And Online Knowledge Transfer},
  author    = {Li, Lujun and Jin, Zhe},
  booktitle = {NeurIPS},
  year      = {2022}
}

@inproceedings{nasser2024reverse,
  title     = {Reverse Knowledge Distillation: Training A Large Model Using A Small One For Retinal Image Matching On Limited Data},
  author    = {Nasser, Sahar Almahfouz and Gupte, Nihar and Sethi, Amit},
  booktitle = {WACV},
  year      = {2024}
}

@article{cheng2021fedgems,
  title   = {FedGEMS: Federated Learning Of Larger Server Models Via Selective Knowledge Fusion},
  author  = {Cheng, Sijie and Wu, Jingwen and Xiao, Yanghua and Liu, Yang},
  journal = {arXiv preprint arXiv:2110.11027},
  year    = {2021}
}

@inproceedings{jin2023parameter,
  title     = {Parameter-Efficient Tuning For Large Language Model Without Calculating Its Gradients},
  author    = {Feihu Jin and Jiajun Zhang and Chengqing Zong},
  booktitle = {EMNLP},
  year      = {2023}
}

@article{wu2025bidirectional,
  title   = {Bidirectional Knowledge Distillation For Enhancing Sequential Recommendation With Large Language Models},
  author  = {Wu, Jiongran and Liu, Jiahao and Li, Dongsheng and Zhang, Guangping and Han, Mingzhe and Gu, Hansu and Zhang, Peng and Shang, Li and Lu, Tun and Gu, Ning},
  journal = {arXiv preprint arXiv:2505.18120},
  year    = {2025}
}

@inproceedings{li2025bild,
  title     = {Bild: Bi-Directional Logits Difference Loss For Large Language Model Distillation},
  author    = {Li, Minchong and Zhou, Feng and Song, Xiaohui},
  booktitle = {COLING},
  year      = {2025}
}

@inproceedings{qin2025beyond,
  title     = {Beyond Output Matching: Bidirectional Alignment For Enhanced In-Context Learning},
  author    = {Qin, Chengwei and Xia, Wenhan and Jiao, Fangkai and Chen, Chen and Hu, Yuchen and Ding, Bosheng and Chen, Ruirui and Joty, Shafiq},
  booktitle = {ACL},
  year      = {2025}
}

@inproceedings{zhang2021dual,
  title     = {Dual Knowledge Distillation For Bidirectional Neural Machine Translation},
  author    = {Zhang, Huaao and Qiu, Shigui and Wu, Shilong},
  booktitle = {IJCNN},
  year      = {2021}
}

@article{fan2023fatellm,
  title   = {Fate-LLM: A Industrial Grade Federated Learning Framework For Large Language Models},
  author  = {Fan, Tao and Kang, Yan and Ma, Guoqiang and Chen, Weijing and Wei, Wenbin and Fan, Lixin and Yang, Qiang},
  journal = {arXiv preprint arXiv:2310.10049},
  year    = {2023}
}

@inproceedings{kuang2023federatedscopellm,
  title     = {Federatedscope-LLM: A Comprehensive Package For Fine-Tuning Large Language Models In Federated Learning},
  author    = {Kuang, Weirui and Qian, Bingchen and Li, Zitao and Chen, Daoyuan and Gao, Dawei and Pan, Xuchen and Xie, Yuexiang and Li, Yaliang and Ding, Bolin and Zhou, Jingren},
  booktitle = {KDD},
  year      = {2024}
}

@inproceedings{ye2024openfedllm,
  title     = {Openfedllm: Training Large Language Models On Decentralized Private Data Via Federated Learning},
  author    = {Ye, Rui and Wang, Wenhao and Chai, Jingyi and Li, Dihan and Li, Zexi and Xu, Yinda and Du, Yaxin and Wang, Yanfeng and Chen, Siheng},
  booktitle = {KDD},
  year      = {2024}
}

@inproceedings{fan2025fedmkt,
  title     = {FedMKT: Federated Mutual Knowledge Transfer For Large And Small Language Models},
  author    = {Fan, Tao and Ma, Guoqiang and Kang, Yan and Gu, Hanlin and Song, Yuanfeng and Fan, Lixin and Chen, Kai and Yang, Qiang},
  booktitle = {COLING},
  year      = {2025}
}

@article{chua2024fedpeat,
  title   = {FedPEAT: Convergence Of Federated Learning, Parameter-Efficient Fine Tuning, And Emulator Assisted Tuning For Artificial Intelligence Foundation Models With Mobile Edge Computing},
  author  = {Chua, Terence Jie and Yu, Wenhan and Zhao, Jun and Lam, Kwok-Yan},
  journal = {arXiv preprint arXiv:2310.17491},
  year    = {2023}
}

@article{deng2023mutual,
  title   = {Mutual Enhancement Of Large And Small Language Models With Cross-Silo Knowledge Transfer},
  author  = {Deng, Yongheng and Qiao, Ziqing and Ren, Ju and Liu, Yang and Zhang, Yaoxue},
  journal = {arXiv preprint arXiv:2312.05842},
  year    = {2023}
}

@inproceedings{yu2023orchestration,
  title     = {Orchestration Of Emulator Assisted 6G Mobile Edge Tuning For AI Foundation Models: A Multi-Agent Deep Reinforcement Learning Approach},
  author    = {Yu, Wenhan and Chua, Terence Jie and Zhao, Jun},
  booktitle = {IEEE VTC},
  year      = {2024}
}

@inproceedings{wumixture,
  title     = {Mixture Of LoRA Experts},
  author    = {Wu, Xun and Huang, Shaohan and Wei, Furu},
  booktitle = {ICLR},
  year      = {2024}
}

@inproceedings{liu2024dora,
  title     = {DoRA: Weight-Decomposed Low-Rank Adaptation},
  author    = {Liu, Shih-Yang and Wang, Chien-Yi and Yin, Hongxu and Molchanov, Pavlo and Wang, Yu-Chiang Frank and Cheng, Kwang-Ting and Chen, Min-Hung},
  booktitle = {ICML},
  year      = {2024}
}

@inproceedings{zhang2024upload,
  title     = {An Upload-Efficient Scheme For Transferring Knowledge From A Server-Side Pre-Trained Generator To Clients In Heterogeneous Federated Learning},
  author    = {Zhang, Jianqing and Liu, Yang and Hua, Yang and Cao, Jian},
  booktitle = {CVPR},
  year      = {2024}
}

@inproceedings{zhang2024federated,
  title     = {Federated Adaptation For Foundation Model-Based Recommendations},
  author    = {Zhang, Chun-xu and Long, Guo-dong and Guo, Hong-kuan and Fang, Xiao and Song, Yang and Liu, Zhao-jie and Zhou, Guo-rui and Zhang, Zi-jian and Liu, Yang and Yang, Bo},
  booktitle = {IJCAI},
  year      = {2024}
}

@article{shen2023split,
  title   = {A Split-And-Privatize Framework For Large Language Model Fine-Tuning},
  author  = {Shen, Xicong and Liu, Yang and Liu, Huiqi and Hong, Jue and Duan, Bing and Huang, Zirui and Mao, Yunlong and Wu, Ye and Wu, Di},
  journal = {arXiv preprint arXiv:2312.15603},
  year    = {2023}
}

@inproceedings{thareja2026dp,
  title     = {DP-Fusion: Token-Level Differentially Private Inference For Large Language Models},
  author    = {Thareja, Rushil and Nakov, Preslav and Vepakomma, Praneeth and Lukas, Nils},
  booktitle = {ICLR},
  year      = {2025}
}

@inproceedings{liu2024tuning,
  title     = {Tuning Language Models By Proxy},
  author    = {Liu, A-lisa and Han, Xiao-chuang and Wang, Yi-zhong and Tsvetkov, Yu-lia and Choi, Ye-jin and Smith, Noah A},
  booktitle = {COLM},
  year      = {2024}
}

@inproceedings{ormazabal2023comblm,
  title     = {CombLM: Adapting Black-Box Language Models Through Small Fine-Tuned Models},
  author    = {Ormazabal, Aitor and Artetxe, Mikel and Agirre, Eneko},
  booktitle = {EMNLP},
  year      = {2023}
}

@inproceedings{mitchell2024emulator,
  title     = {An Emulator For Fine-Tuning Large Language Models Using Small Language Models},
  author    = {Mitchell, Eric and Rafailov, Rafael and Sharma, Archit and Finn, Chelsea and Manning, Christopher D},
  booktitle = {ICLR},
  year      = {2024}
}

@article{huang2024offset,
  title   = {Offset Unlearning For Large Language Models},
  author  = {Huang, James Y and Zhou, Wenxuan and Wang, Fei and Morstatter, Fred and Zhang, Sheng and Poon, Hoifung and Chen, Muhao},
  journal = {TMLR},
  year    = {2025}
}

@inproceedings{fan2024giant,
  title     = {On Giant's Shoulders: Effortless Weak To Strong By Dynamic Logits Fusion},
  author    = {Fan, Chenghao and Lu, Zhenyi and Wei, Wei and Tian, Jie and Qu, Xiaoye and Chen, Dangyang and Cheng, Yu},
  booktitle = {NeurIPS},
  year      = {2024}
}

@article{he2024cpt,
  title   = {CPT: Consistent Proxy Tuning For Black-Box Optimization},
  author  = {He, Yuanyang and Huang, Zitong and Xu, Xinxing and Goh, Rick Siow Mong and Khan, Salman and Zuo, Wangmeng and Liu, Yong and Feng, Chun-Mei},
  journal = {arXiv preprint arXiv:2407.01155},
  year    = {2024}
}

@inproceedings{li2023contrastive,
  title     = {Contrastive Decoding: Open-Ended Text Generation As Optimization},
  author    = {Li, Xiang Lisa and Holtzman, Ari and Fried, Daniel and Liang, Percy and Eisner, Jason and Hashimoto, Tatsunori and Zettlemoyer, Luke and Lewis, Mike},
  booktitle = {ACL},
  year      = {2023}
}

@article{zhang2024fast,
  title   = {Fast And Slow Generating: An Empirical Study On Large And Small Language Models Collaborative Decoding},
  author  = {Zhang, Kaiyan and Wang, Jianyu and Ding, Ning and Qi, Biqing and Hua, Ermo and Lv, Xingtai and Zhou, Bowen},
  journal = {arXiv preprint arXiv:2406.12295},
  year    = {2024}
}

@article{yao2024privacy,
  title   = {Privacy-Preserving Language Model Inference With Instance Obfuscation},
  author  = {Yao, Yixiang and Wang, Fei and Ravi, Srivatsan and Chen, Muhao},
  journal = {arXiv preprint arXiv:2402.08227},
  year    = {2024}
}

@article{mehta2024openelm,
  title   = {Openelm: An Efficient Language Model Family With Open Training And Inference Framework},
  author  = {Mehta, Sachin and Sekhavat, Mohammad Hossein and Cao, Qingqing and Horton, Maxwell and Jin, Yanzi and Sun, Chenfan and Mirzadeh, Iman and Najibi, Mahyar and Belenko, Dmitry and Zatloukal, Peter and others},
  journal = {arXiv preprint arXiv:2404.14619},
  year    = {2024}
}

@article{grattafiori2024llama,
  title   = {The Llama 3 Herd Of Models},
  author  = {Grattafiori, Aaron and Dubey, Abhimanyu and Jauhri, Abhinav and Pandey, Abhinav and Kadian, Abhishek and Al-Dahle, Ahmad and Letman, Aiesha and Mathur, Akhil and Schelten, Alan and Vaughan, Alex and others},
  journal = {arXiv preprint arXiv:2407.21783},
  year    = {2024}
}

@inproceedings{shen2024learning,
  title     = {Learning To Decode Collaboratively With Multiple Language Models},
  author    = {Shen, Zejiang and Lang, Hunter and Wang, Bailin and Kim, Yoon and Sontag, David},
  booktitle = {ACL},
  year      = {2024}
}

@article{10882951,
  title   = {CoLLM: Integrating Collaborative Embeddings Into Large Language Models For Recommendation},
  author  = {Zhang, Yang and Feng, Fuli and Zhang, Jizhi and Bao, Keqin and Wang, Qifan and He, Xiangnan},
  journal = {TKDE},
  year    = {2025},
  volume  = {37},
  number  = {5},
  pages   = {2329-2340}
}

@inproceedings{kang2017neurosurgeon,
  title     = {Neurosurgeon: Collaborative Intelligence Between The Cloud And Mobile Edge},
  author    = {Kang, Yiping and Hauswald, Johann and Gao, Cao and Rovinski, Austin and Mudge, Trevor and Mars, Jason and Tang, Lingjia},
  booktitle = {ACM ASPLOS},
  year      = {2017}
}

@inproceedings{li2022federated,
  title     = {Federated Learning With Gan-Based Data Synthesis For Non-IID Clients},
  author    = {Li, Zi-jian and Shao, Jia-wei and Mao, Yu-yi and Wang, Jessie Hui and Zhang, Jun},
  booktitle = {International Workshop on Trustworthy Federated Learning},
  year      = {2022}
}

@inproceedings{cheng2023gfl,
  title     = {GFL: Federated Learning On Non-IID Data Via Privacy-Preserving Synthetic Data},
  author    = {Cheng, Yihang and Zhang, Lan and Li, Anran},
  booktitle = {PerCom},
  year      = {2023}
}

@article{gunasekar2023textbooks,
  title   = {Textbooks Are All You Need},
  author  = {Gunasekar, Suriya and Zhang, Yi and Aneja, Jyoti and Mendes, Caio C{\'e}sar Teodoro and Del Giorno, Allie and Gopi, Sivakanth and Javaheripi, Mojan and Kauffmann, Piero and de Rosa, Gustavo and Saarikivi, Olli and others},
  journal = {arXiv preprint arXiv:2306.11644},
  year    = {2023}
}

@article{zhou2020distilled,
  title   = {Distilled One-Shot Federated Learning},
  author  = {Zhou, Yanlin and Pu, George and Ma, Xiyao and Li, Xiaolin and Wu, Dapeng},
  journal = {arXiv preprint arXiv:2009.07999},
  year    = {2020}
}

@article{wang2018dataset,
  title   = {Dataset Distillation},
  author  = {Wang, Tongzhou and Zhu, Jun-Yan and Torralba, Antonio and Efros, Alexei A},
  journal = {arXiv preprint arXiv:1811.10959},
  year    = {2018}
}

@inproceedings{fancore,
  title     = {CORE: Reducing UI Exposure In Mobile Agents Via Collaboration Between Cloud And Local LLMs},
  author    = {Fan, Gucongcong and Niu, Chaoyue and Wu, Fan and Chen, Guihai and others},
  booktitle = {NeurIPS},
  year      = {2025}
}

@inproceedings{yi2025ecoagent,
  title     = {EcoAgent: An Efficient Edge-Cloud Collaborative Multi-Agent Framework For Mobile Automation},
  author    = {Yi, Biao and Hu, Xavier and Chen, Yurun and Zhang, Shengyu and Yang, Hongxia and Wu, Fan and Wu, Fei},
  booktitle = {AAAI},
  year      = {2026}
}

@inproceedings{guo2025accrag,
  title     = {Enhancing RAG Efficiency With Adaptive Context Compression},
  author    = {Guo, Shuyu and Zhang, Shuo and Ren, Zhaochun},
  booktitle = {EMNLP Findings},
  year      = {2025}
}

@inproceedings{zhao2021dataset,
  title     = {Dataset Condensation With Gradient Matching},
  author    = {Zhao, Bo and others},
  booktitle = {ICLR},
  year      = {2021}
}

@inproceedings{chen2025fedc4,
  title     = {Rethinking Client-Oriented Federated Graph Learning},
  author    = {Chen, Ze-kai and Li, Xun-kai and Zhu, Yin-lin and Li, Rong-Hua and Wang, Guoren},
  booktitle = {CIKM},
  year      = {2025}
}

@inproceedings{zhang2018mixup,
  title     = {Mixup: Beyond Empirical Risk Minimization},
  author    = {Hongyi Zhang and Moustapha Cisse and Yann N. Dauphin and David Lopez-Paz},
  booktitle = {ICLR},
  year      = {2018}
}

@inproceedings{yan2025federated,
  title     = {Federated Graph Condensation With Information Bottleneck Principles},
  author    = {Yan, Bo and He, Sihao and Yang, Cheng and Liu, Shang and Cao, Yang and Shi, Chuan},
  booktitle = {AAAI},
  year      = {2025}
}

@article{kim2022stable,
  title   = {Stable Federated Learning With Dataset Condensation.},
  author  = {Kim, Seong-Woong and others},
  journal = {JCSE},
  year    = {2022},
  volume  = {16},
  number  = {1},
  pages   = {52--62}
}

@inproceedings{song2023federated,
  title     = {Federated Learning Via Decentralized Dataset Distillation In Resource-Constrained Edge Environments},
  author    = {Song, Rui and Liu, Dai and Chen, Dave Zhenyu and Festag, Andreas and Trinitis, Carsten and Schulz, Martin and Knoll, Alois},
  booktitle = {IJCNN},
  year      = {2023}
}

@article{li2024introducing,
  title   = {Introducing Edge Intelligence To Smart Meters Via Federated Split Learning},
  author  = {Li, Yehui and Qin, Dalin and Poor, H Vincent and Wang, Yi},
  journal = {Nature Communications},
  year    = {2024},
  volume  = {15},
  number  = {1},
  pages   = {9044}
}

@article{kan2023protecting,
  title   = {Protecting User Privacy In Remote Conversational Systems: A Privacy-Preserving Framework Based On Text Sanitization},
  author  = {Kan, Zhigang and Qiao, Linbo and Yu, Hao and Peng, Liwen and Gao, Yifu and Li, Dongsheng},
  journal = {arXiv preprint arXiv:2306.08223},
  year    = {2023}
}

@article{chen2023hide,
  title   = {Hide And Seek (Has): A Lightweight Framework For Prompt Privacy Protection},
  author  = {Chen, Yu and Li, Tingxin and Liu, Huiming and Yu, Yang},
  journal = {arXiv preprint arXiv:2309.03057},
  year    = {2023}
}

@inproceedings{chong2024casper,
  title     = {Casper: Prompt Sanitization For Protecting User Privacy In Web-Based Large Language Models},
  author    = {Chong, Chun Jie and Hou, Chenxi and Yao, Zhihao and Talebi, Seyed Mohammadjavad Seyed},
  booktitle = {CSCloud},
  year      = {2025}
}

@inproceedings{zhang2023latticegen,
  title     = {LatticeGen: Hiding Generated Text In A Lattice For Privacy-Aware Large Language Model Generation On Cloud},
  author    = {Zhang, Mengke and He, Tianxing and Wang, Tianle and Mi, Lu and Mireshghallah, Fatemehsadat and Chen, Binyi and Wang, Hao and Tsvetkov, Yulia},
  booktitle = {NAACL Findings},
  year      = {2024}
}

@article{hochreiter1997long,
  title   = {Long Short-Term Memory},
  author  = {Hochreiter, S},
  journal = {Neural Computation},
  year    = {1997}
}

@article{chung2014empirical,
  title   = {Empirical Evaluation Of Gated Recurrent Neural Networks On Sequence Modeling},
  author  = {Chung, Junyoung and Gulcehre, Caglar and Cho, KyungHyun and Bengio, Yoshua},
  journal = {arXiv preprint arXiv:1412.3555},
  year    = {2014}
}

@inproceedings{he2016deep,
  title     = {Deep Residual Learning For Image Recognition},
  author    = {He, Kaiming and Zhang, Xiangyu and Ren, Shaoqing and Sun, Jian},
  booktitle = {CVPR},
  year      = {2016}
}

@inproceedings{rombach2022high,
  title     = {High-Resolution Image Synthesis With Latent Diffusion Models},
  author    = {Rombach, Robin and Blattmann, Andreas and Lorenz, Dominik and Esser, Patrick and Ommer, Bj{\"o}rn},
  booktitle = {CVPR},
  year      = {2022}
}

@inproceedings{ramesh2021zero,
  title     = {Zero-Shot Text-To-Image Generation},
  author    = {Ramesh, Aditya and Pavlov, Mikhail and Goh, Gabriel and Gray, Scott and Voss, Chelsea and Radford, Alec and Chen, Mark and Sutskever, Ilya},
  booktitle = {ICML},
  year      = {2021}
}

@article{ramesh2022hierarchical,
  title   = {Hierarchical Text-Conditional Image Generation With Clip Latents},
  author  = {Ramesh, Aditya and Dhariwal, Prafulla and Nichol, Alex and Chu, Casey and Chen, Mark},
  journal = {arXiv preprint arXiv:2204.06125},
  year    = {2022},
  volume  = {1},
  number  = {2},
  pages   = {3}
}

@article{betker2023improving,
  title   = {Improving Image Generation With Better Captions},
  author  = {Betker, James and Goh, Gabriel and Jing, Li and Brooks, Tim and Wang, Jianfeng and Li, Linjie and Ouyang, Long and Zhuang, Juntang and Lee, Joyce and Guo, Yufei and others},
  journal = {Computer Science},
  year    = {2023},
  volume  = {2},
  number  = {3},
  pages   = {8}
}

@inproceedings{nichol2022glide,
  title     = {GLIDE: Towards Photorealistic Image Generation And Editing With Text-Guided Diffusion Models},
  author    = {Nichol, Alexander Quinn and Dhariwal, Prafulla and Ramesh, Aditya and Shyam, Pranav and Mishkin, Pamela and Mcgrew, Bob and Sutskever, Ilya and Chen, Mark},
  booktitle = {ICML},
  year      = {2022}
}

@inproceedings{redmon2016you,
  title     = {You Only Look Once: Unified, Real-Time Object Detection},
  author    = {Redmon, J},
  booktitle = {CVPR},
  year      = {2016}
}

@inproceedings{redmon2017yolo9000,
  title     = {Yolo9000: Better, Faster, Stronger},
  author    = {Redmon, Joseph and Farhadi, Ali},
  booktitle = {CVPR},
  year      = {2017}
}

@article{redmon2018yolov3,
  title   = {YOLOv3: An Incremental Improvement},
  author  = {Redmon, Joseph},
  journal = {arXiv preprint arXiv:1804.02767},
  year    = {2018}
}

@article{bochkovskiy2020yolov4,
  title   = {YOLOv4: Optimal Speed And Accuracy Of Object Detection},
  author  = {Bochkovskiy, Alexey and Wang, Chien-Yao and Liao, Hong-Yuan Mark},
  journal = {arXiv preprint arXiv:2004.10934},
  year    = {2020}
}

@article{li2022yolov6,
  title   = {YOLOv6: A Single-Stage Object Detection Framework For Industrial Applications},
  author  = {Li, Chuyi and Li, Lulu and Jiang, Hongliang and Weng, Kaiheng and Geng, Yifei and Li, Liang and Ke, Zaidan and Li, Qingyuan and Cheng, Meng and Nie, Weiqiang and others},
  journal = {arXiv preprint arXiv:2209.02976},
  year    = {2022}
}

@inproceedings{wang2023yolov7,
  title     = {YOLOv7: Trainable Bag-Of-Freebies Sets New State-Of-The-Art For Real-Time Object Detectors},
  author    = {Wang, Chien-Yao and Bochkovskiy, Alexey and Liao, Hong-Yuan Mark},
  booktitle = {CVPR},
  year      = {2023}
}

@article{regulation2016regulation,
  title   = {Regulation (EU) 2016/679 Of The European Parliament And Of The Council},
  author  = {Regulation, Protection},
  journal = {Regulation (eu)},
  year    = {2016},
  volume  = {679},
  pages   = {2016}
}

@article{bonta2022california,
  title   = {California Consumer Privacy Act (CCPA)},
  author  = {Bonta, Rob},
  journal = {State of California Department of Justice},
  year    = {2022}
}

@article{act1996health,
  title   = {Health Insurance Portability And Accountability Act Of 1996},
  author  = {Act, Accountability},
  journal = {Public law},
  year    = {1996},
  volume  = {104},
  pages   = {191}
}

@article{joshi2022federated,
  title   = {Federated Learning For Healthcare Domain-Pipeline, Applications And Challenges},
  author  = {Joshi, Madhura and Pal, Ankit and Sankarasubbu, Malaikannan},
  journal = {ACM Transactions on Computing for Healthcare},
  year    = {2022},
  volume  = {3},
  number  = {4},
  pages   = {1--36}
}

@article{li2023review,
  title   = {Review On Security Of Federated Learning And Its Application In Healthcare},
  author  = {Li, Hao and Li, Chengcheng and Wang, Jian and Yang, Aimin and Ma, Zezhong and Zhang, Zunqian and Hua, Dianbo},
  journal = {FGCS},
  year    = {2023},
  volume  = {144},
  pages   = {271--290}
}

@article{courtiol2019deep,
  title   = {Deep Learning-Based Classification Of Mesothelioma Improves Prediction Of Patient Outcome},
  author  = {Courtiol, Pierre and Maussion, Charles and Moarii, Matahi and Pronier, Elodie and Pilcer, Samuel and Sefta, Meriem and Manceron, Pierre and Toldo, Sylvain and Zaslavskiy, Mikhail and Le Stang, Nolwenn and others},
  journal = {Nature Medicine},
  year    = {2019},
  volume  = {25},
  number  = {10},
  pages   = {1519--1525}
}

@misc{cordis2019machine,
  title  = {Machine Learning Ledger Orchestration For Drug Discovery},
  author = {CORDIS},
  year   = {2019}
}

@article{chatterjee2023use,
  title   = {Use Of Federated Learning And Blockchain Towards Securing Financial Services},
  author  = {Chatterjee, Pushpita and Das, Debashis and Rawat, Danda B},
  journal = {arXiv preprint arXiv:2303.12944},
  year    = {2023}
}

@article{liu2023efficient,
  title   = {Efficient And Secure Federated Learning For Financial Applications},
  author  = {Liu, Tao and Wang, Zhi and He, Hui and Shi, Wei and Lin, Liangliang and An, Ran and Li, Chenhao},
  journal = {Applied Sciences},
  year    = {2023},
  volume  = {13},
  number  = {10},
  pages   = {5877}
}

@inproceedings{chen-etal-2023-customizedText,
  title     = {A Customized Text Sanitization Mechanism With Differential Privacy},
  author    = {Chen, Sai and Mo, Fengran and Wang, Yanhao and Chen, Cen and Nie, Jian-Yun and Wang, Chengyu and Cui, Jamie},
  booktitle = {ACL Findings},
  year      = {2023}
}

@article{wu2023bloomberggpt,
  title   = {Bloomberggpt: A Large Language Model For Finance},
  author  = {Wu, Shijie and Irsoy, Ozan and Lu, Steven and Dabravolski, Vadim and Dredze, Mark and Gehrmann, Sebastian and Kambadur, Prabhanjan and Rosenberg, David and Mann, Gideon},
  journal = {arXiv preprint arXiv:2303.17564},
  year    = {2023}
}

@article{lee2020biobert,
  title   = {BioBERT: A Pre-Trained Biomedical Language Representation Model For Biomedical Text Mining},
  author  = {Lee, Jinhyuk and Yoon, Wonjin and Kim, Sungdong and Kim, Donghyeon and Kim, Sunkyu and So, Chan Ho and Kang, Jaewoo},
  journal = {Bioinformatics},
  year    = {2020},
  volume  = {36},
  number  = {4},
  pages   = {1234--1240}
}

@article{luo2022biogpt,
  title   = {BioGPT: Generative Pre-Trained Transformer For Biomedical Text Generation And Mining},
  author  = {Luo, Renqian and Sun, Liai and Xia, Yingce and Qin, Tao and Zhang, Sheng and Poon, Hoifung and Liu, Tie-Yan},
  journal = {Briefings in Bioinformatics},
  year    = {2022},
  volume  = {23},
  number  = {6},
  pages   = {bbac409}
}

@article{openai2023gpt4,
  title   = {GPT-4 Technical Report},
  author  = {OpenAI},
  journal = {OpenAI},
  year    = {2023}
}

@inproceedings{maas2011learning_imdb,
  title     = {Learning Word Vectors For Sentiment Analysis},
  author    = {Maas, Andrew L. and Daly, Raymond E. and Pham, Peter T. and Huang, Dan and Ng, Andrew Y. and Potts, Christopher},
  booktitle = {ACL},
  year      = {2011}
}

@inproceedings{bonawitz2019towards,
  title     = {Towards Federated Learning At Scale: System Design},
  author    = {Bonawitz, Keith},
  booktitle = {SysML},
  year      = {2019}
}

@article{gao2023retrieval,
  title   = {Retrieval-Augmented Generation For Large Language Models: A Survey},
  author  = {Gao, Yunfan and Xiong, Yun and Gao, Xinyu and Jia, Kangxiang and Pan, Jinliu and Bi, Yuxi and Dai, Yi and Sun, Jiawei and Wang, Meng and Wang, Haofen},
  journal = {arXiv preprint arXiv:2312.10997},
  year    = {2023}
}

\end{document}